\documentclass[conference]{IEEEtran}
\IEEEoverridecommandlockouts
\usepackage{cite}
\usepackage{amsmath,amssymb,amsfonts}
\usepackage{graphicx}
\usepackage{textcomp}
\usepackage{xcolor}
\usepackage[font={footnotesize}]{caption}

\usepackage[para,online,flushleft]{threeparttable}

\usepackage{hyperref}
\usepackage{url}

\usepackage{algpseudocode}

\usepackage{booktabs}
\usepackage{adjustbox}
\usepackage{subfig}
\usepackage{enumitem}
\usepackage[ruled,vlined]{algorithm2e}
\usepackage[ruled]{algorithm2e}
\usepackage{multirow}
\usepackage{marvosym}
\usepackage{dblfloatfix} 

\usepackage{wrapfig}

\SetKwInput{kwSym}{Notation}
\SetKwInput{kwCalc}{Calculate range of feature embeddings}
\SetKwInput{kwSel}{Select Enemy Examples
 and Assign Sampling Probabilities}

\def\BibTeX{{\rm B\kern-.05em{\sc i\kern-.025em b}\kern-.08em
    T\kern-.1667em\lower.7ex\hbox{E}\kern-.125emX}}
    
  \makeatletter
    \newcommand{\linebreakand}{%
      \end{@IEEEauthorhalign}
      \hfill\mbox{}\par
      \mbox{}\hfill\begin{@IEEEauthorhalign}
    }
    \makeatother

\begin{document}
\title{Efficient Augmentation for Imbalanced Deep Learning}

\author{\IEEEauthorblockN{Damien A. Dablain}
\IEEEauthorblockA{\textit{Dept. Computer Science and Engineering} \\
\textit{University of Notre Dame}\\
Notre Dame, IN 46556, USA \\
ddablain@nd.edu}
\and
\IEEEauthorblockN{Colin Bellinger}
\IEEEauthorblockA{\textit{National Research Council of Canada}\\
 Ottawa, Canada, K1A 0R6\\
colin.bellinger@nrc-cnrc.gc.ca}
\linebreakand
\IEEEauthorblockN{Bartosz Krawczyk}
\IEEEauthorblockA{\textit{Dept. Computer Science} \\
\textit{Virginia Commonwealth University}\\
Richmond, VA 23824, USA \\
bkrawczyk@vcu.edu}
\and
\IEEEauthorblockN{Nitesh V. Chawla}
\IEEEauthorblockA{\textit{Dept. Computer Science and Engineering} \\
\textit{University of Notre Dame}\\
Notre Dame, IN 46556, USA \\
nchawla@nd.edu}

}

\maketitle

\begin{abstract}
Deep learning models tend to memorize training data, which hurts their ability to generalize to under-represented classes.  We empirically study a convolutional neural network’s internal representation of imbalanced image data and  measure the \textit{generalization gap} between a model’s \emph{feature embeddings} in the training and test sets, showing that the gap is wider for minority classes.  This insight enables us to design an efficient three-phase CNN training framework for imbalanced data. The framework involves training the network end-to-end on imbalanced data to learn accurate feature embeddings, performing data augmentation in the learned embedded space to balance the train distribution, and fine-tuning the classifier head on the embedded balanced training data. We propose Expansive Over-Sampling (EOS) as a data augmentation technique to utilize in the training framework. EOS forms synthetic training instances as convex combinations between the minority class samples and their nearest adversaries in the embedded space to reduce the generalization gap. The proposed framework improves the accuracy over leading cost-sensitive  and resampling methods commonly used in imbalanced learning. Moreover, it is more computationally efficient than standard data pre-processing methods, such as SMOTE and GAN-based over-sampling, as it requires fewer parameters and less training time. The source code for the proposed framework is available at: \href{https://github.com/dd1github/EOS}{https://github.com/dd1github/EOS}.
\end{abstract}

\begin{IEEEkeywords}
machine learning, deep learning, class imbalance, over-sampling
\end{IEEEkeywords}

 \section{Introduction}
 \label{sec:int}
\begin{figure}[ht!]
\vspace{-.2cm}
\centering
  \includegraphics[width=0.47\textwidth]{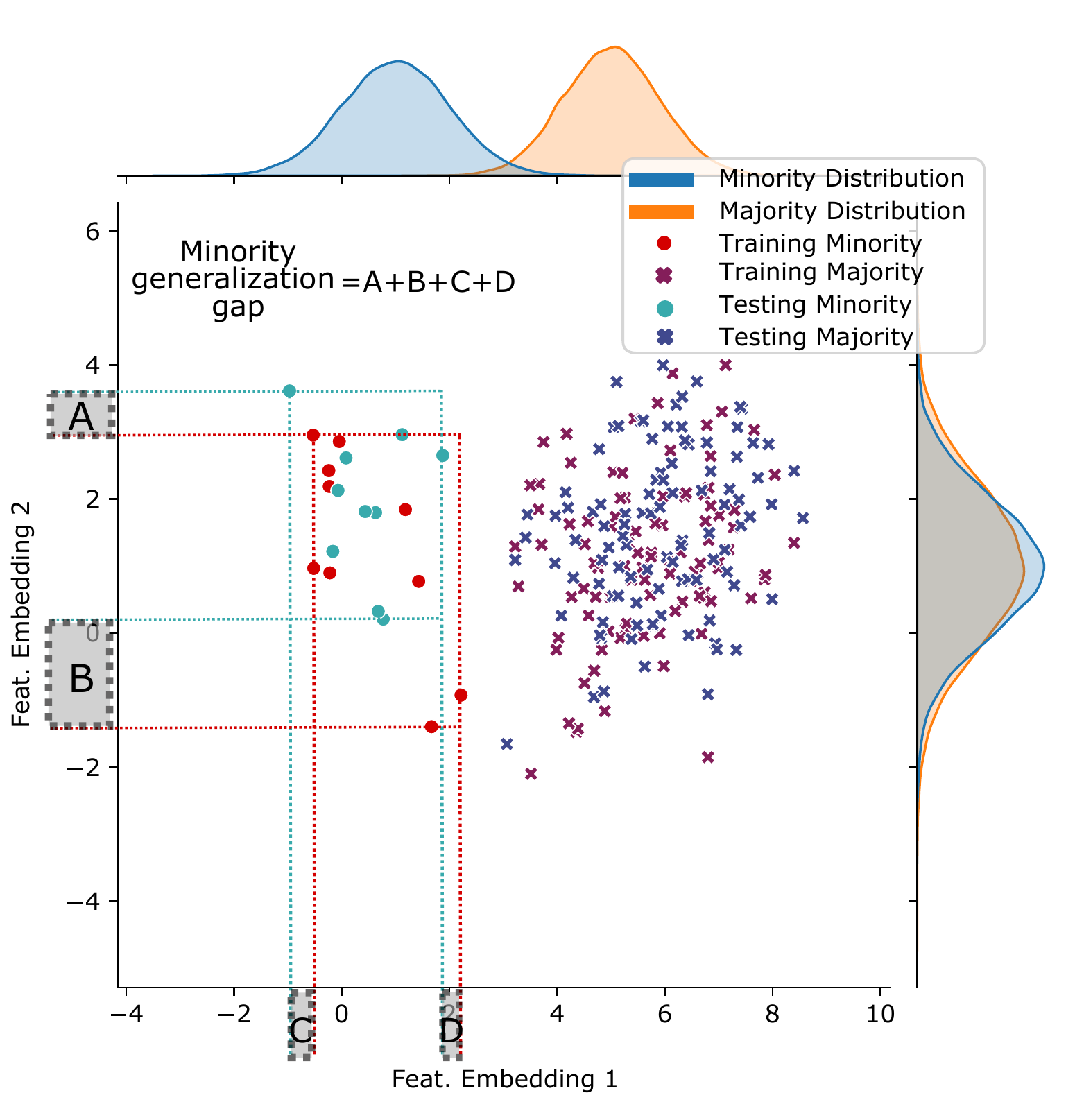}
  \caption{This figure demonstrates the calculation of the generalization gap in a two-dimensional embedding space. For simplicity, we only show the gap for the minority class. By definition, however, we calculate the total generalization gap as the sum of the generalization gap for each individual class. In comparing the training and testing data of the minority and majority classes, this figure shows that with i.i.d sampling of the underlying data distributions, the generalization gap is much larger for the minority class than the majority class.}
  \label{fig:genGap}
  \vspace{-0.2cm}
\end{figure}

Convolutional Neural Networks (CNN) are progressively being combined with imbalanced data sets \cite{johnson2019survey} and yet they have been shown to memorize training data due to over-parameterization \cite{zhang2021understanding}. When learning from imbalanced classification data, a neural network's  affinity for memorization limits its generalization capacity for minority classes \cite{Garcin:2022}. Traditionally, a machine learning model's ability to generalize has been measured by the difference between training and testing accuracy rates \cite{smith2018disciplined}.  In order to gain more insight into the generalization gap in the context of imbalanced deep learning, in this work we measure the generalization gap as the class-wise difference between a model's internal representation of training and testing \textit{features}. Specifically, we measure the difference in the ranges of the embedded features, which is illustrated in Figure ~\ref{fig:genGap}. This provides a wealth of information on where generalization is failing and how it can be rectified. 
 
\begin{figure*}[ht!]
\centering
  \includegraphics[width=0.8\textwidth]{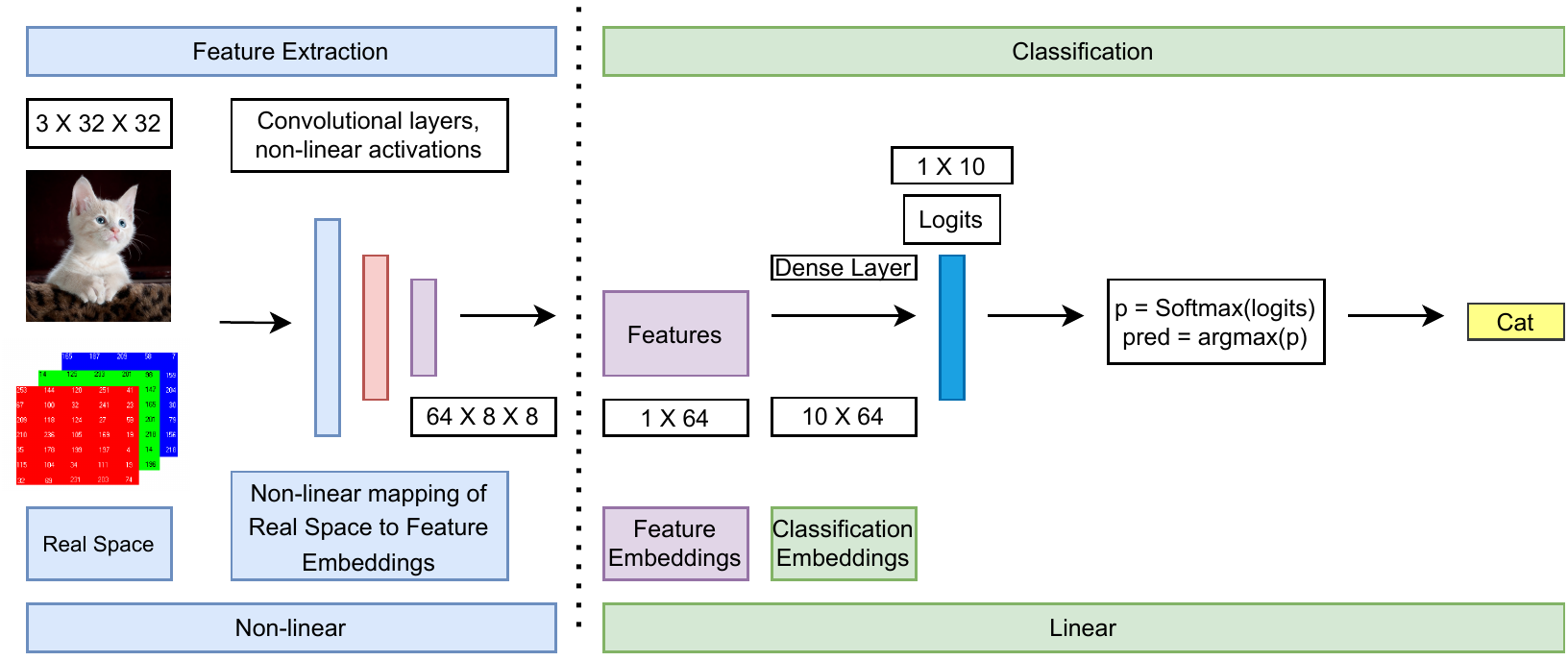}
  \caption{Illustration of feature and classification embeddings, using the Resnet 32 architecture. Feature embeddigs (FE) are extracted by the CNN's non-linear layers and are combined with the final classification layer weights to form classification layer embeddings (CLE), before they are summed.}
  \label{fig:cnn}
  \vspace{-0.2cm}
\end{figure*}

 The generalization gap is particularly large in imbalanced deep learning because training requires a significant number of diverse samples, and yet, there are few minority class examples available.  For majority classes, which have a rich supply of varied examples, the model's internal representation of training and test features is expected to be similar. Therefore the generalization gap in embedding-space will be low and the difference between train and test accuracy rates is expected to be small. On the other hand, the generalization gap for the poorly sampled minority classes is expected to be large. This is because the low-probability edge cases that are needed for good class coverage are unlikely to exist in the training and test sets. We measure the generalization gap at the penultimate layer of a CNN, which outputs \textit{feature} embeddings (see Figure~\ref{fig:cnn} and description in Section~\ref{sec:note}).

Our results demonstrate that a generalization gap exists and that it is correlated with reduced accuracy on  minority classes. Thus, a natural question arises: can we exploit the geometric information in our quantification of the generalization gap to design a minority class feature augmentation procedure that improves model accuracy and generalization? Our analysis illustrates that the answer is yes, and resulted in the development of an efficient CNN training framework for imbalanced data that includes a novel data augmentation algorithm. In the proposed training framework, augmentations are performed in the \textit{embedding} space of deep neural networks, rather than as a pre-processing step in pixel-space. 

Our training framework is based on the recognition that a classifier's ability to generalize is related to the learned feature embeddings. We hypothesize that an effective way to rectify a classifier's generalization gap, in the context of class imbalance, is with data augmentation in embedded space. Specifically, in our framework, we separate the training of a CNN into three phases. Initially, the CNN is trained end-to-end on imbalanced data. Subsequent to convergence, we perform data augmentation in the learned embedding space to acquire additional minority samples. We propose  Expansive Over-Sampling (EOS), an augmentation technique that forms convex combinations between minority instances in the embedding space and their nearest adversary class examples (nearest enemies). By generating convex combinations with nearest enemies, EOS expands the ranges of the minority classes in embedding space in the direction of the neighboring majority classes. Next, the balanced embedding training set is used to fine tune the classification layers of the CNN. Once updated, the full CNN is employed for inference on the test set. Our results show that the proposed training framework improves generalization and accuracy on minority classes.

\noindent \textbf{Contributions.}  This paper offers the following contributions to the field of deep learning from imbalanced data:

\begin{itemize}
    \item \textbf{Generalization Gap Measure.} We propose a novel way to quantify the generalization gap for imbalanced data. Whereas the gap has traditionally been measured as the difference between the accuracy on training and test sets, our method compares the distributions of training and test data in the model's embedding-space. The gap is found to be wider for minority classes and generally follows the class imbalance level. This provides insight into where the model's generalization is failing and can be leveraged to design more robust training techniques. 
    \item \textbf{Three-Phased CNN Training Framework.} We propose an efficient training framework for CNNs with imbalanced data. In particular, the framework involves training the network end-to-end on the imbalanced data, performing data augmentation in the learned embedding space, and fine-tuning the classifier head with the embedding and augmented training data. Our results show that the proposed framework out-performs standard pre-processing and cost sensitive methods for imbalanced learning. 
    
    \item \textbf{Novel Over-sampling Technique.} We propose a novel imbalanced data augmentation algorithm, Expansive Over-Sampling (EOS). EOS is employed in embedding space after the end-to-end training in our CNN framework. It finds nearest adversary class examples (nearest enemies) of the minority class samples to form convex combinations in order to expand the ranges of minority class feature embeddings. EOS outperforms  alternative resampling techniques, such as SMOTE, that can be utilized for augmentation in our framework.  
\end{itemize}

\section{Related work}
\label{sec:bac}
\subsection{Imbalanced Learning}
Imbalanced learning is concerned with techniques that deal with differences in the number of training examples, by class.  When the number of training examples are skewed toward one or a few classes, machine learning models struggle to differentiate minority classes and tend to treat them as "noise" with respect to majority classes.  Many types of class imbalance are possible: exponential and step are some of the more common.  In this paper, we study exponential imbalance, which is most often found in real-world image data \cite{cui2019class,krishna2017visual,chang2017active}. 

The leading methods used to address imbalanced data are resampling \cite{Koziarski:2021}, cost-sensitive algorithms \cite{Siers:2021}, ensemble methods \cite{Wegier:2022}, and decoupling feature representations from classification \cite{Huang:2016,Khan:2018}.  Resampling generally involves under-sampling majority classes or over-sampling minority classes.  These methods attempt to re-balance class weight norms in parametric machine learning models by altering the number of samples per class used during training. 

Over-sampling is a proven technique for combating class imbalance for traditional (i.e., shallow) machine learning models \cite{krawczyk2016learning}. Three leading over-sampling methods are: Synthetic Minority Over-Sampling (SMOTE) \cite{chawla2002smote}, Border-line SMOTE \cite{han2005borderline}, and Balanced SVM  \cite{farquad2012preprocessing}.  All three methods are interpolative in the sense that they generate new samples by randomly inserting data points between minority class samples.  

SMOTE works by randomly selecting a base and a nearest neighbor example from within a minority class and creating an artificial instance with features calculated as the difference between the features of the two instances multiplied by a random number from [0;1]. Border-line SMOTE works in a similar fashion, except that it only performs interpolation from "borderline" examples in the minority class.  An instance is considered to be border-line if its set of \textit{k}-nearest neighbors includes one or more instances from another class.  Balanced SVM uses SMOTE to generate synthetic samples and then replaces the "true" label with a label predicted by a Support Vector Machine model.

These over-sampling methods share a common limitation: since they interpolate between same-class instances, they do not expand the feature \textit{ranges} of the training data. It is well-known, for example, that SMOTE-based methods do not generate instances outside the convex-hull of the individual minority class \cite{belkin2018understand}.  Hence, they can adjust class weight norms, but do not address the \textit{generalization gap} as defined herein.  

Several methods have been introduced to overcome these perceived short-comings.  ADASYN \cite{he2008adasyn} adaptively selects minority examples for over-sampling based on learning difficulty and is generally implemented with binary classification.  Remix \cite{bellinger2021calibrated} aims to improve recall on minority classes by expanding the footprint of the minority class data; however, it is designed to work in pixel space instead of a model's latent embedding space. As a result, the augmentations are expected to be more noisy and less targeted towards reducing the generalization gap than EOS. Like EOS and Remix, manifold-based over-sampling \cite{bellinger2018manifold} aims to improve generalization by expanding the minority class footprint by generating samples along the underlying manifold of the target data. Similar to recent GAN-based over-sampling techniques, however, manifold-based over-sampling requires leaning an additional model for data generation. 

Cost-sensitive algorithms are also used to address class imbalance.  Unlike resampling methods, which are generally applied as a pre-processing step to data, cost-sensitive algorithms modify the model itself by assigning a higher cost to misclassified examples. Cost-sensitive methods have gained considerable interest in deep learning because they are more efficient than resampling methods, and recently deep learning-specific methods have been developed.  Three leading cost-sensitive algorithms are: the Asymmetric Loss (ASL) \cite{ben2020asymmetric}, the Focal loss \cite{lin2017focal}, and the Label-Distribution-Aware Margin loss (LDAM) \cite{cao2019learning}. ASL assigns different costs to positive and negative samples through a targeted decay rate.  The focal loss assigns higher losses to hard examples (which presumably are from the minority class) and lower relative costs to easy examples (which presumably are from classes with a greater number of samples).  LDAM encourages higher margins for minority classes and also employs a class reweighting strategy that is deferred until later training epochs.

Another approach to imbalanced learning involves separating a neural network into feature extraction and classification stages. Leading methods in this area include: Open Long Tailed Recognition (OLTR) \cite{liu2019large}, the Bilateral Branch Network (BBN) \cite{zhou2020bbn}, and Decoupling Recognition and Classifier for Long-Tailed Recognition (Decoupling) \cite{kang2019decoupling}. BBN employs a learning branch, a rebalancing branch and a weighted cross-entropy loss to retrain the classifier. The Decoupling technique uses a variety of methods to retrain a classifier, including: random reinitialization and optimization of the weights, computation of the mean feature representation of each class, and direct adjustment of the classifier norms. Our CNN training framework for imbalanced data takes inspiration from the above work on model decoupling. Our approach, however, includes a specialized data augmentation process in the learned embedded space and classifier fine-tuning that efficiently reduces the generalization gap.

\subsection{Neural Network Memorization \& Generalization}
Being over-parameterized, deep neural networks have significant power to represent complex data distributions \cite{choromanska2015loss, du2019gradient,du2018gradient}. On the other hand, their capacity renders them susceptible to over-fitting and data memorization \cite{goodfellow2016deep}. As a result, studying and analyzing the over-fitting and memorizing behaviour of deep neural networks is an active field of research \cite{domingos2020every,jacot2018neural,zhang2021understanding}. This includes the study of generalization and the generalization gap \cite{arpit2017closer,smith2018disciplined,hoffer2017train,ye2020identifying}, which is related to our work. In \cite{arpit2017closer,smith2018disciplined}, the authors examine the generalization gap in the context of the training and test losses. This provides a good overall assessment of the model's generalization at the classification level. In contrast, our work focuses on the generalization of the feature extraction layers, which we deem to be a prerequisite for good generalization at the classifier level. 

In \cite{ye2020identifying}, the authors quantified the generalization gap based on the per-class feature means in the training and testing data using squared Euclidean distance.  Our generalization gap is similar to their feature deviation computation, except that we do not use the class mean of features. By utilizing the feature ranges (maximum and minimum), we are able to assess deviations in the footprints of the embedded classes. This is a critical concept in the context of imbalanced learning. In addition, we use the Manhattan distance with a floor (i.e., zero distance). The floor takes effect if the test feature embedding minimum or maximum falls within the training feature embedding range.  The Manhattan distance has the advantage of being less sensitive to outliers and the floor prevents the generalization gap from being reduced when the test distribution falls within the training range. 

In general, much of the work on the generalization gap in deep learning has focused on balanced classification problems. It is well-know, however, that generalization is a major issue in the context of imbalanced classification. In the case of majority classes, a neural network may have a sufficient number and variety of training examples such that it can generalize well to the test set.  On the other hand, given i.i.d. sampling of the underlying distribution, the sparsely sampled minority classes are likely to have a wide gap between the training and test distributions.  The generalization gap problem is further compounded in imbalanced learning because the majority classes may contain sub-concepts that overlap with minority classes \cite{lopez2012analysis,japkowicz2002class, prati2004class,weiss2003learning}.  Feldman has shown that minority classes, which share features with majority class sub-concepts, can be treated as "noisy" majority class examples that get misclassified  \cite{feldman2020does}.

Our work quantifies the generalization gap on a class-wise basis according to the difference between the footprint of the training and test data in the model's embedding space. This provides greater insight into generalization in imbalanced deep learning.  

\section{Method Description}
\label{sec:eos}

\subsection{Notation}
\label{sec:note}
We adapt notations used by Ye et al. \cite{ye2020identifying} and Badola et al. \cite{badola2021identifying}.  A dataset, $D=\{X,Y\}$ is comprised of instances, $X$, and labels, $Y$. An instance of $D$, $d=\{x,y\} \in \{X,Y\}$, consists of an image, where $x \in \mathbb{R}^{cXhXw}$, such that $c$, $h$, and $w$ are image channels, height and width, respectively. $D$ can be partitioned into training and test sets ($D=\{Train,Test\}$).

A CNN can be described as a network of weights, $W$, arranged in layers, $L$, that operate on $x$ to produce an output, $y$. The layers, $L$, of a CNN can be decomposed into two principal parts: extraction layers and a classification layer.  The extraction layers, $f_\theta(\cdot)=AW^l(...AW^1(AW^0(x))) \rightarrow P$, apply weight kernels $W$ and activation functions $A$, where the final extraction layer performs a pooling function, $P$. We refer to the output of $f_\theta(\cdot)$ as feature embeddings (FE), which are of size $b \cdot  d$, where $b$ is a batch of transformed $x$ and $d$ is the dimension of the pooling layer. The final classification layer outputs a label $y=\sigma(FE \cdot W_c)$, where $\sigma$ is the softmax function, $exp(FE) / \Sigma exp(FE)$.  Classification layer embeddings (CLE) represent the output of the classification layer before summation and softmax are applied.  CLE are of size $b \cdot d$, which is the same dimension as the pooling layer.

\subsection{Generalization Gap}
  This section attempts to take a step toward answering the question: if neural networks memorize training data, how do they generalize to unseen examples?

\begin{algorithm}[h!]
\scriptsize
\caption{Generalization Gap}\label{alg:gen_gap_algo}

\DontPrintSemicolon 
\kwSym{\\
$Train$ = training data;\\
$Test$ = test data;\\
$D$ = \{Train, Test\};\\
$FE$ = feature embeddings;\\
$N_{Train}$ = number of Train examples; \\ 
$N_{Test}$ = number of Test examples; \\ 
$C$ = class labels; \\ 
$FRange_{Trn}$ = Train feature embedding ranges; \\ 

$FRange_{Tst}$ = Test feature embedding ranges; \\ 

}
\BlankLine
\kwCalc{}
\For{$d\ in\ D$}{
\For {$c\ in\ C$}{
$FE_c \gets FE[c]$\;
$cnt = 0$\;
\For {$f\ in\ FE$}{
 $min \gets minimum(f)$\;
 $max \gets maximum(f)$\;
 $FR[c, cnt] \gets min$\;
 $FR[c, cnt+1] \gets max$\;
 $increment\ cnt\ by\ 2$\;
 }}
 
 \eIf{$d$ equals Train}{
 $FRange_{Trn}$ = $FR$\;

 }{{
 $FRange_{Tst}$ = $FR$\;
 } }  }
$GGap = minimum(FRange_{Trn} - FRange_{Tst},0)$\; 
$GGap = mean(GGap)$\ 
\end{algorithm}


We hypothesize that a neural network's ability to generalize lies in the \textit{range} of the internal representations learned during training.  Neural networks can generalize better if the range of the representations closely match the range of the representations present during inference.  The generalization gap can then be quantified as the difference in the model's internal representation between the training and test data. This is illustrated in Figure \ref{fig:genGap}.  We measure the generalization gap based on feature embeddings, or the output of a CNN's penultimate layer (depicted in Figure ~\ref{fig:cnn}).  The generalization gap is simply calculated as the mean of the difference between the range of each feature embedding for the training and test sets, with a zero floor (see Algorithm ~\ref{alg:gen_gap_algo}).

More specifically, once a CNN is trained, the FE of the training and test instances are extracted. For each class in each of the training and test sets, the FE are compiled, such that the size of the compiled feature embeddings are $FE_{Train}=FE_d \cdot N_{Train}$ and $FE_{Test}=FE_d \cdot N_{Test}$, where $N_{Train}$ and $N_{Test}$ are the number of instances in the training and test sets, respectively.  The maximum and minimum of each feature embedding is selected for the training and test sets.  The sum of the differences in FE maximums and minimums in the training and test sets is calculated by class and the mean of the class differences is the net generalization gap for the dataset.

Our formulation of the generalization gap is well-suited to neural networks that accept bounded inputs (i.e., pixels with real values between 0 and 255), where the inputs are normalized before entry to the model, and that use batch normalization \cite{ioffe2015batch} and ReLU activation functions, which further standardize and constrain the model's internal representations. 

All of these factors reduce the impact of potential data outliers and make the comparison of feature embedding values between the train and test sets more uniform.  The Resnet architecture \cite{he2016deep} that was used in our experiments meets these criteria (i.e., normalized pixel inputs, layers with batch normalization and ReLU activation function).

\subsection{Three-Phased CNN Training Framework}
Over-sampling is a proven method for addressing class imbalance \cite{he2009learning}. However, an often cited drawback of over-sampling methods is that they cause a substantial increase in training time for deep learning models because they dramatically increase the number of training batches, which is a natural result of increasing the number of minority class samples \cite{cao2019learning,cui2019class}.  A number of prior attempts to incorporate over-sampling into deep learning have focused on pre-processed minority class augmentations, which generally increase training time \cite{mariani2018bagan,mullick2019generative,bellinger2020framework,dablain2022deepsmote}.  

To address these issues, we propose a three-phased CNN training framework for imbalanced data. First, a CNN is trained with imbalanced data so that it can learn class-specific features.  Once trained, the training set feature embeddings are extracted from the model. Second, the imbalanced FE are augmented so that the number of samples in each class are balanced.  For this purpose, any suitable over-sampling technique can be employed. Then, the classification layer is separated from the model and re-trained with augmented data for a limited number of epochs.  Finally, once re-trained, the extractor and classifier are recombined so that inference can be performed.

\subsection{Expansive Over-Sampling}

\begin{algorithm}[h]
\scriptsize
\caption{Expansive Over-Sampling}\label{alg:EOS_algo}
\DontPrintSemicolon 
\kwSym{\\
$B$ = base examples;\\
$P$ = sampling probabilities;\\
$K$ = number of nearest neighbors;\\
$NNb$ = K Nearest Neighbors of an example;\\
$FE$ = feature embeddings; \\ 
$C$ = class labels; \\ 
$N$ = NNb of size length(B) X length(C); \\ 

$Enemy/ NNb$ = Non-Same Class NNB; \\ 

}
\BlankLine
\kwSel{}

\For {$c\ in\ C$}{
$FE_c \gets FE[c]$\;
$FE_{NNb} = Enemy\ NNb\ of\ FE_c$
$determine\ NNb\ for\ examples\ in\ FE_c$\;
\If{an example in $FE_c$ has Enemy NNb}{
\For {$each\ NNb\ of\ each\ example\ in\ FE_c$}{
 $p \gets 0\ for\ same\ class\ NNb\ OR$\;
 $p \gets uniform\ prob. \ for\ Enemy\ NNb$\;
 $B.append(FE_c\ base\ example)$\;
 $N.append(Enemy\ NNb\ of\ base\ example)$\;
 
 }} } 
\For {$c\ in\ C$}{
$B_c \gets randomly\ select\ N_s\ from\ B$\;
$NNb_c \gets randomly\ select\ N_s\ from\ NNb\ w \backslash  P$\;
$R \gets randomly\ select\ [0,1]$\;
$samples \gets B + R * (B - N)$\;
$Samples.append(samples)$\;

 } 

\vspace{-.2cm}
\end{algorithm}

We propose a method that takes into account the three-phased training framework and additionally that generates synthetic training samples for the classifier that reduce the generalization gap, improve performance and is more efficient.  

EOS works on a CNN's \textit{feature embeddings}.  It identifies minority class examples in the training set whose nearest neighbors contain adversary class members. EOS then creates synthetic examples built on the difference between a minority class example and one of its nearest enemy class neighbors (see Algorithm ~\ref{alg:EOS_algo}).  By pairing examples that are on the border of a minority class with neighbors from another class, the range of minority class features is expanded.  In contrast, many existing over-sampling methods create synthetic examples solely from same class neighbors (intra-class), and hence, they  inherently do not expand inter-class decision boundaries.

EOS generates synthetic minority class examples from nearest adversaries because we believe these examples are more discriminative. Nearest adversary examples likely rest on the class decision boundary because they represent non-same class data instances that are most similar to the reference class. EOS essentially generates synthetic examples in a lower dimensional embedding space (FE), which the classifier uses to reach its decision. By creating more examples that the model has difficulty classifying, EOS increases the related loss, which should modify classifier weights such that they are better able to discern borderline instances.

EOS consists of two essential parts: a novel three-phased training procedure and a resampling technique that relies on nearest adversary class instances.  The EOS training procedure is unique because it augments minority samples in CNN feature embedding space instead of pixel space.  Thus, it is unlike over-sampling methods that rely on GANs\cite{mariani2018bagan,mullick2019generative}.  With EOS, a CNN is first trained on imbalanced image data to learn how to extract and classify images.  The loss function can either be cross-entropy or a cost-sensitive algorithm.  The performance of the CNN is then enhanced by partitioning it into extraction and classification layers. The extraction layers output feature embeddings with imbalanced classes.  These imbalanced FE are over-sampled using nearest adversaries. The classifier is re-trained for a limited number of epochs with real and synthetic feature data with a much lower dimension than in pixel space.

Overall, the EOS framework can be summarized as: train a CNN on imbalanced data, re-sample in embedded space, update the classifier layer only with synthetic samples, and then re-assemble the extraction and updated classification layers for inference. We propose EOS because: a) it expands the minority feature embedding class space in a more precise manner, and b) it enables more efficient training. 

\section{Experimental study}
\label{sec:exp}

We designed an experimental study to answer the following research questions:

\begin{itemize}
    \item RQ1:  Is the generalization gap for minority classes greater than for majority classes? 
    \item RQ2:  Is resampling in CNN embedding space superior to image space? 
    \item RQ3:  Does interpolation with nearest adversaries reduce skew in the minority distribution, lessen the generalization gap and improve prediction bias?
        \item RQ4:  How does the proposed EOS compare with state-of-the-art GAN-based over-sampling that rely on artificial image generation?
\end{itemize}

\subsection{Algorithms, datasets \& training procedure}
\label{sec:set}
To address these questions, we examine four popular image datasets, CIFAR-10 \cite{krizhevsky2009learning}, SVHN \cite{netzer2011reading}, CIFAR-100 \cite{krizhevsky2009learning} and CelebA \cite{liu2018large}.  CIFAR-10 and SVHN each contain ten classes.  The training sets are exponentially imbalanced (maximum imbalance ratio of 100:1), similar to Cui et al. \cite{cui2019class}.  CIFAR-100, which contains 100 classes and has 10 times fewer training examples per class than CIFAR-10 and SVHN, was exponentially imbalanced with a ratio of 10:1 to allow for sufficient training examples in the minority classes. For the CelebA dataset, we selected 5 classes based on hair style (black, brown, blond, gray, bald) with an maximum imbalance ratio of 40:1.

Because the internal representation learned by a CNN may be influenced by the learning algorithm, four different loss functions commonly used in imbalanced learning are tested: cross-entropy (CE) and three cost-sensitive methods - ASL, Focal Loss, and LDAM. In addition, to assess the impact of over-sampling at different points in the training process and as baselines to EOS, three popular over-sampling algorithms are used:  SMOTE, Borderline-SMOTE and Balanced SVM. Additionally, we use three GAN-based over-sampling methods: GAMO \cite{mullick2019generative}, BAGAN \cite{mariani2018bagan}, and CGAN \cite{Dong:2022}.

All models are trained based on a training regime established by Cui et al. \cite{cui2019class} (200 epochs, except for CelebA, which is trained for 50 epochs). Before final selection, all models and datasets are run on three different cuts of the training set.  Since the variation in balanced accuracy was less than 2 points for all cuts, a single cut is selected for experimentation. The best performing model for each dataset and algorithm is selected for further investigation. Model performance is assessed using three skew-insensitive metrics: balanced accuracy (BAC), geometric mean (GM), the F1 measure (FM) \cite{sokolova2009systematic}.

\subsection{Experimental setup}\label{esetup}
\noindent \textbf{Generalization gap experimental setup.} To measure the generalization gap, a CNN ( Resnet-32 for CIFAR-10, CIFAR-100 and SVHN and Resnet-56 for CelebA) is first trained to extract features based on four cost-sensitive algorithms (CE, ASL, Focal Loss and LDAM) and four datasets. The accuracy of each algorithm is recorded. Feature embeddings from the trained models are extracted for both training and test sets.  Multiple algorithms are used so that accuracies and generalization gaps can be compared.

\smallskip
\noindent \textbf{Feature vs. pixel space resampling experimental setup.} To determine if there is a difference in test accuracy when over-sampling is performed in pixel versus embedding space, the same procedure as above is followed, except that a single algorithm is selected and a model is trained with over-sampling at different stages.  More specifically, the cross-entropy loss function is used to train a model where over-sampling is performed in pixel space with SMOTE, Borderline SMOTE, Balanced SVM, and Remix.  Then, the same model architecture and loss function is used to train a model to classify images on the same datasets, but with minority class over-sampling performed in feature embedding space.  For this purpose, the trained model is separated into extraction and classification layers.  The classification layer is retrained with synthetic data generated with respect to feature embeddings instead of in pixel space. The final classification layer is re-trained for a greatly reduced number of epochs (10). The EOS algorithm is run with K nearest neighbors equal to 10. See Section ~\ref{sec:knn} for further discussion.

\smallskip
\noindent \textbf{EOS experimental setup.} For this experiment, models with the same CNN architecture are trained on four loss functions.  Then, the trained models are separated into extraction and classification layers.  The classification layer is retrained with three over-sampling methods and EOS.  The norms, accuracies and generalization gaps of the models trained on cost-sensitive algorithms are compared with models trained with over-sampling in feature embedding space.  The impact of over-sampling in feature embedding space and with EOS are then compared to the baseline approaches. 

\section{Results and discussion}
\label{sec:res}

\subsection{Generalization Gap}

\begin{figure*}[t!]
   \vspace{-.5cm}
  \subfloat[CIFAR10:Base]{\includegraphics[width=0.2\textwidth]{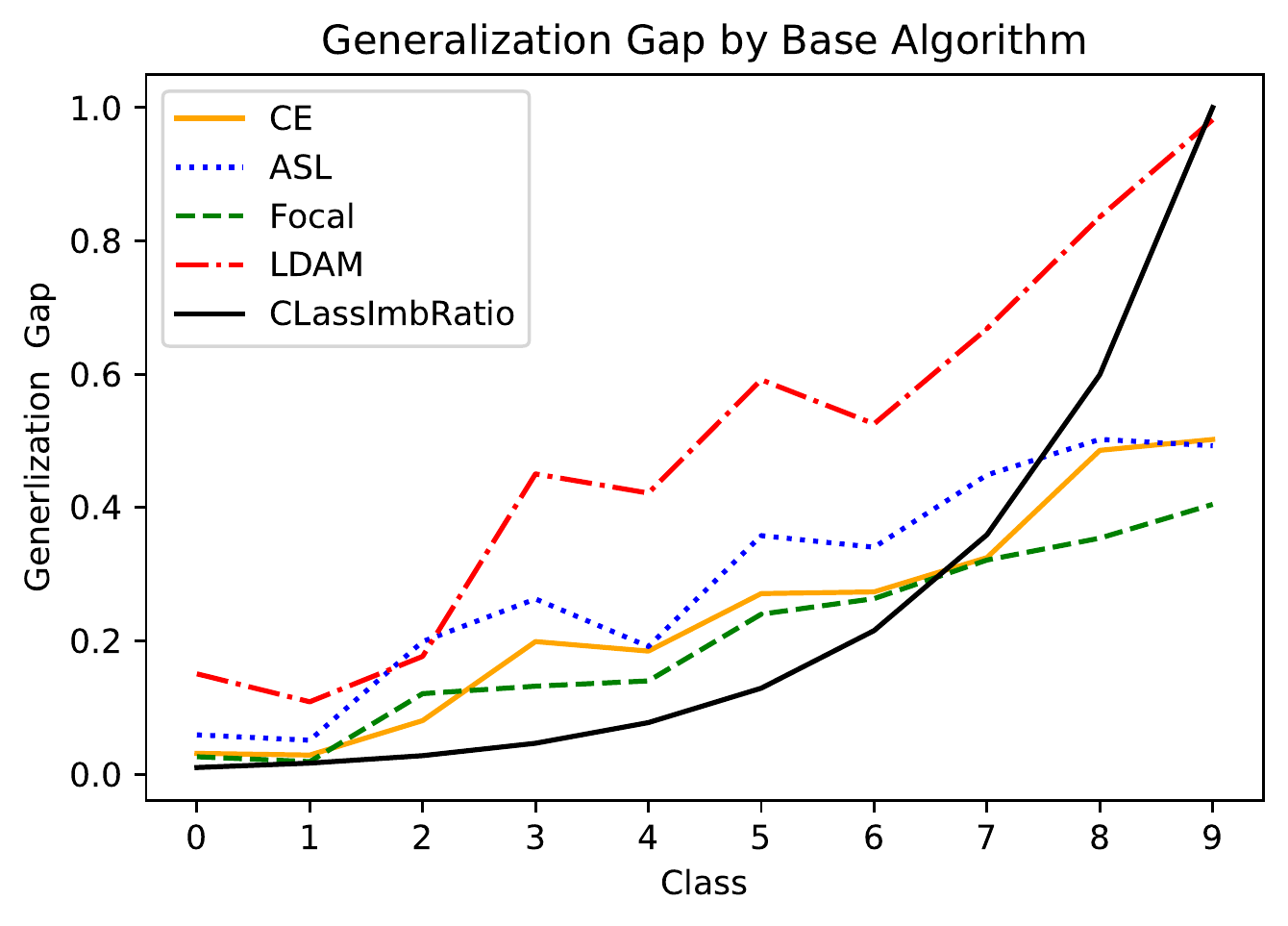}\label{fig:f31}}
  \hfill
  \subfloat[CIFAR10:ASL]{\includegraphics[width=0.2\textwidth]{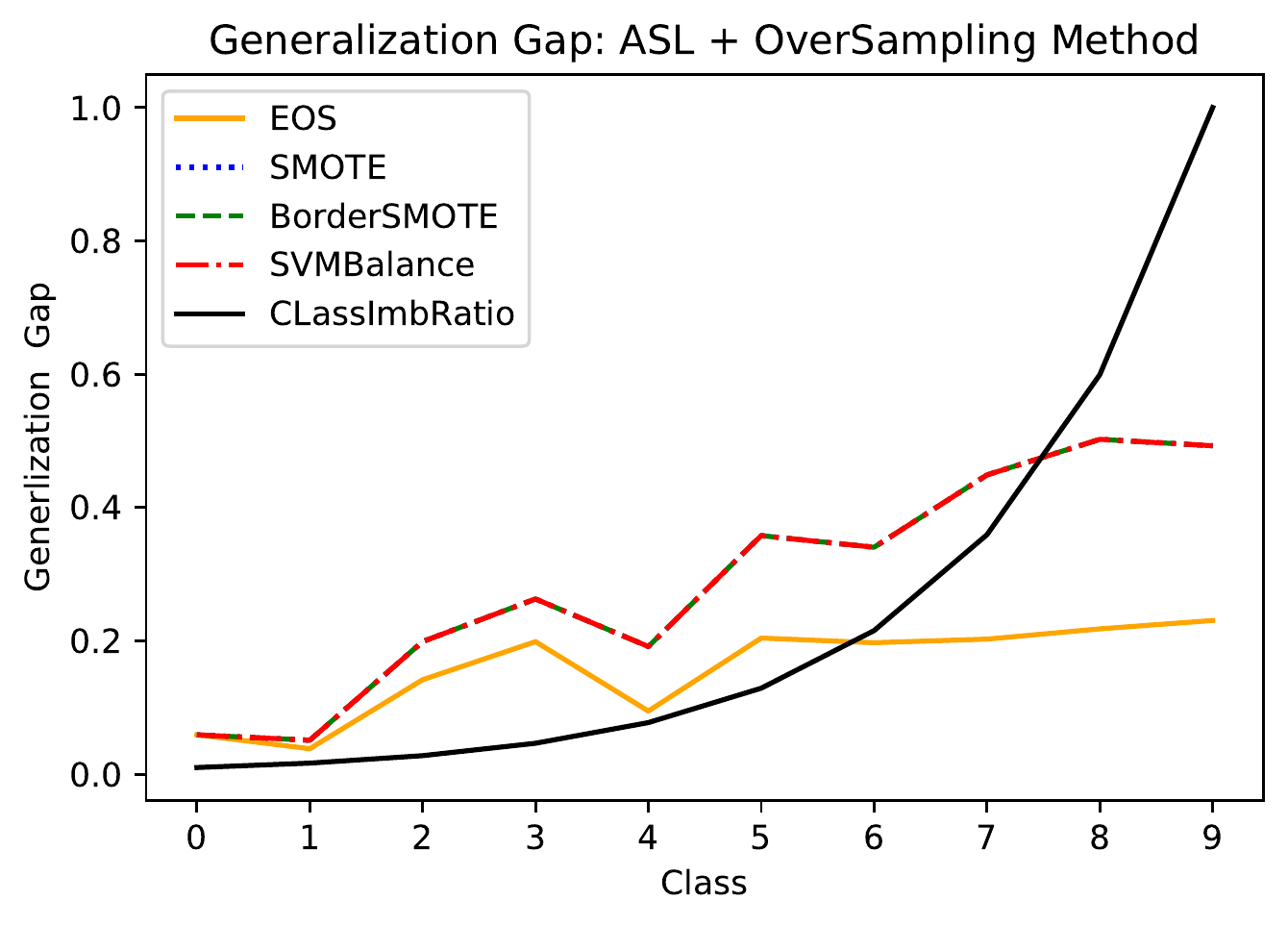}\label{fig:f32}}
  \hfill
  \subfloat[CIFAR10:CE]{\includegraphics[width=0.2\textwidth]{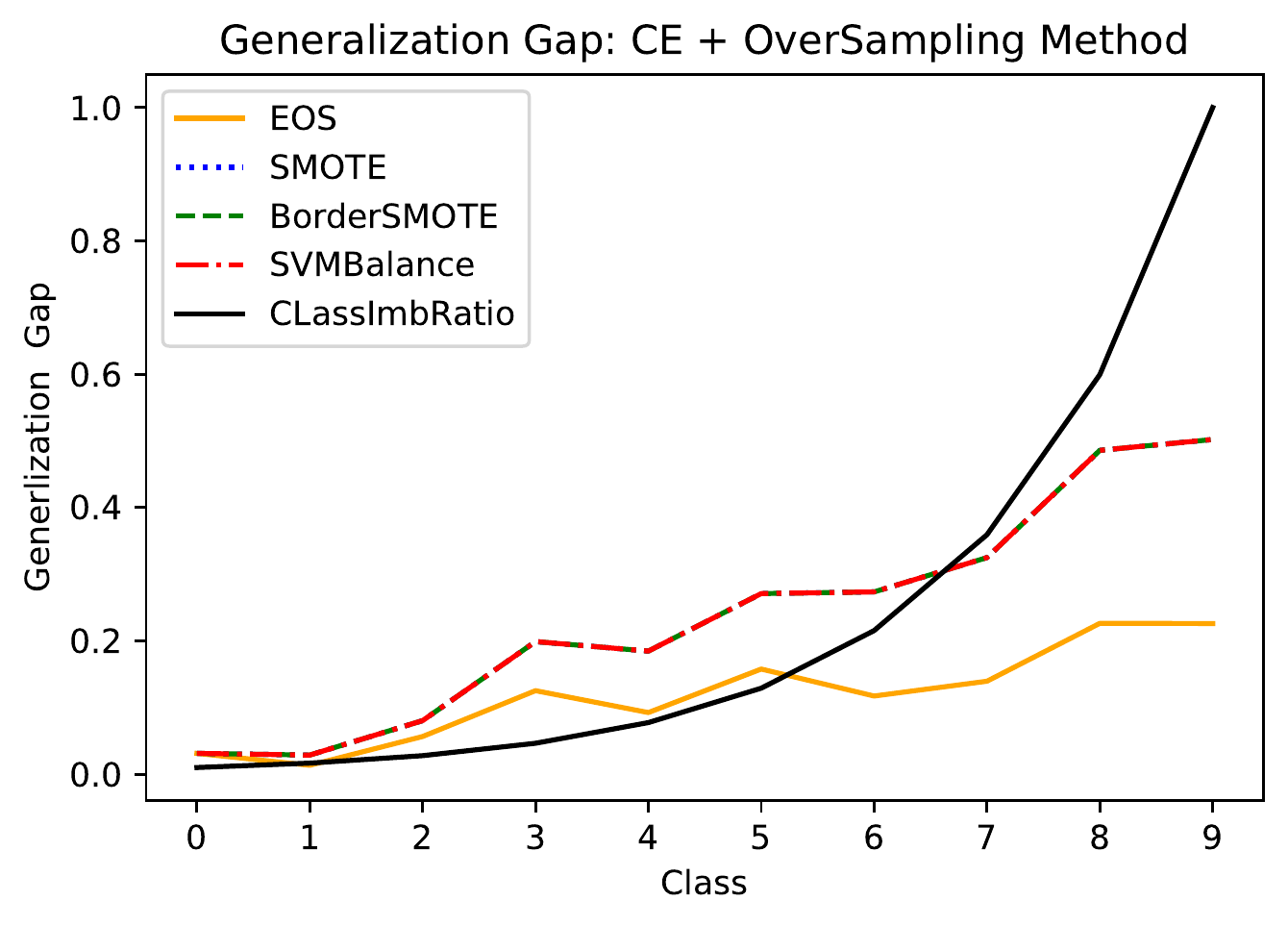}\label{fig:f33}}
   \hfill
  \subfloat[CIFAR10:Focal]{\includegraphics[width=0.2\textwidth]{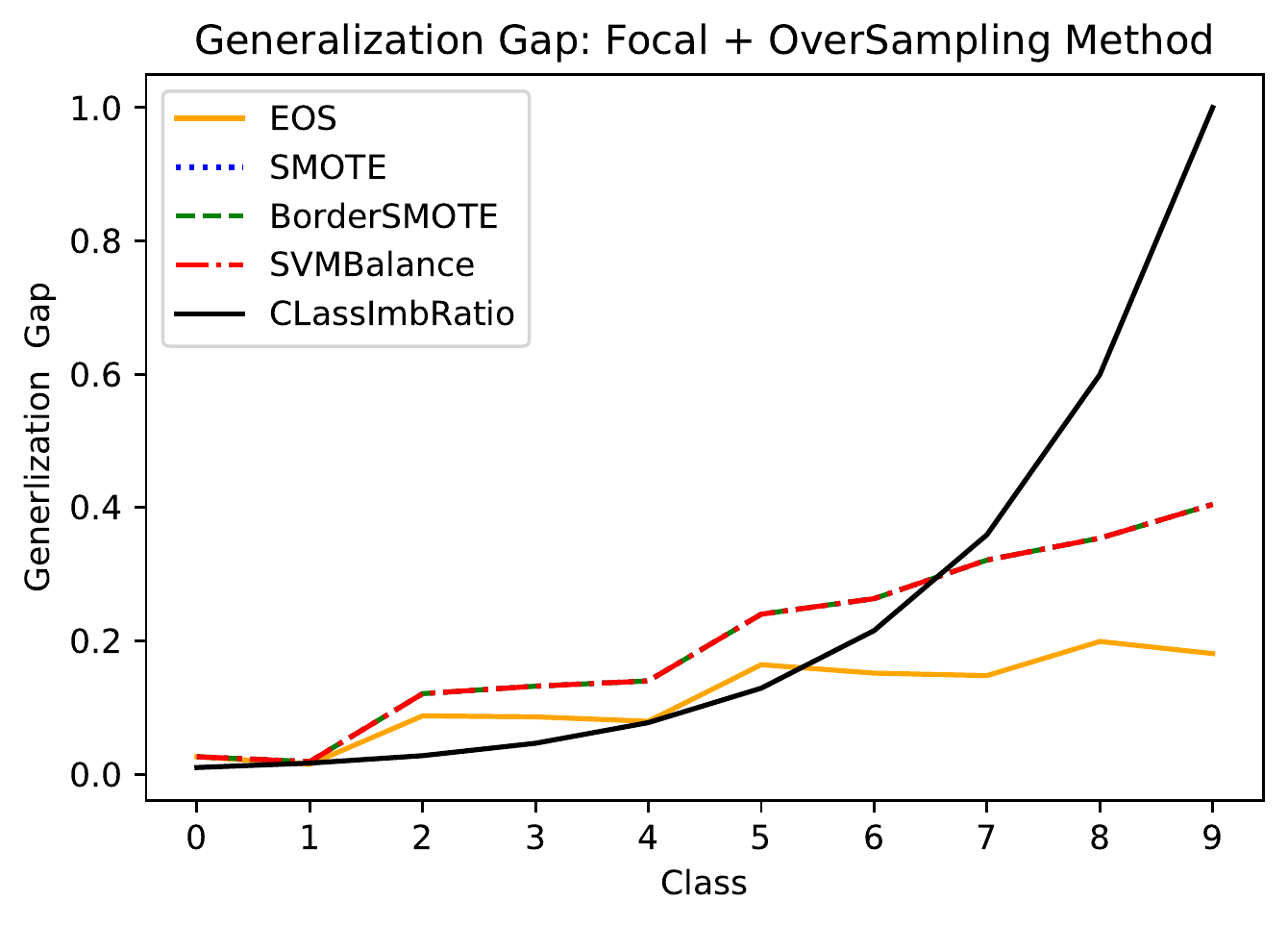}\label{fig:f34}}
   \hfill
  \subfloat[CIFAR10:LDAM]{\includegraphics[width=0.2\textwidth]{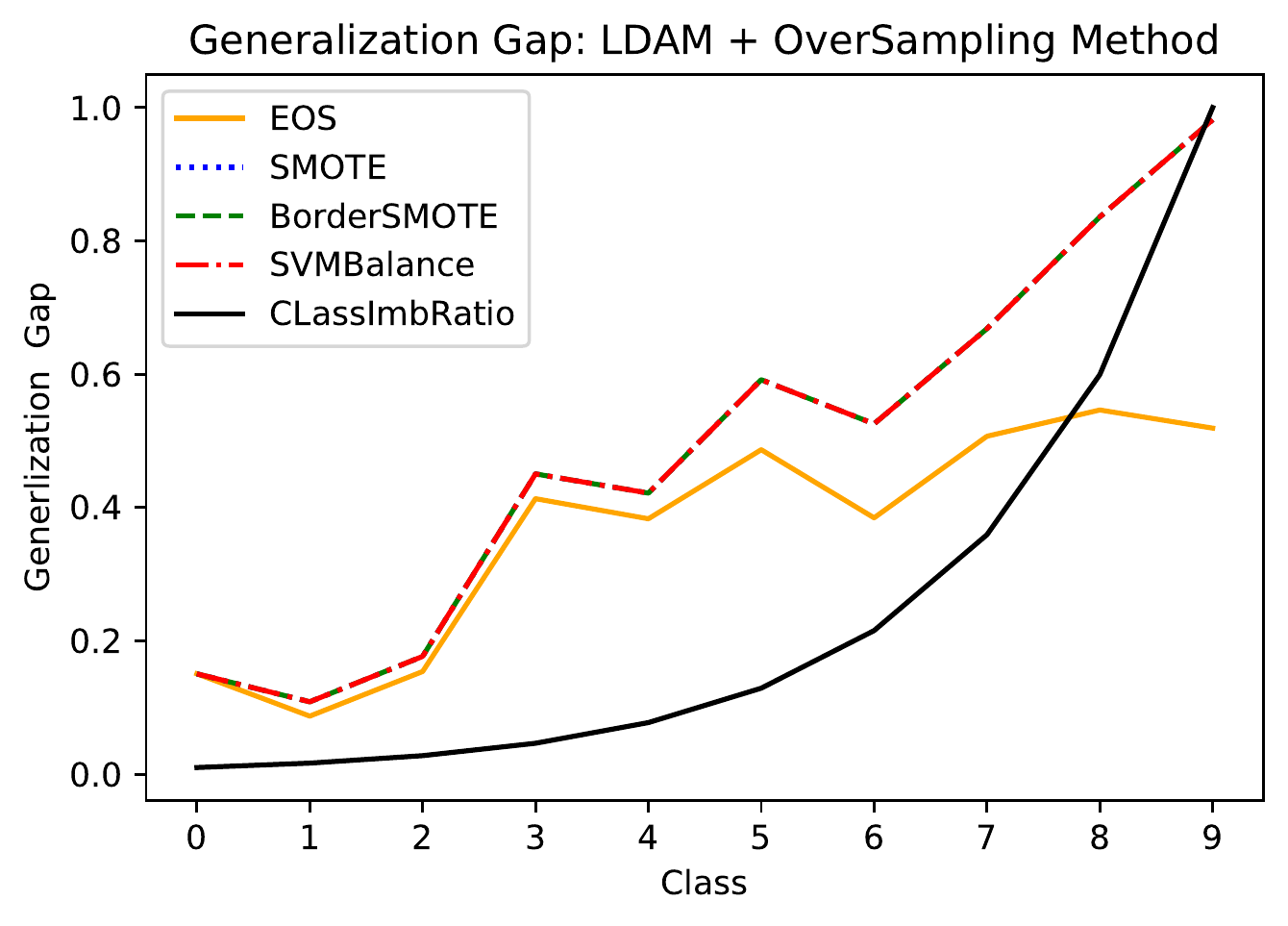}\label{fig:f35}}
  \hfill
  \subfloat[SVHN:Base]{\includegraphics[width=0.2\textwidth]{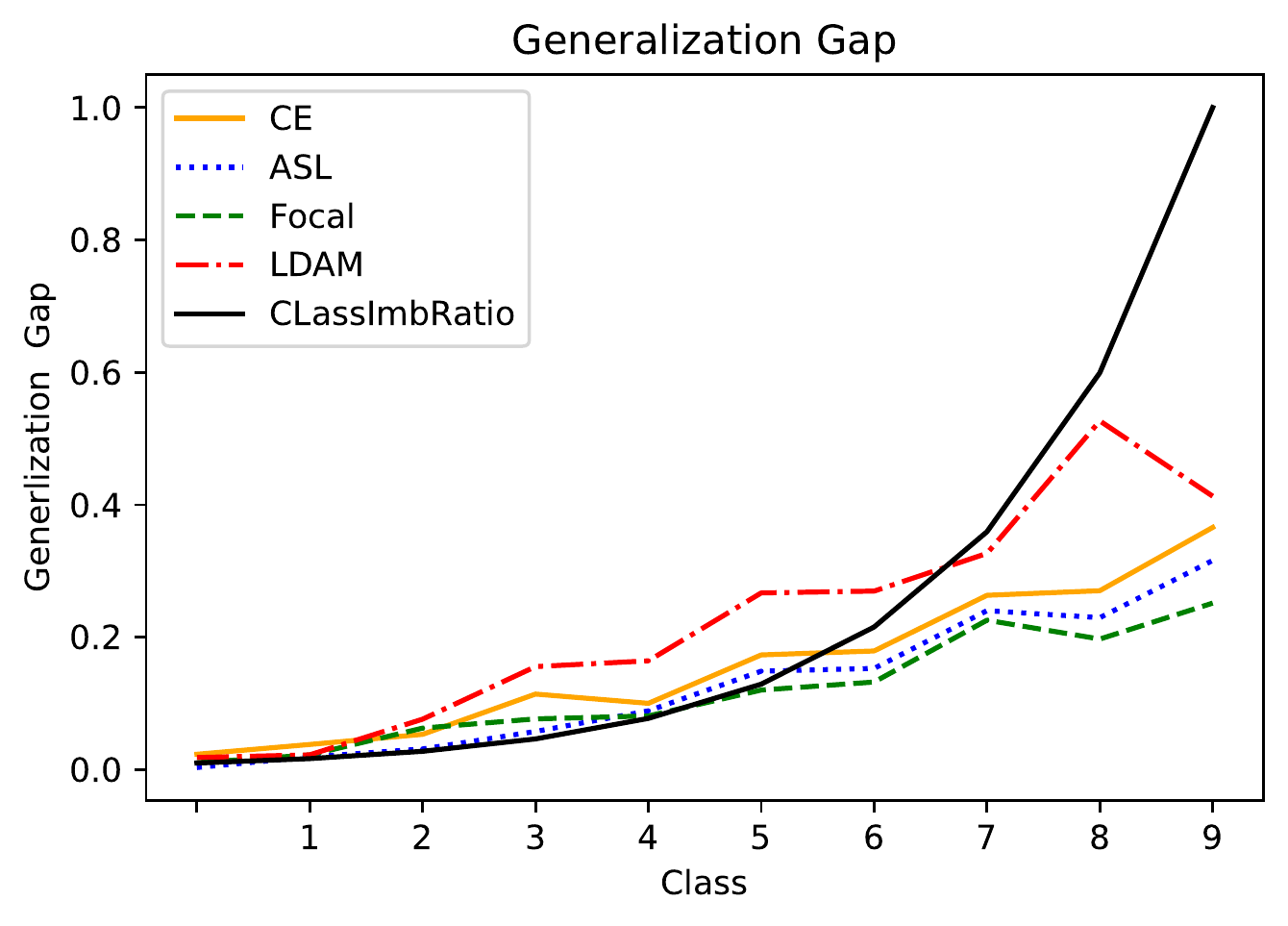}\label{fig:f35a}}
  \hfill
  \subfloat[SVHN:ASL]{\includegraphics[width=0.2\textwidth]{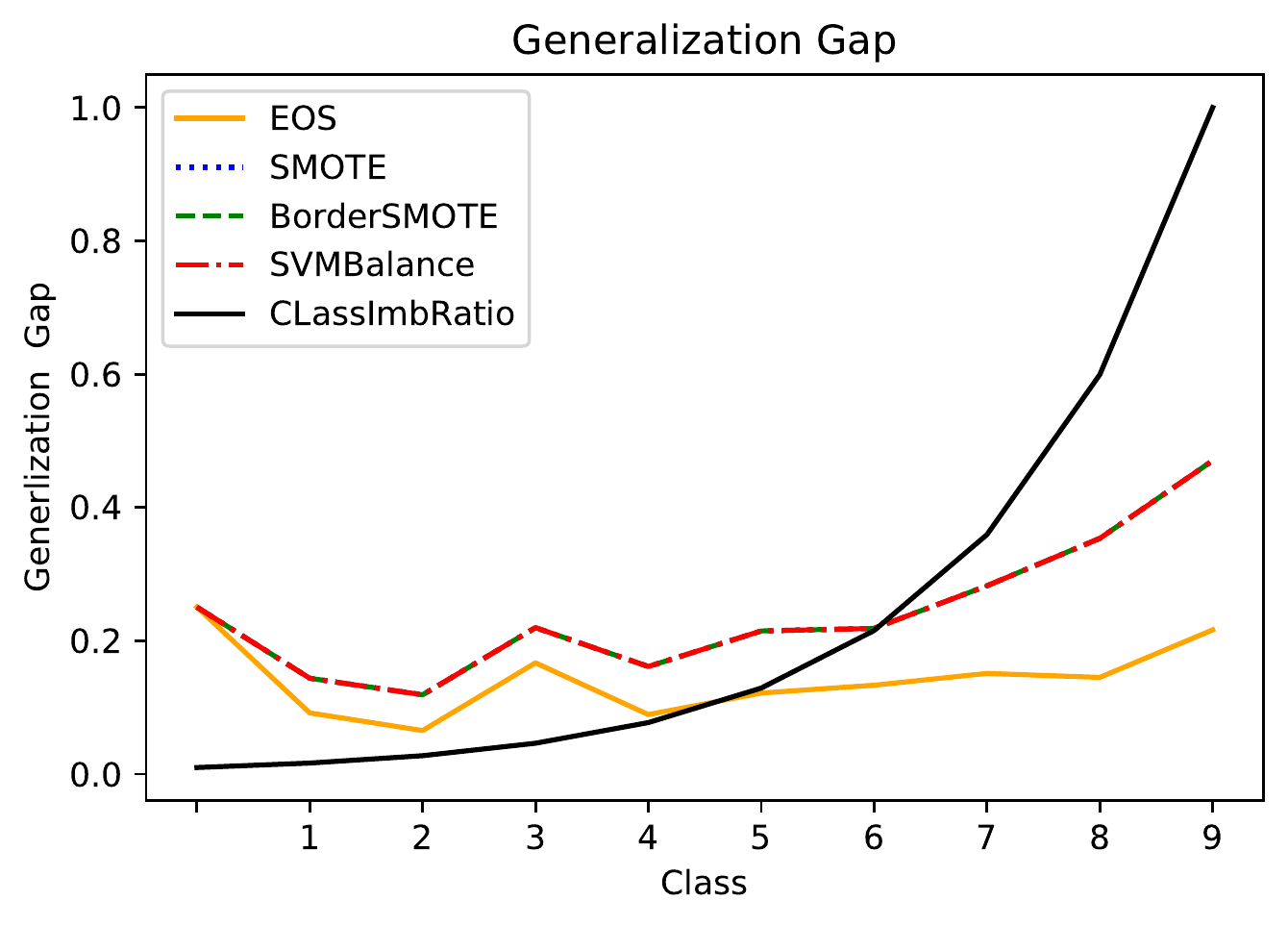}\label{fig:f35a}}
  \hfill
  \subfloat[SVHN:CE]{\includegraphics[width=0.2\textwidth]{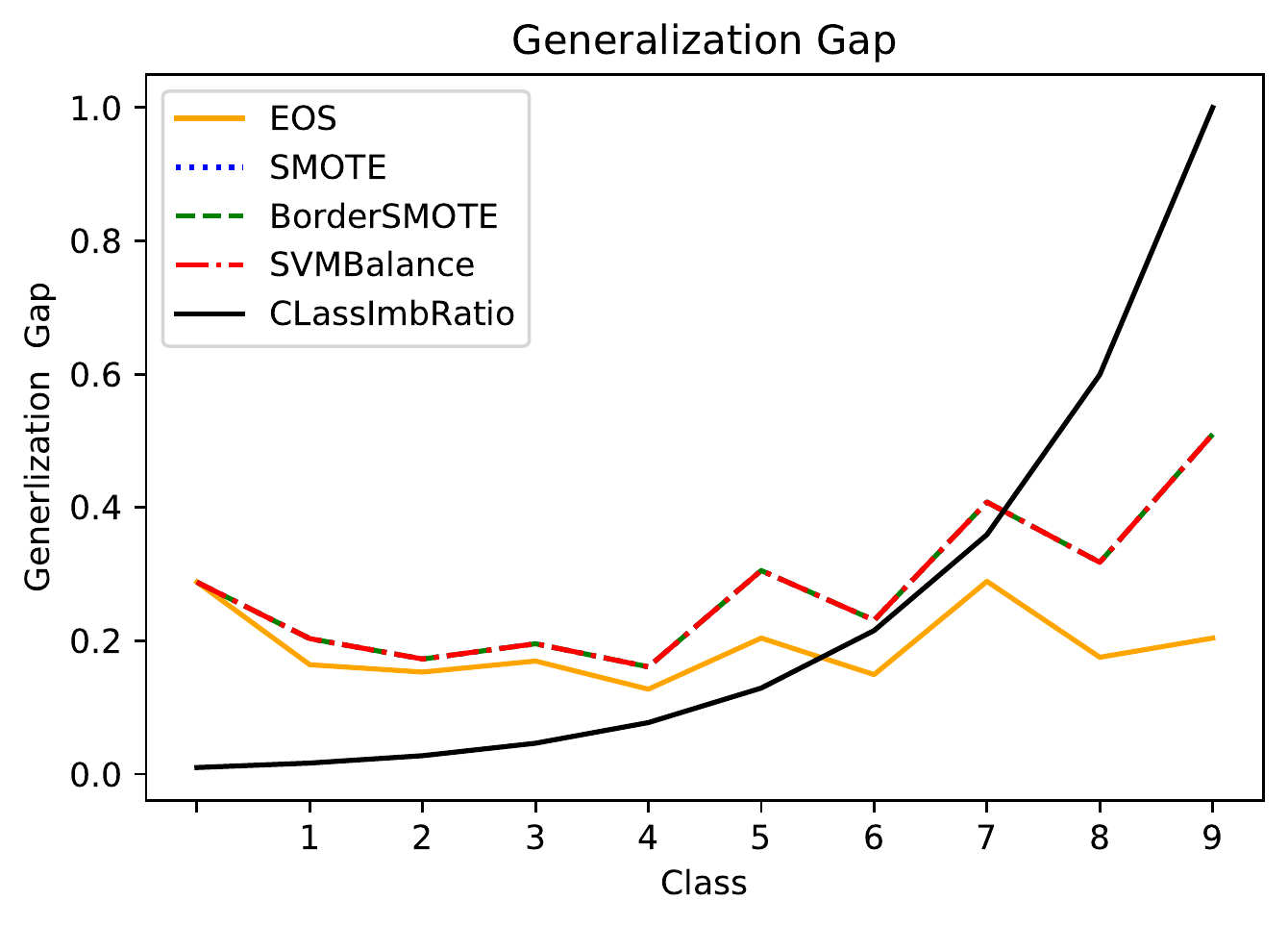}\label{fig:f35b}}
  \hfill
  \subfloat[SVHN:Focal]{\includegraphics[width=0.2\textwidth]{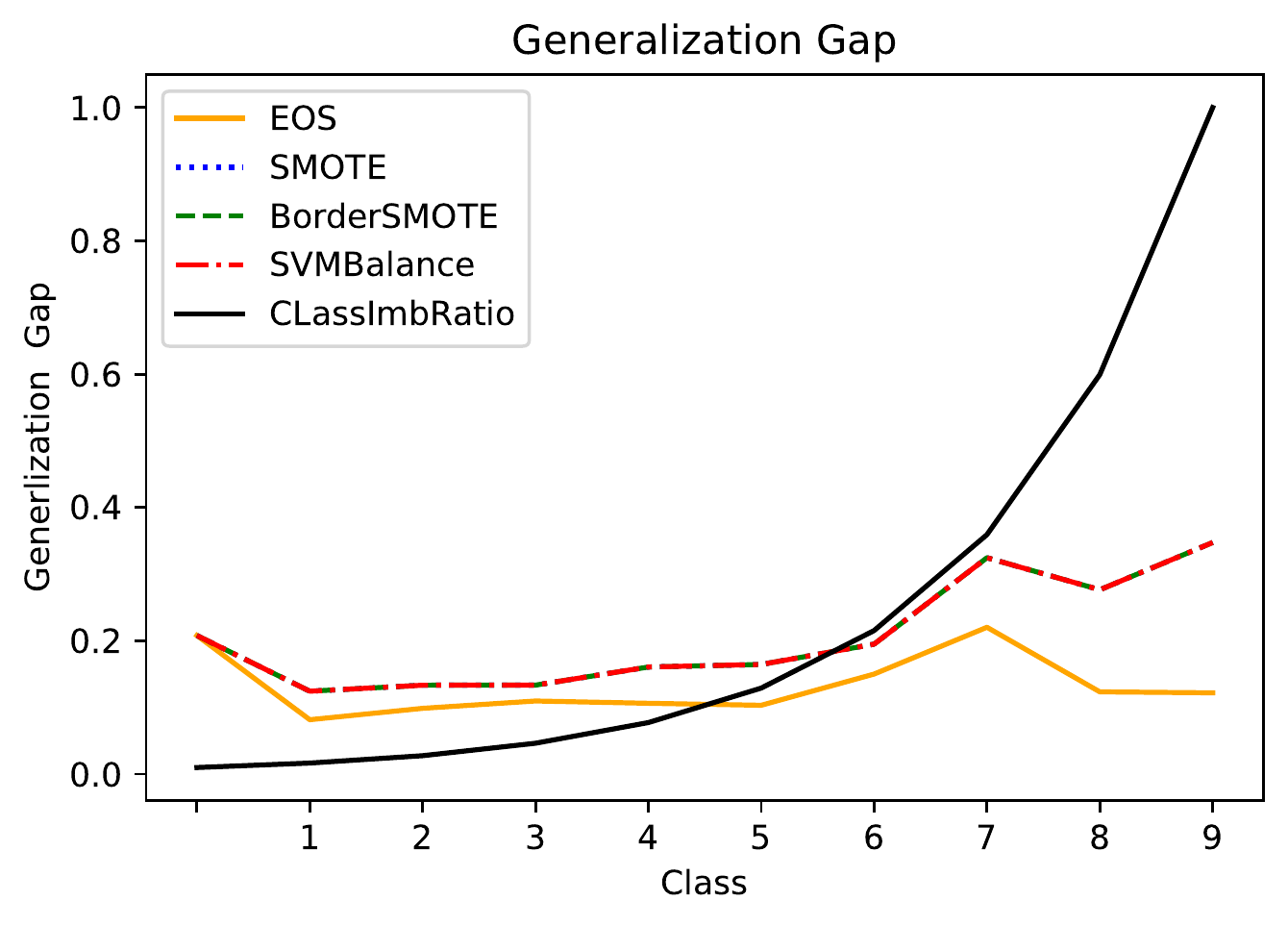}\label{fig:f35c}}
  \hfill
  \subfloat[SVHN:LDAM]{\includegraphics[width=0.2\textwidth]{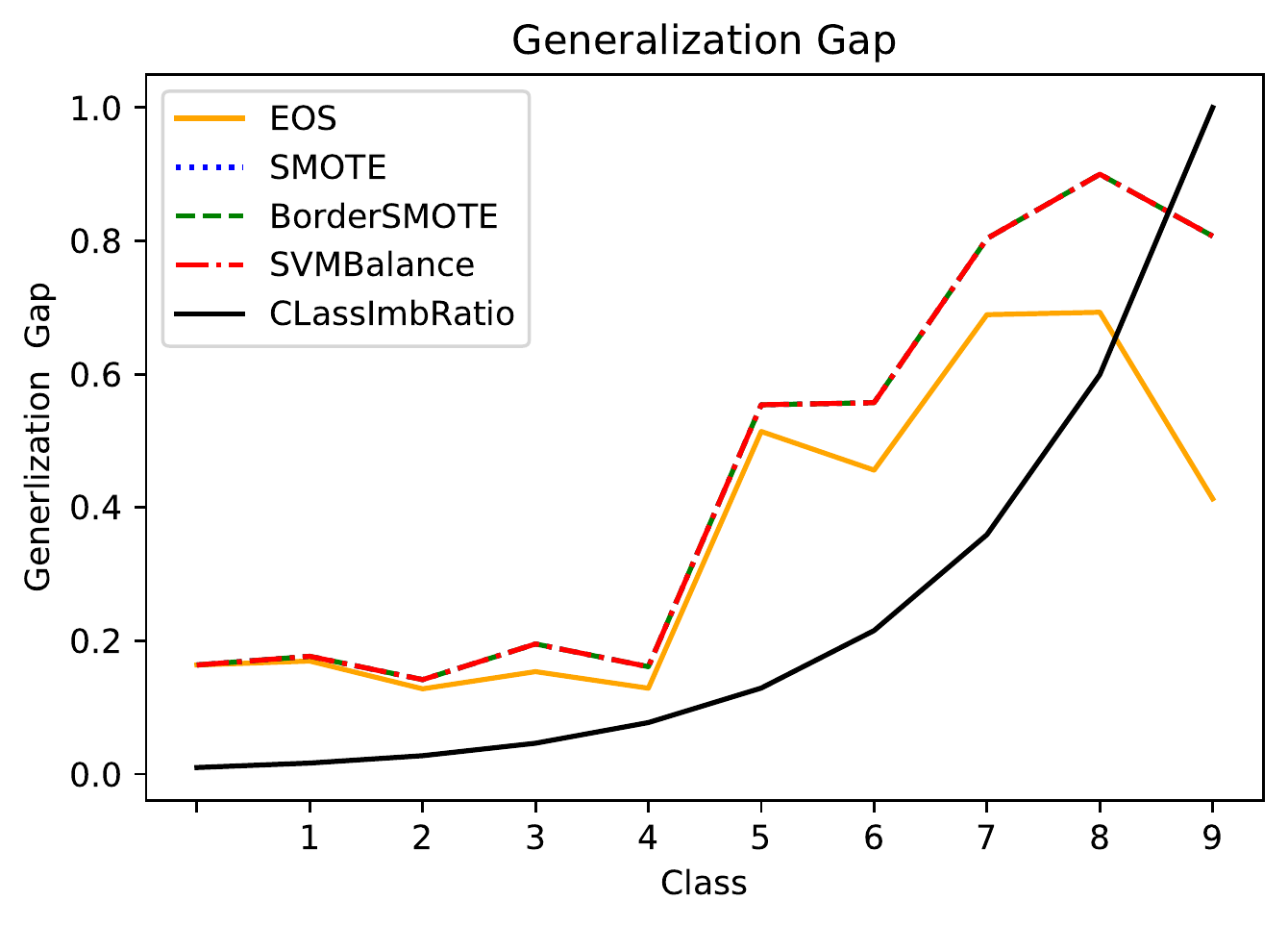}\label{fig:f35d}}
  \hfill
  \subfloat[CIFAR100:Base]{\includegraphics[width=0.2\textwidth]{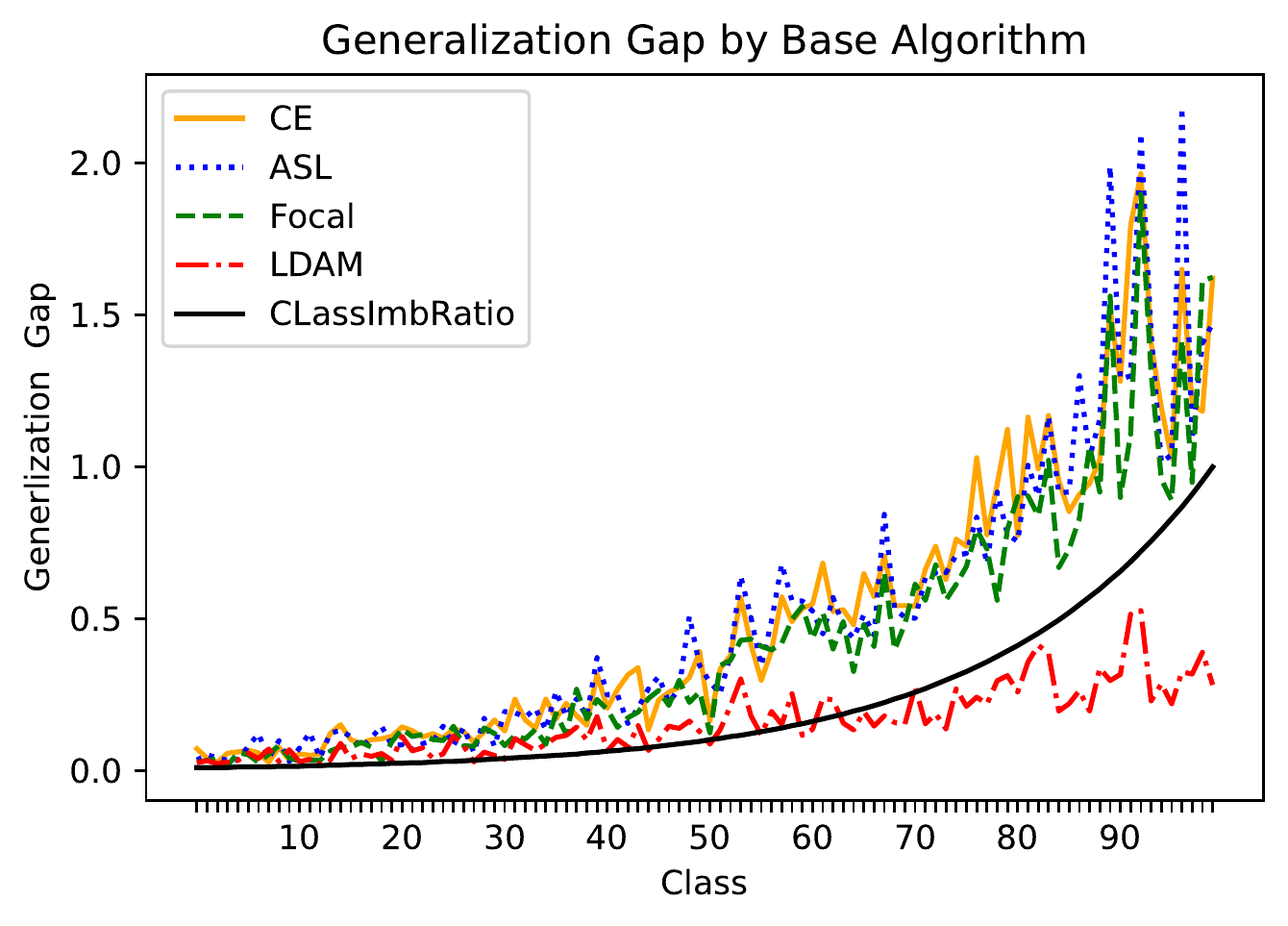}\label{fig:f35e}}
  \hfill
  \subfloat[CIFAR100:ASL]{\includegraphics[width=0.2\textwidth]{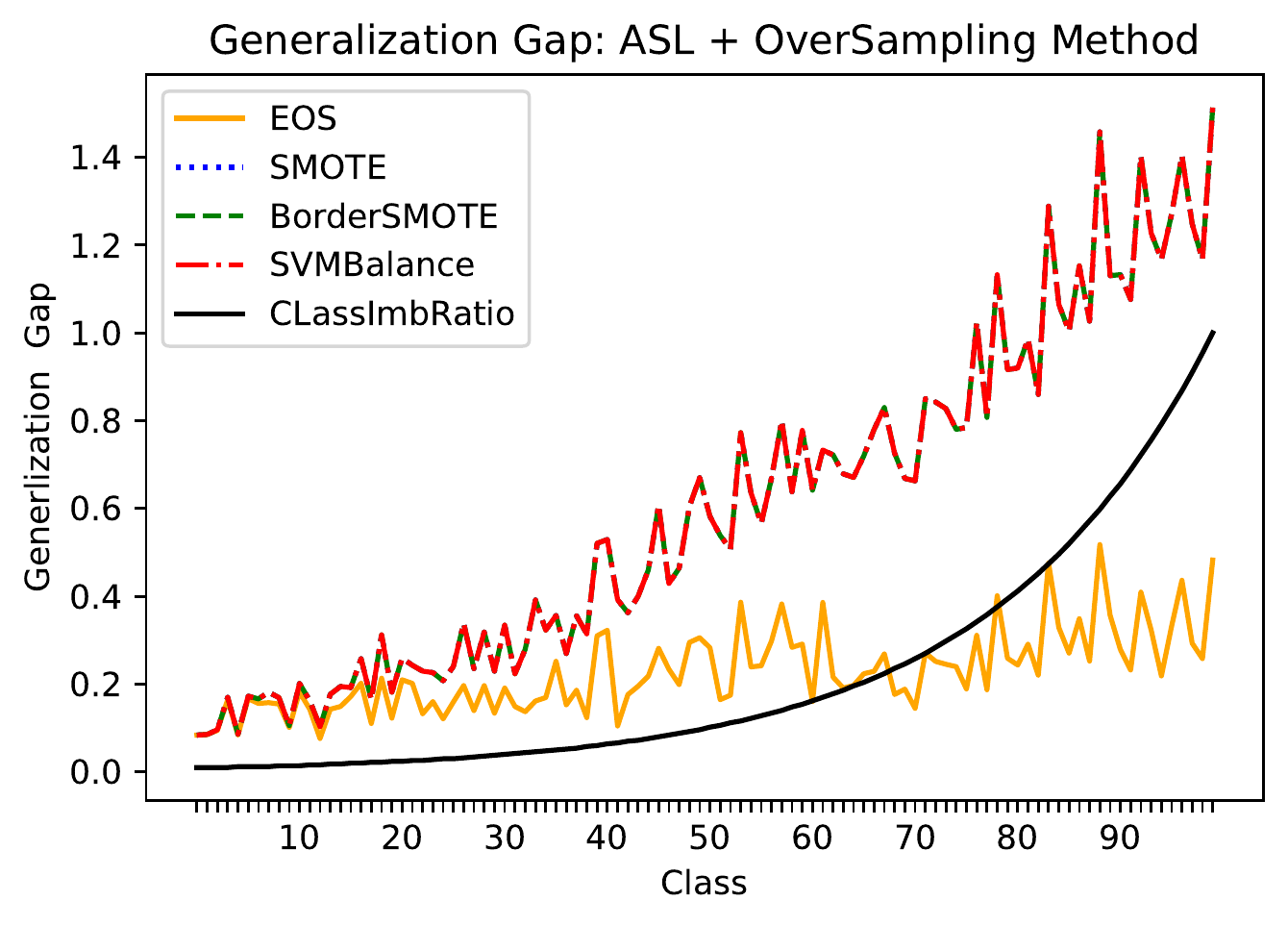}\label{fig:f35f}}
   \hfill
  \subfloat[CIFAR100:CE]{\includegraphics[width=0.2\textwidth]{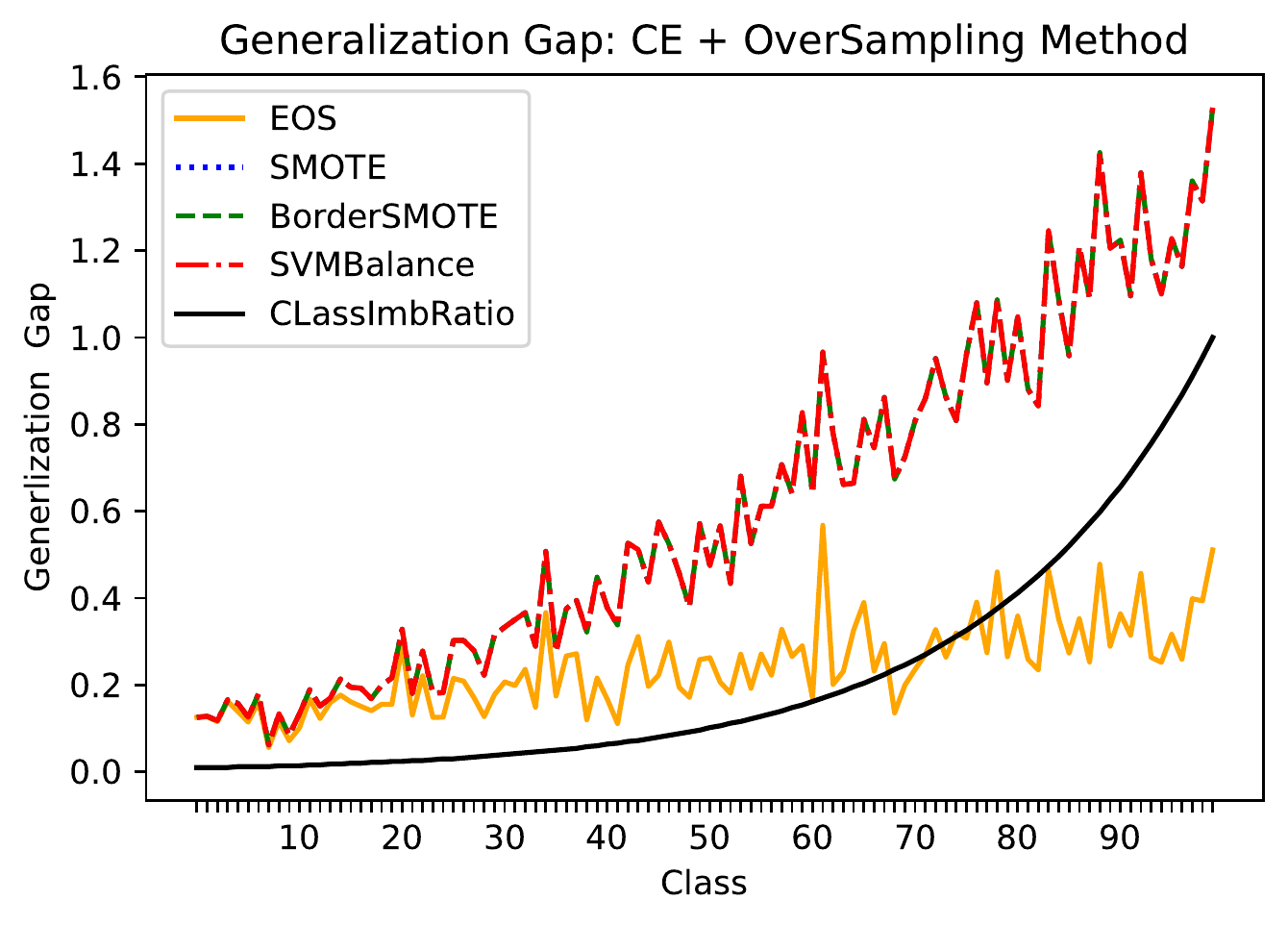}\label{fig:f35g}}
   \hfill
  \subfloat[CIFAR100:Focal]{\includegraphics[width=0.2\textwidth]{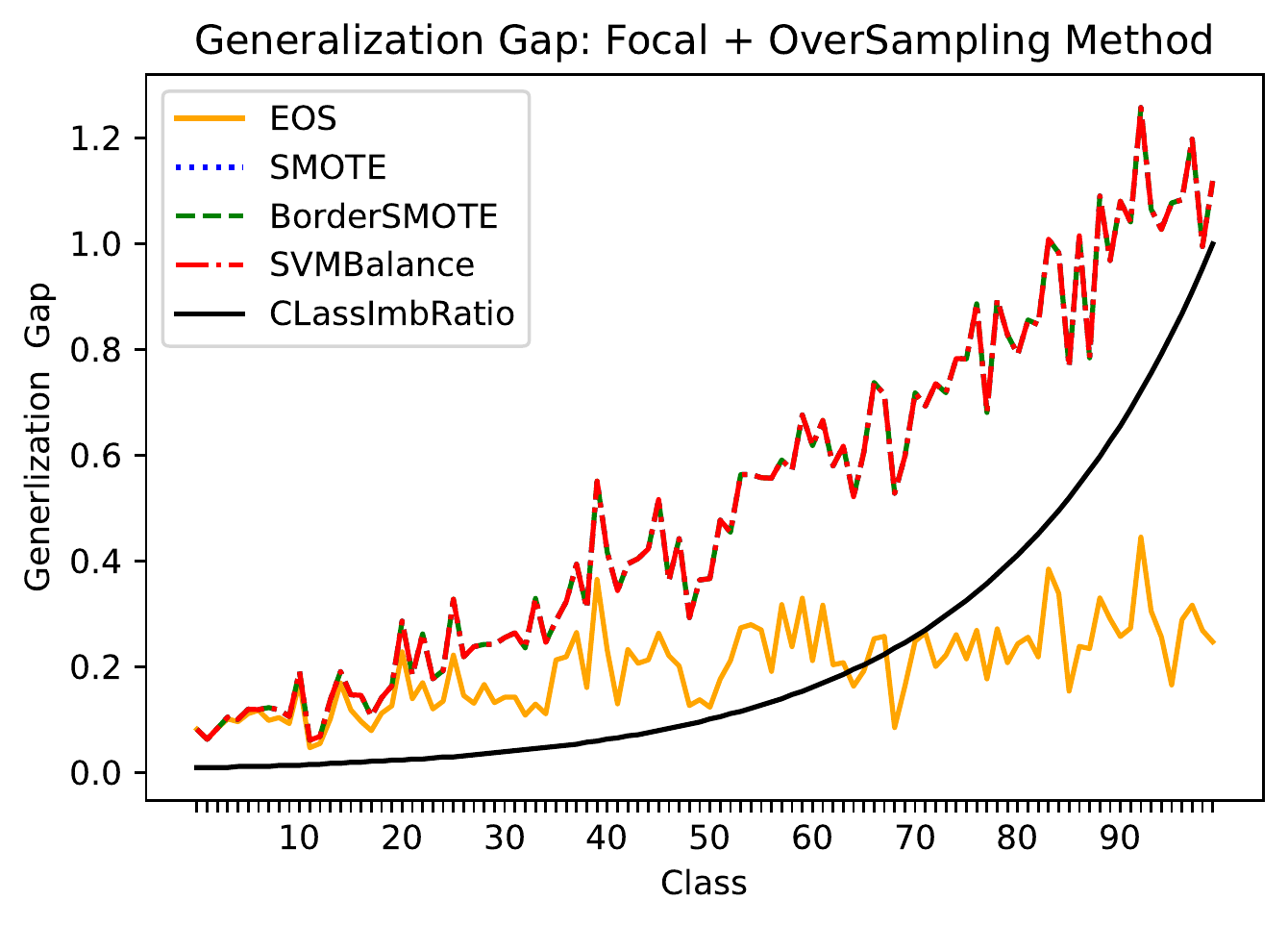}\label{fig:f35g}}
   \hfill
  \subfloat[CIFAR100:LDAM]{\includegraphics[width=0.2\textwidth]{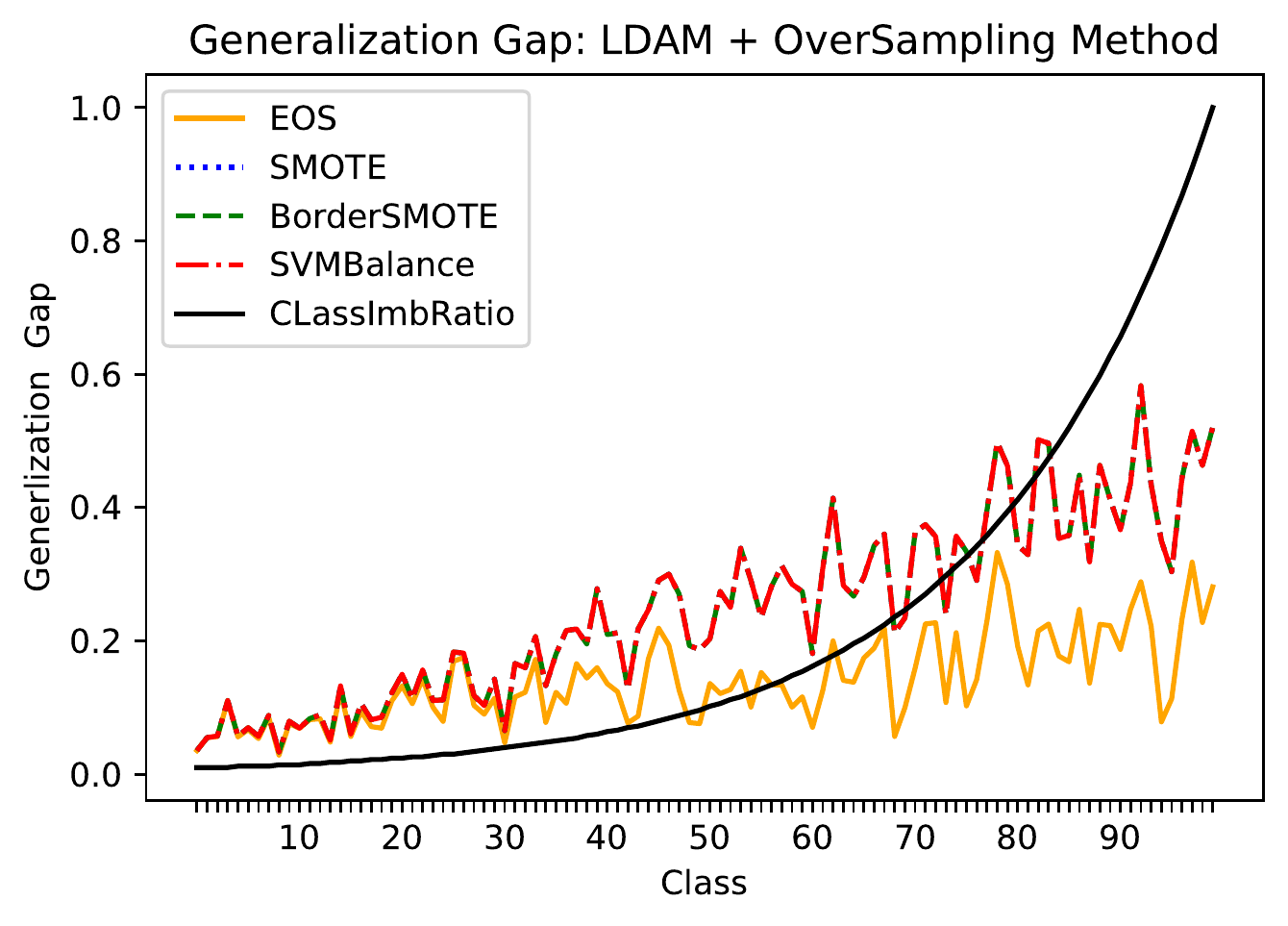}\label{fig:f35h}}
  
  \hfill
  \subfloat[CelebA:Base]{\includegraphics[width=0.2\textwidth]{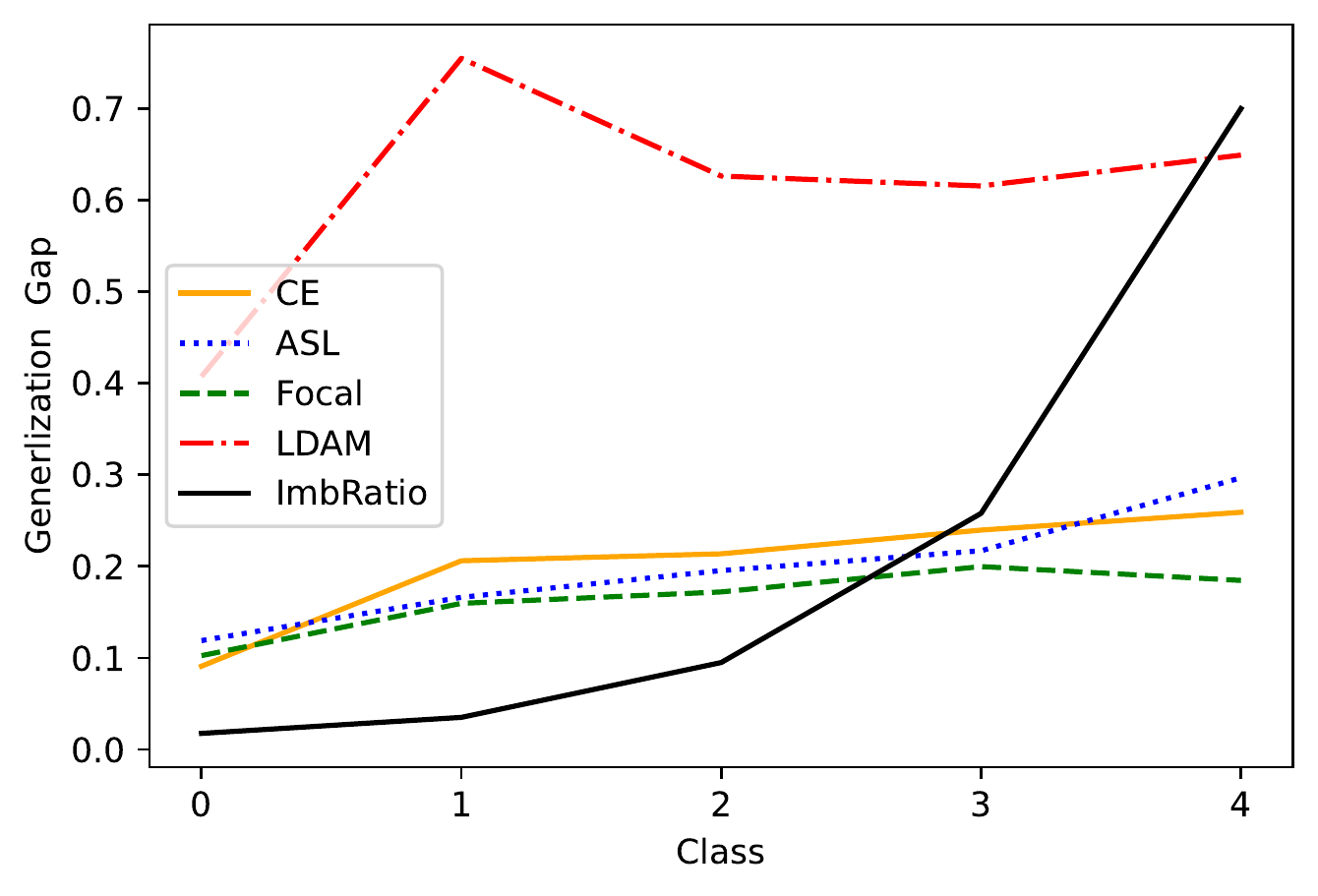}\label{fig:f35e1}}
  \hfill
  \subfloat[CelebA:ASL]{\includegraphics[width=0.2\textwidth]{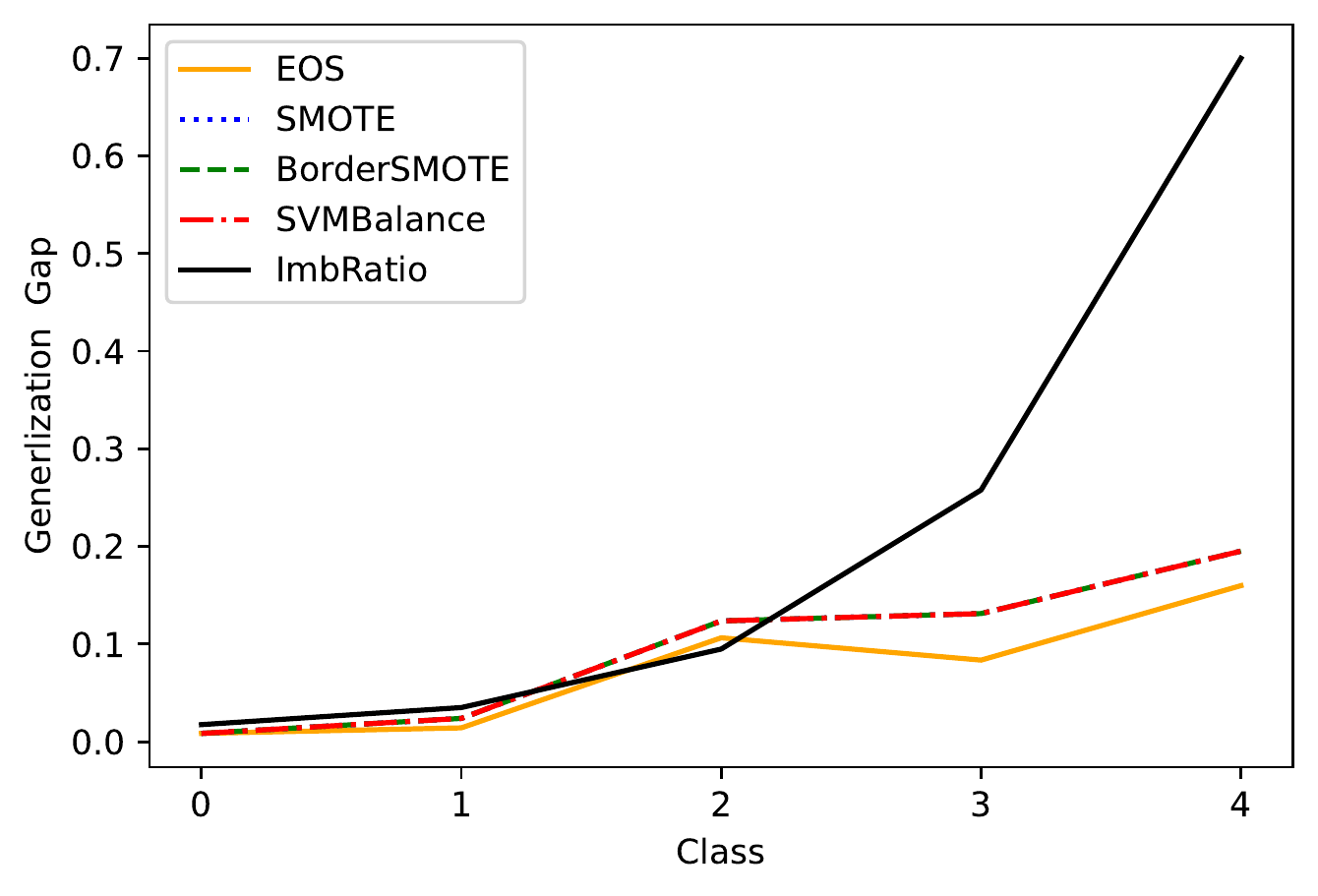}\label{fig:f35f1}}
   \hfill
  \subfloat[CelebA:CE]{\includegraphics[width=0.2\textwidth]{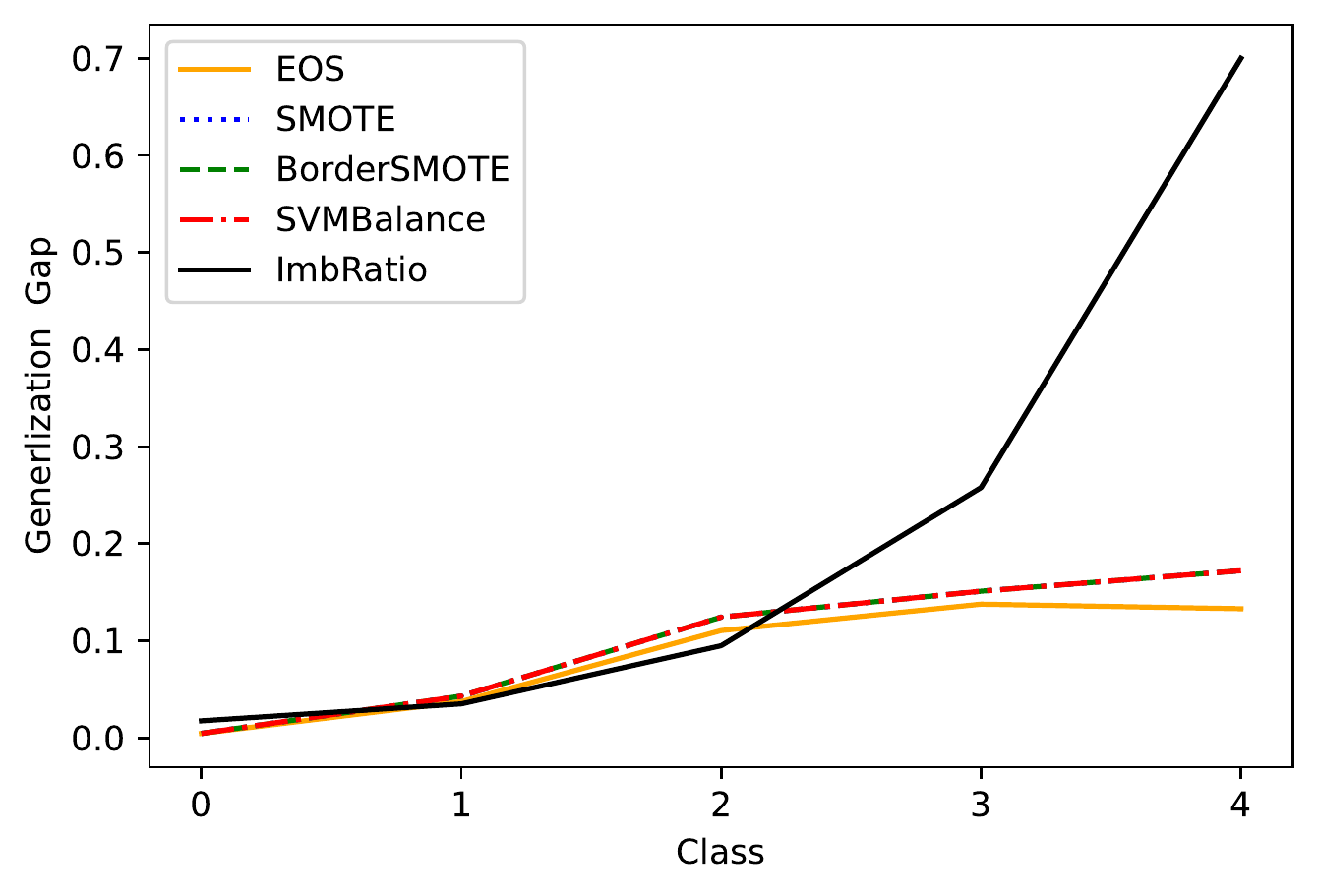}\label{fig:f35g1}}
   \hfill
  \subfloat[CelebA:Focal]{\includegraphics[width=0.2\textwidth]{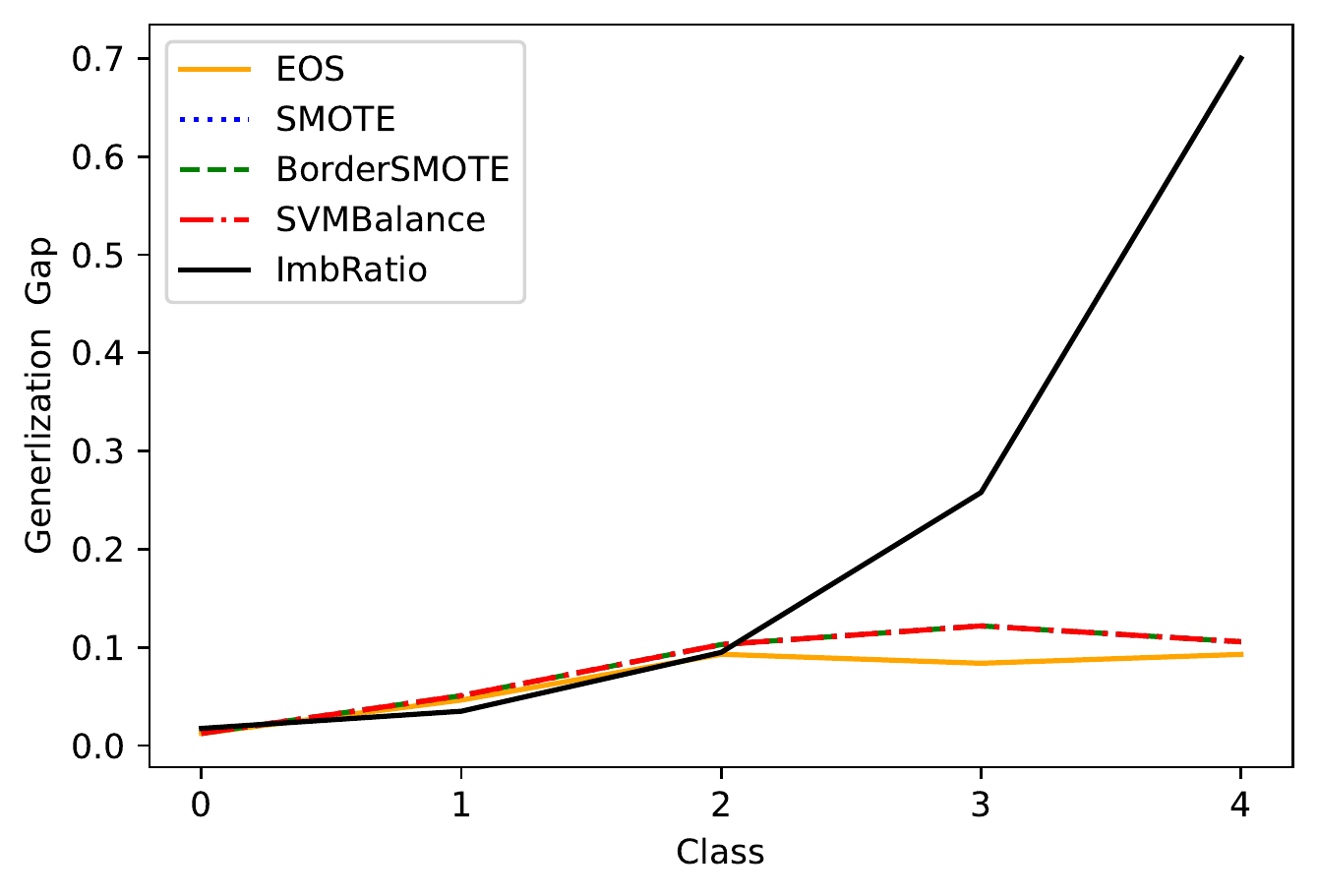}\label{fig:f35g1}}
   \hfill
  \subfloat[CelebA:LDAM]{\includegraphics[width=0.2\textwidth]{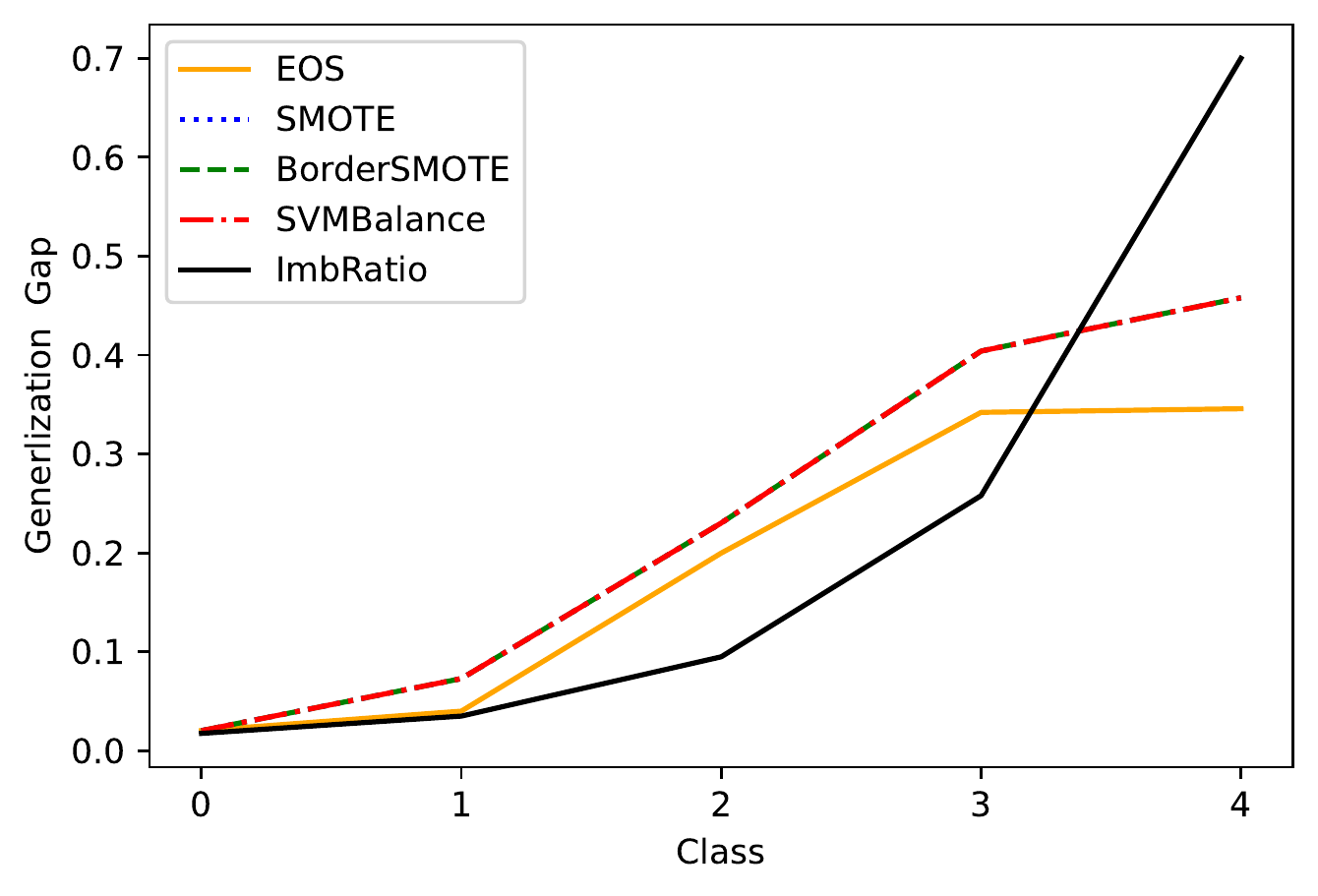}\label{fig:f35h1}}

  \caption{ These figures show that EOS reduces the generalization gap (y-axis) in feature embedding space for all four reference algorithms and classes (x-axis).  The black line represents the class imbalance ratio.  For most algorithms and datasets, the generalization gap approximately parallels the class imbalance ratio (black line); except in the case of EOS, which helps "flat-line" the generalization gap between classes in the face of imbalance.  EOS has less of an impact in the case of CelebA, which has a small overall difference between training and test accuracy.}
  \label{fig_gen_gap}
  \vspace{-0.2cm}
\end{figure*}

 In subfigures (a), (f), (k) and (p) of Figure~\ref{fig_gen_gap}, the generalization gaps of four cost-sensitive algorithms are shown. In all cases, the generalization gap increases as class imbalance increases. In Figure~\ref{fig_gen_gap}, the subplots feature a black line, which depicts the exponentially increasing class imbalance.  Although far from a perfect relationship, all of the baseline algorithms (ASL, CE, Focal loss, and LDAM) show a gradually rising generalization gap as class imbalance increases. Even after over-sampling is applied in the classification layer with SMOTE, Borderline SMOTE and Balanced SVM (depicted with an over-lapping red dotted line), the minority class generalization gap continues. 
 The plot lines for these methods overlap because they have not changed the range of feature embeddings.  Because these methods generate synthetic examples by interpolating between same class instances, the range of the training set minority class feature embeddings will remain the same.
 
 Only in the case of EOS (orange line), is the increase in the generalization gap somewhat arrested for minority classes. In summary, these diagrams confirm that the generalization gap is greater for minority classes, which is likely due to fewer examples in the training set, for all cost-sensitive algorithms and over-sampling methods, except in the case of EOS.

\begin{figure}[h!]
\vspace{-0.2cm}
  \centering
  \includegraphics[width=0.45\textwidth]{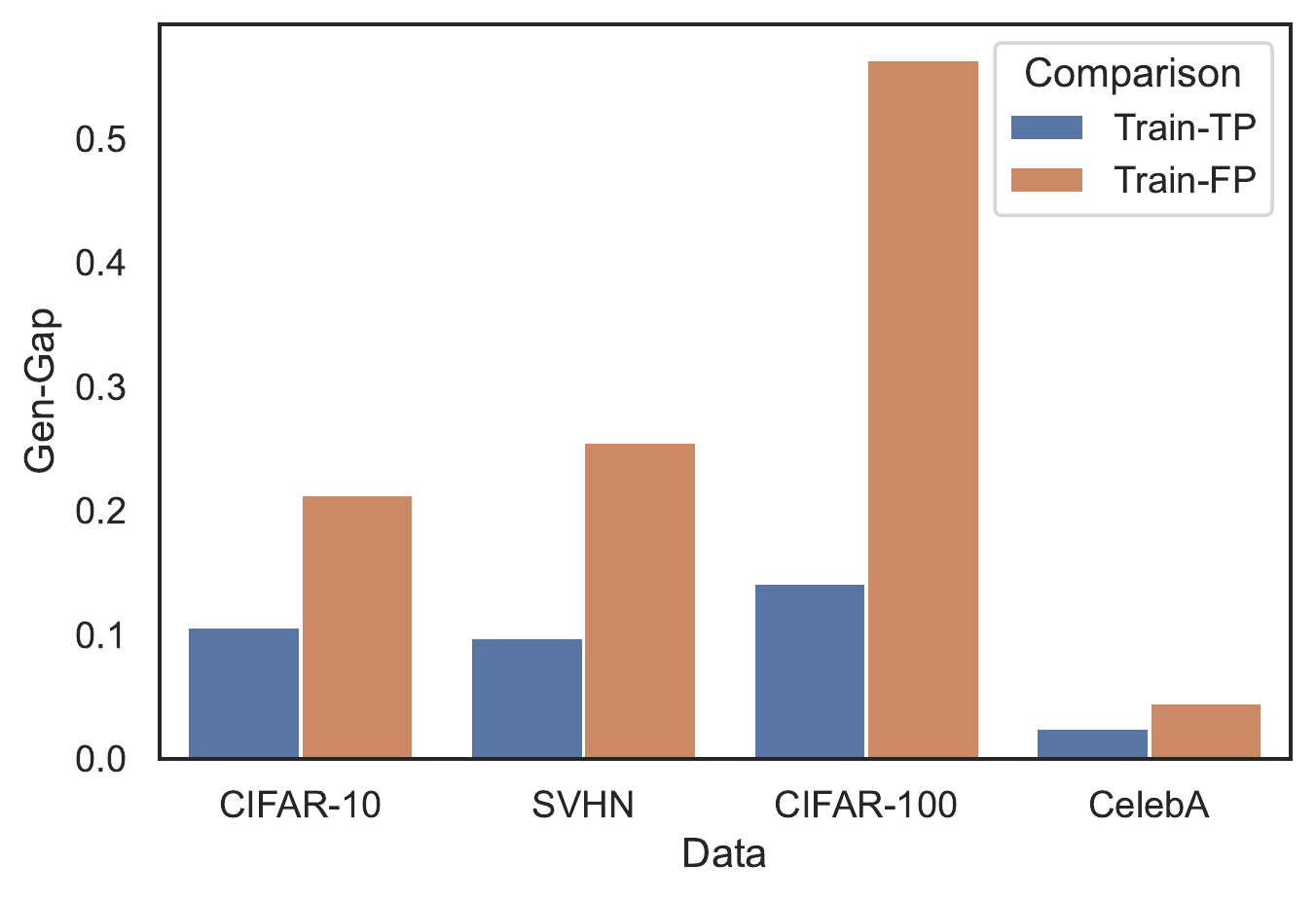}\label{fig:gg_tp_fp}
  
  \caption{ This figure shows that the generalization gap, as measured by the difference in the \textit{range} of learned feature embeddings from the training to the test sets, is higher for all datasets in the case of false positives (FPs) compared to true positives (TPs). This confirms that a model is better able to generalize (TPs) when the feature embedding \textit{ranges} between the training and test sets are more closely aligned.}
  \label{fig:gg_tp_fp}
  \vspace{-0.6cm}
\end{figure}

\begin{table*}[t!]
\footnotesize
\caption{Pre-Processing vs. Feature Embedding Space Over-Sampling (Cross-Entropy Loss)}
\label{tab: pre-res}
\centering
\begin{threeparttable}
\begin{tabular}{ p{1.8cm}p{.6cm}
p{.6cm}p{.6cm}p{.6cm}p{.6cm}
p{.6cm}p{.6cm}p{.6cm}p{.6cm}p{.6cm}p{.6cm}p{.6cm}}
\toprule

\multicolumn{1}{l}{\textbf{Descr}} & 
\multicolumn{3}{l}{\textbf{CIFAR-10}} & 
\multicolumn{3}{l}{\textbf{SVHN}} &
\multicolumn{3}{l}{\textbf{CIFAR-100}} &
\multicolumn{3}{l}{\textbf{CelebA}} 
\\
\midrule

Cross-Entropy & BAC & GM & FM & BAC & GM & FM &
BAC & GM & FM & BAC & GM & FM 
\\

\midrule


Pre-SMOTE & .7293 & .8410 & .7284 & .8539 & .9224 & .8638 & .5635 & .7490 & .5585 & .7392 & .8313 & .7300\\

Pre-BSMOTE & .7104 & ,8292 & .7054 & .8525 & .9157 & .8512 & .5620 & .7480 & .5539 & .7784 & .8575 & .7739\\

Pre-BalSVM & .7418 & ,8488 & .7421 & .8697 & .9258 & .8695 & .5637 & .7491 & .5583 & .7254 & .8219 & .7123\\

Remix[1] & .7331 & ,8434 & .7318 & .8393 & .9079 & .8380 & \textbf{.5769} & \textbf{.7579} & .5671 & .7286 & .8241 & .7233 \\
\midrule
Post-SMOTE & .7390 & .8471 & .7375 & \textbf{.8937} & \textbf{.9397} & \textbf{.8936} &
.5761 & .7574 & \textbf{.5731} & .7968 & .8697 & .7931\\

Post-BSMOTE & \textbf{.7446} & \textbf{.8506} & \textbf{.7428} &
.8238 & .8987 & .8237 &
.5722 & .7547 & .5673 
& .7916 & .8662 & .7878 \\
Post-BalSVM & .7381 & .8465 & .7367 & .8931 & .9394 & .8931 &
.5751 & .7567 & .5743 & \textbf{.7980} & \textbf{.8705} & \textbf{.7940}\\

\bottomrule

\end{tabular}
\begin{tablenotes}
\item[1] Remix balances class samples in pixel space; showing it in post-processing would represent double sample balancing.
\end{tablenotes}

\end{threeparttable}
\vspace{-.2cm}

\end{table*}

 In addition, Figure~\ref{fig:gg_tp_fp} shows that the generalization gap is larger for test set false positives (FP) than true positives (TP). True positives represent test set instances that a model correctly predicts based on information that it has learned during training. For all four datasets, a CNN is better able to generalize, in the form of TPs, from the training to the test set when there is a small difference in the learned feature embedding \textit{ranges}. In the case of FPs (i.e., when the model has difficulty generalizing and predicts the wrong label), the difference in the \textit{range} of learned feature embeddings is empirically shown in the figure to be 2X to almost 4X greater in the case of FPs, as measured by the generalization gap. For purposes of this comparison, two different CNN model architectures were used (a Resnet-32 and Resnet-56).  This provides further confirmation that the learned feature embedding ranges have a powerful impact on a CNN's ability to generalize  
 (\textbf{RQ1 answered}).    
\subsection{Over-sampling in embedding vs. pixel space}

Table~\ref{tab: pre-res} shows that in 7 out of 9 cases, the same over-sampling method used on feature embeddings outperformed the same method used in pre-processing.  Retraining the classifier in feature embedding space with over-sampling generally outperformed the same algorithm used as a pre-processing step in pixel space. This experiment confirms that three leading \textit{over-sampling} techniques work better in FE space than pixel space.  (\textbf{RQ2 answered}).   

\subsection{Analyzing EOS performance}

\begin{table*}[b!]
\centering
\footnotesize
\caption{Baseline Algorithms \& Over-Sampling Accuracy Results}
\label{tab: results}
\begin{tabular}{ p{.8cm}p{.6cm}p{.6cm}
p{.6cm}p{.6cm}p{.6cm}p{.6cm}
p{.6cm}p{.6cm}p{.6cm}p{.6cm}
p{.6cm}p{.6cm}p{.6cm}p{.6cm}
p{.6cm}}
\toprule

\multicolumn{1}{l}{\textbf{Descr}} & 
\multicolumn{3}{l}{\textbf{Baseline}} & 
\multicolumn{3}{l}{\textbf{SMOTE}} &
\multicolumn{3}{l}{\textbf{B.SMOTE}} &
\multicolumn{3}{l}{\textbf{Bal. SVM}} &
\multicolumn{3}{l}{\textbf{EOS}}
\\
\midrule

Algo & BAC & GM & FM & BAC & GM & FM &
BAC & GM & FM & BAC & GM & FM &
BAC & GM & FM  \\

\midrule

\multicolumn{16}{l}{\textbf{CIFAR 10}}\\
\midrule

CE & .7256 & .8387 & .7246 &
.7390 & .8471 & .7375 &
.7446 & .8506 & .7428 &
.7381 & .8465 & .7367 &
\textbf{.7581} & \textbf{.8589} & \textbf{.7571} \\

ASL & .7100 & .8254 & .6997 &
.7465 & .8517 & .7454 &
.7441 & .8503 & .7422 &
.7481 & .8527 & .7474 &
\textbf{.7825} & \textbf{.8738} & \textbf{.7827}\\

Focal & .7020 & .8239 & .6969 &
.7438 & .8501 & .7426 &
.7503 & .8541 & .7502 &
.7430 & .8496 & .7422 &
\textbf{.7831} & \textbf{.8742} & \textbf{.7830}\\

LDAM & .7780 & .8711 & .7779 &
.7668 & .8643 & .7662 &
.7630 & .8619 & .7617 &
.7672 & .8645 & .7670 &
\textbf{.7865} & \textbf{.8763} & \textbf{.7862}\\

\midrule

\multicolumn{16}{l}{\textbf{SVHN}}\\
\midrule

CE & .8811 & .9324 & .8815 &
.8937 & .9397 & .8936 &
.8238 & .8987 & .8237 &
.8931 & .9394 & .8931 &
\textbf{.9016} & \textbf{.9443} & \textbf{.9014} \\

ASL & .8726 & .9275 & .8728 &
.8978 & .9421 & .8976 &
.8740 & .9283 & .8737 &
.8991 & .9429 & .8988 &
\textbf{.9005} & \textbf{.9437} & \textbf{.9003}\\

Focal & .8727 & .9275 & .8726 &
.8835 & .9339 & .8835 &
.8574 & .9186 & .8564 &
.8831 & .9336 & .8831 &
\textbf{.8913} & \textbf{.9384} & \textbf{.8909}\\

LDAM & .8961 & .9411 & .8959 &
.9030 & .9451 & .9029 &
.8933 & .9395 & .8930 &
.8992 & .9429 & .8992 &
\textbf{.9093} & \textbf{.9487} & \textbf{.9092}\\

\midrule

\multicolumn{16}{l}{\textbf{CIFAR 100}}\\

\midrule
CE & .5694 & .7529 & .5648 &
.5761 & .7574 & .5731 &
.5722 & .7548 & .5673 &
.5751 & .7567 & .5743 &
\textbf{.5794} & \textbf{.7596} & \textbf{.5786} \\

ASL & .5646 & .7497 & .5485 &
.5690 & .7527 & .5674 &
.5643 & .7495 & .5580 &
.5687 & .7525 & .5655 &
\textbf{.5722} & \textbf{.7548} & \textbf{.5727}\\

Focal & .5536 & .7423 & .5485 &
.5602 & .7468 & .5571 &
.5561 & .7440 & .5515 &
.5554 & .7436 & .5510 &
\textbf{.5633} & \textbf{.7489} & \textbf{.5623}\\

LDAM & \textbf{.5808} & \textbf{.7605} & \textbf{.5707} &
.5728 & .7552 & .5636 &
.5713 & .7542 & .5572 &
.5693 & .7529 & .5588 &
.5732 & .7555 & .5624\\

\midrule

\multicolumn{16}{l}{\textbf{CelebA}}\\

\midrule
CE & .7598 & .8451 & .7542 &
.7968 & .8697 & .7931 &
.7916 & .8662 & .7878 &
.7980 & .8705 & .7940 &
\textbf{.8044} & \textbf{.8747} & \textbf{.8023} \\

ASL & .7630 & .8472 & .7597 &
.7790 & .8579 & .7751 &
\textbf{.7846} & \textbf{.8616} & \textbf{.7828} &
.7738 & .8544 & .7696 &
.7758 & .8558 & .7729\\

Focal & .7458 & .8357 & .7382 &
.8000 & .8718 & .7970 &
.7996 & .8715 & .7936 &
.7996 & .8715 & .7967 &
\textbf{.8010} & \textbf{.8724} & \textbf{.7990}\\

LDAM & .7444 & .8348 & .7300 &
.8030 & .8738 & .8008 &
.7948 & .8683 & .7923 &
.7978 & .8703 & .7953 &
\textbf{.8064} & \textbf{.8760} & \textbf{.8055}\\

\bottomrule

\end{tabular}
\vspace{.1cm}
\end{table*}

As can be seen in Table~\ref{tab: results}, EOS outperforms other over-sampling methods, even when the other over-sampling methods are implemented in embedding space, and consistently improves classifier accuracy when compared to baselines.  Only in the case of CIFAR-100 did a baseline cost-sensitive algorithm outperform EOS. Table ~\ref{tab: results} also shows that EOS can be combined with cost-sensitive algorithms, improving their performance, and also introduce data augmentations efficiently at the back-end of training in a low-cost manner. 

We should further notice that the quality of the feature embedding extracted by the CNN has a large impact on overall performance.  With the exception of CIFAR-100, EOS performed best when it was combined with FE extracted by a CNN trained with the LDAM cost-sensitive algorithm.  This confirms our hypothesis that the loss function used to train the extraction layers has a significant impact on classifier accuracy, even when the classifier is separated from the extraction layers and retrained on cross-entropy with augmented data.

In order to investigate the reason for EOS' performance, we considered the impact of the various cost-sensitive algorithms and over-sampling methods on classifier weight norms.   
Over-sampling methods like SMOTE, Border-line SMOTE and Balanced SVM work by balancing the number of class examples. In theory, this approach  should cause classifier parameters to be more uniform, such that the norms for each class are more equally weighted.

\begin{figure*}[!t]
   \vspace{-0.5cm}
  \centering
  \subfloat[CIFAR10:Base]{\includegraphics[width=0.2\textwidth]{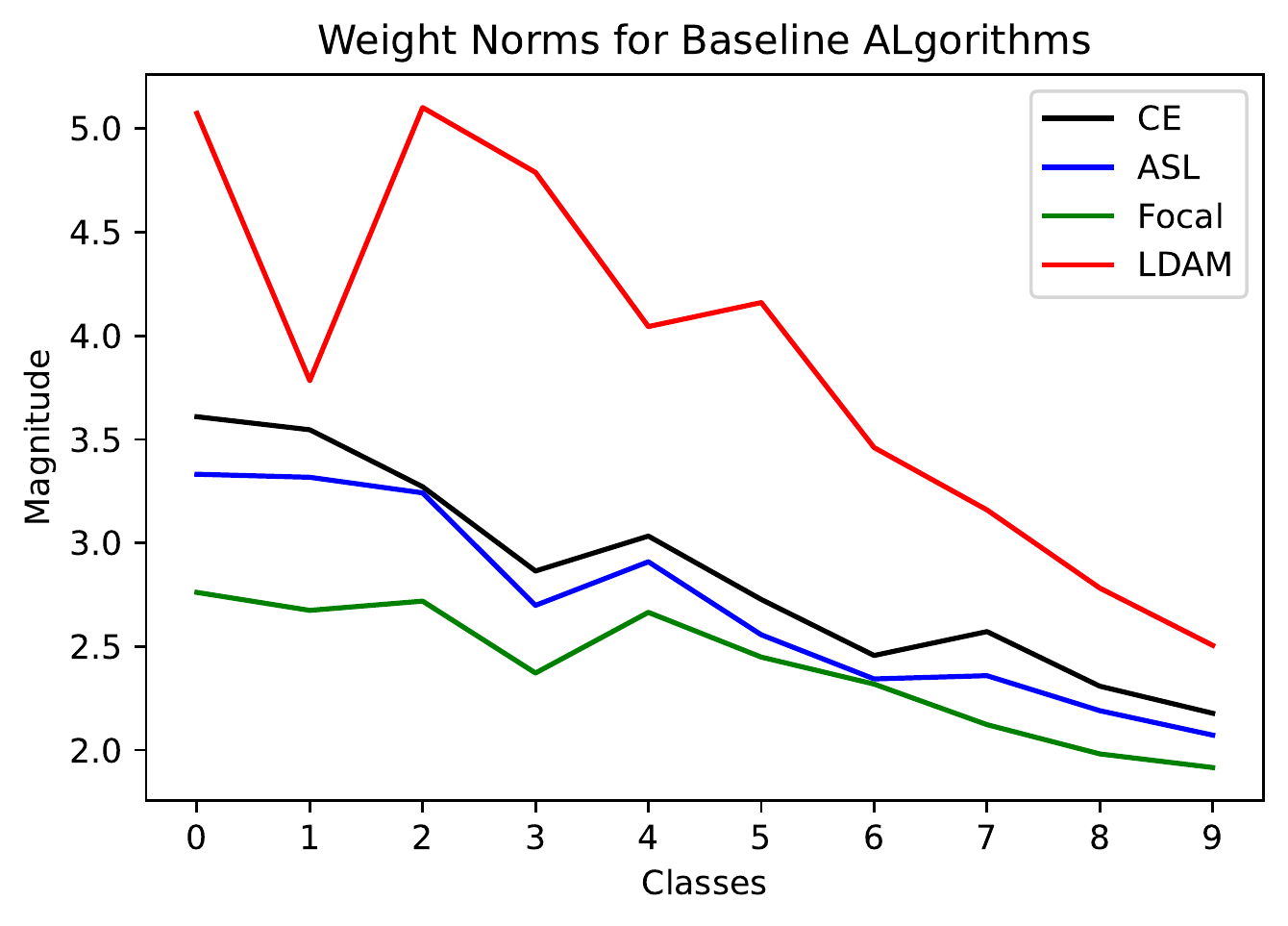}\label{fig:f6}}
  \hfill
  \subfloat[CIFAR10:CE]{\includegraphics[width=0.2\textwidth]{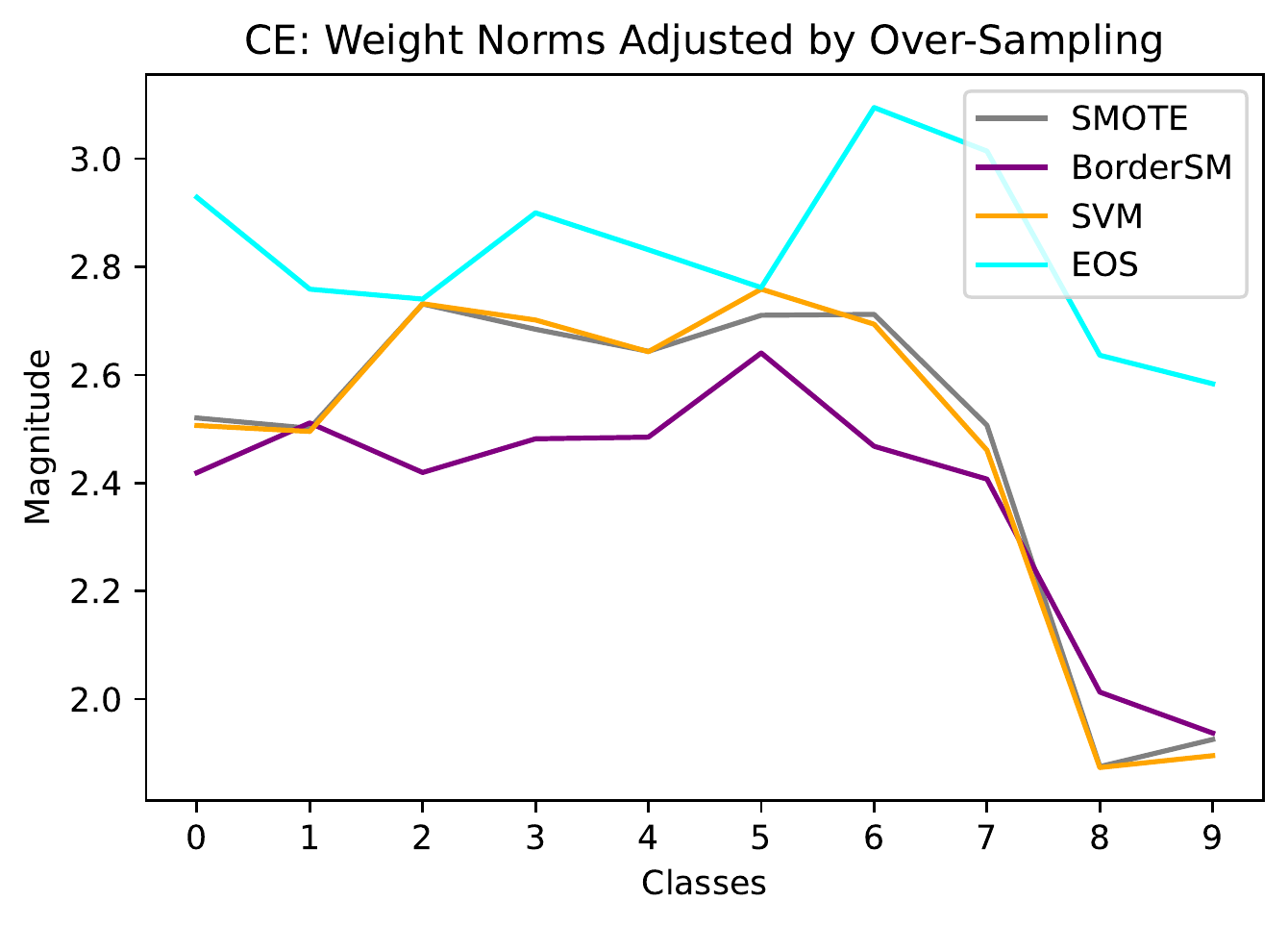}\label{fig:f7}}
   \hfill
  \subfloat[CIFAR10:ASL]{\includegraphics[width=0.2\textwidth]{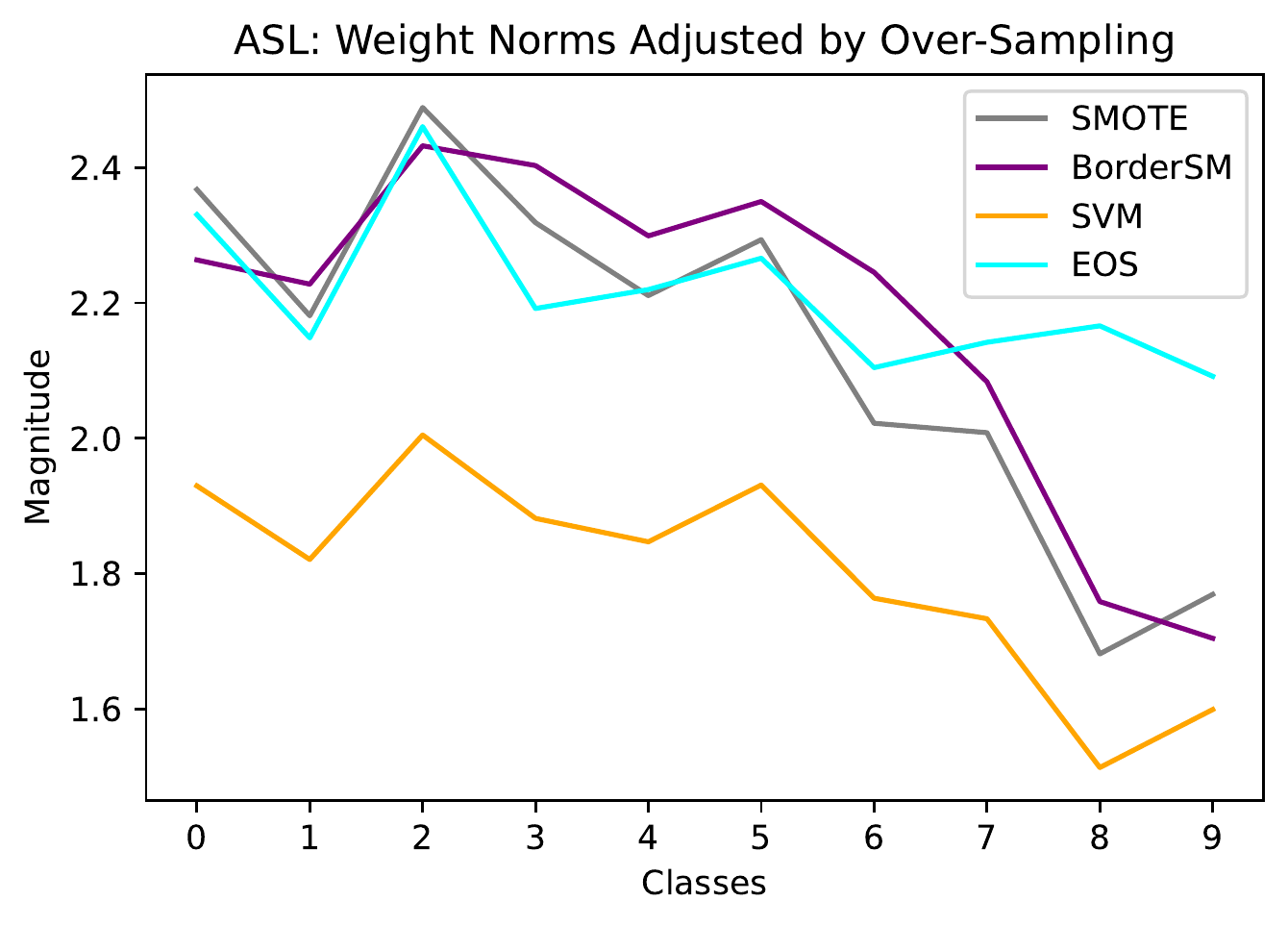}\label{fig:f8}}
   \hfill
  \subfloat[CIFAR10:Focal]{\includegraphics[width=0.2\textwidth]{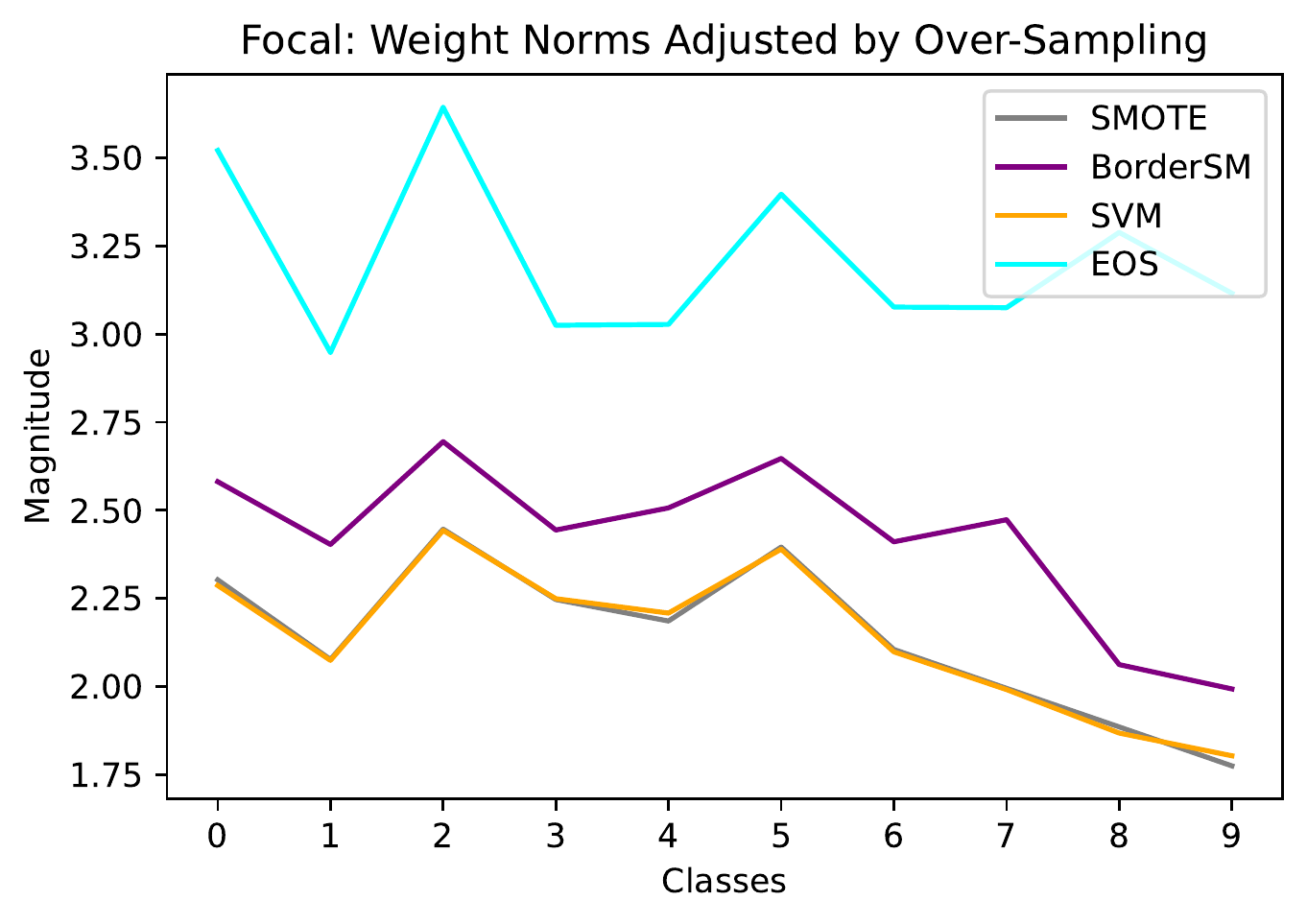}\label{fig:f9}}
  \hfill
  \subfloat[CIFAR10:LDAM]{\includegraphics[width=0.2\textwidth]{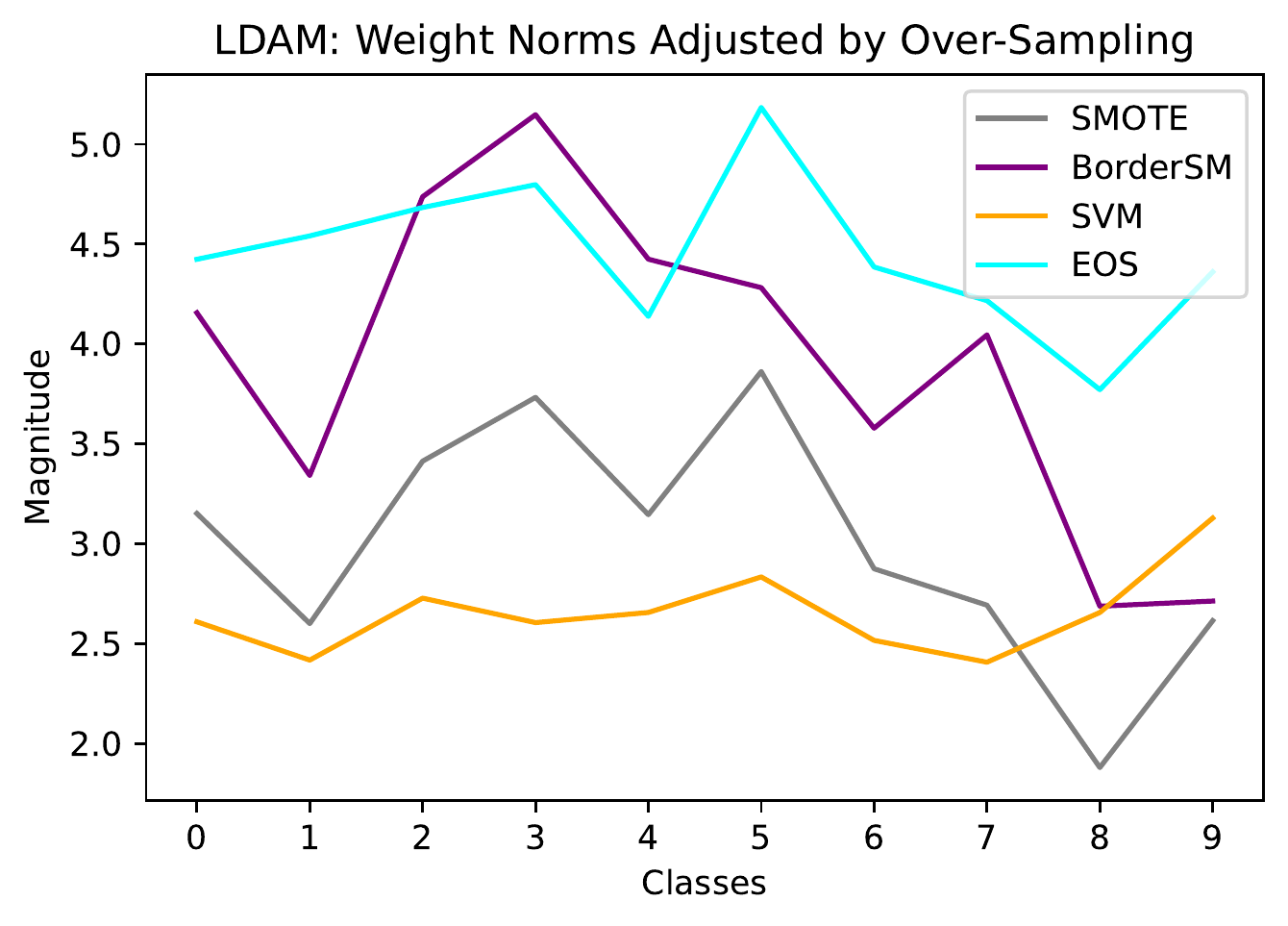}\label{fig:f10}}
  \hfill
  \subfloat[SVHN:Base]{\includegraphics[width=0.2\textwidth]{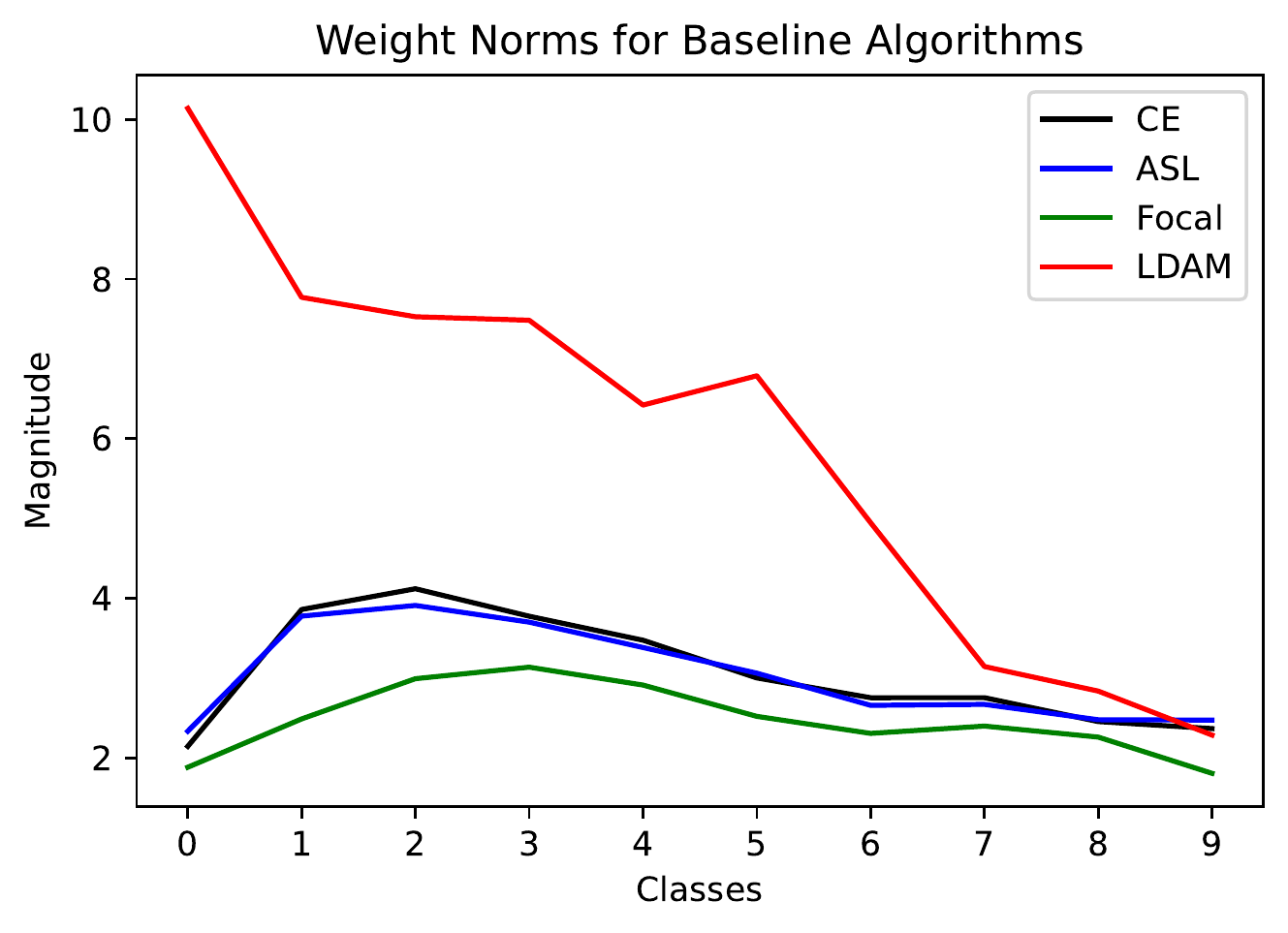}\label{fig:f10a}}
   \hfill
  \subfloat[SVHN:CE]{\includegraphics[width=0.2\textwidth]{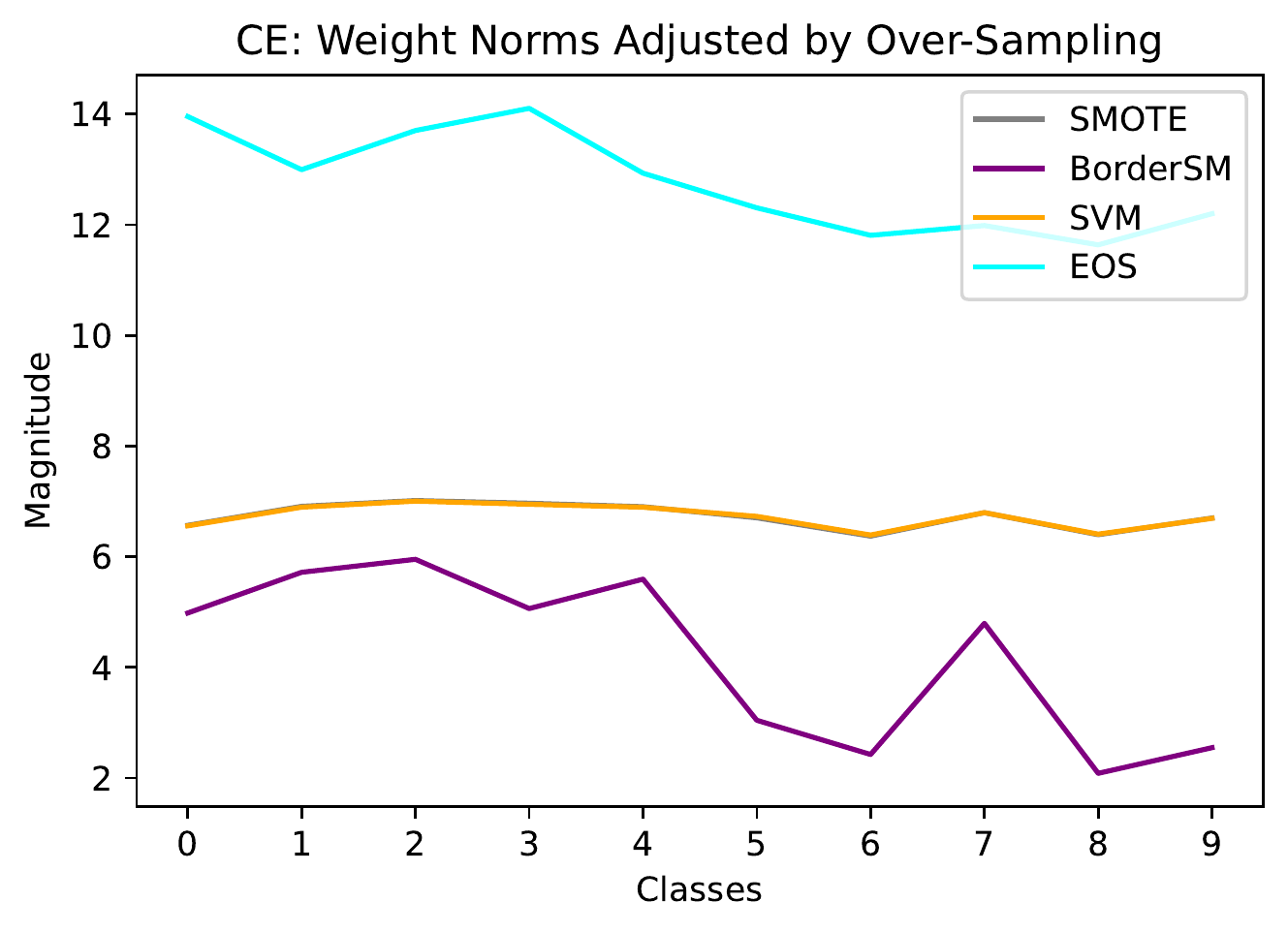}\label{fig:f10a}}
  \hfill
  \subfloat[SVHN:ASL]{\includegraphics[width=0.2\textwidth]{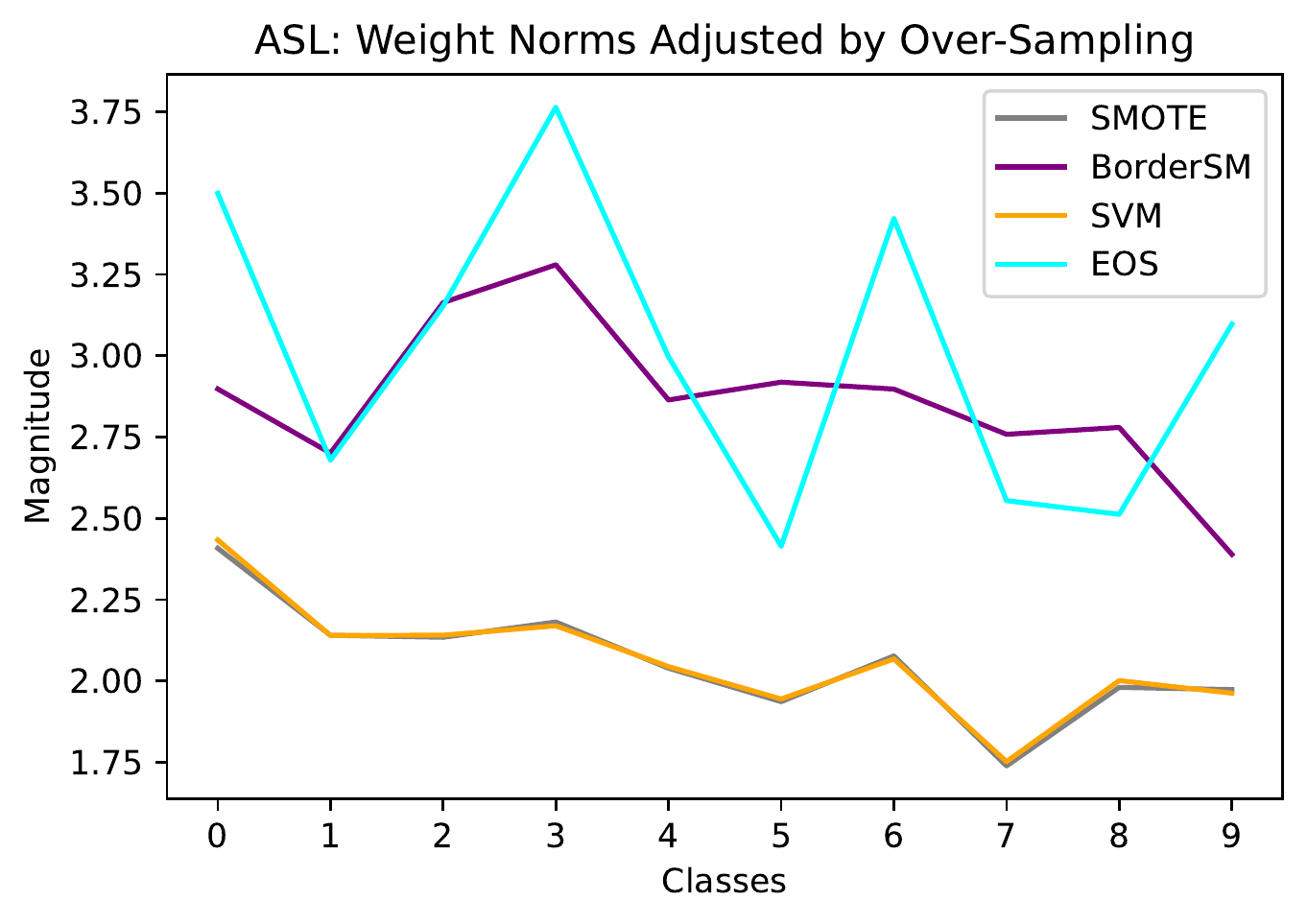}\label{fig:f10b}}
  \hfill
  \subfloat[SVHN:Focal]{\includegraphics[width=0.2\textwidth]{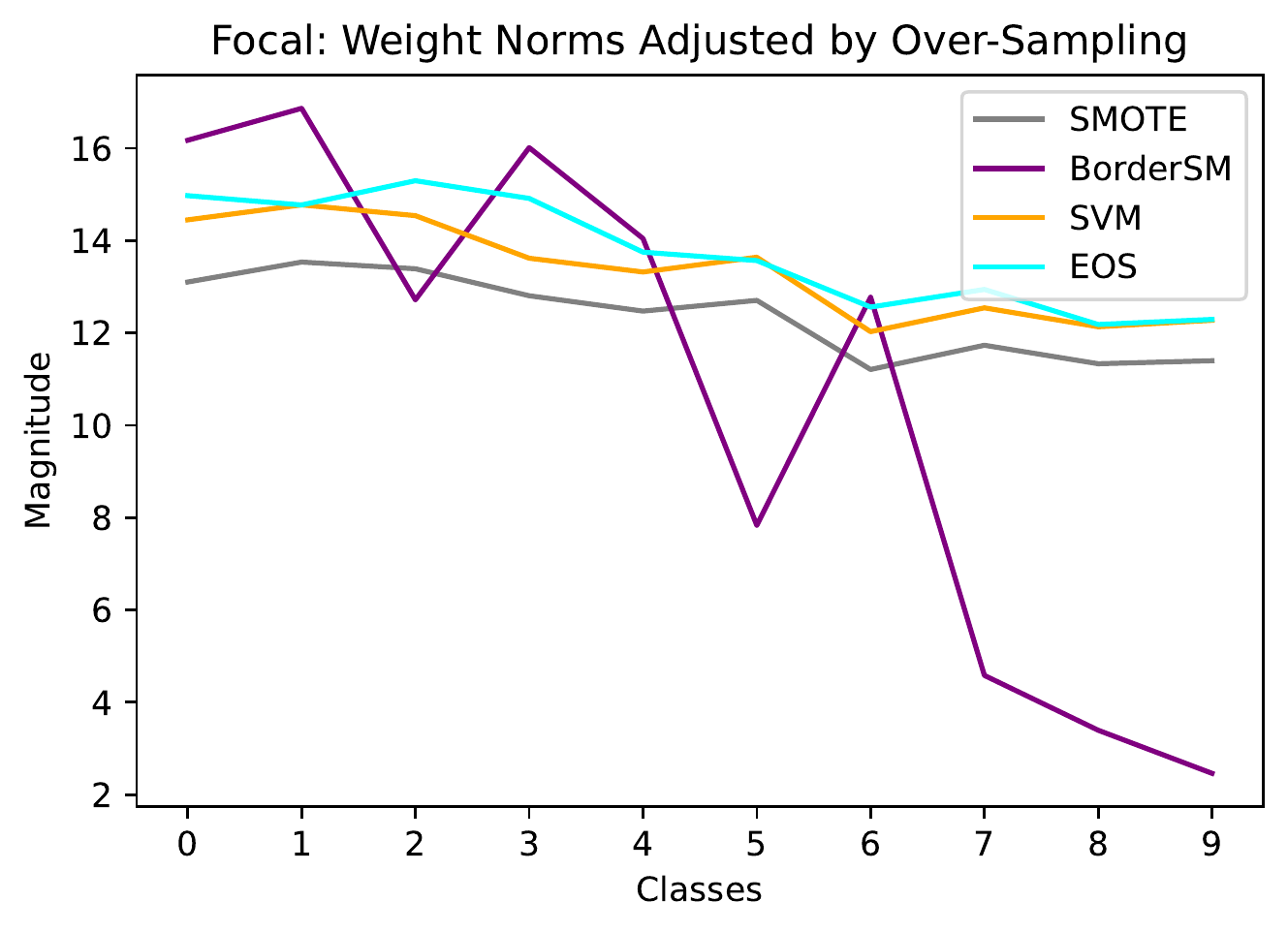}\label{fig:f10c}}
  \hfill
  \subfloat[SVHN:LDAM]{\includegraphics[width=0.2\textwidth]{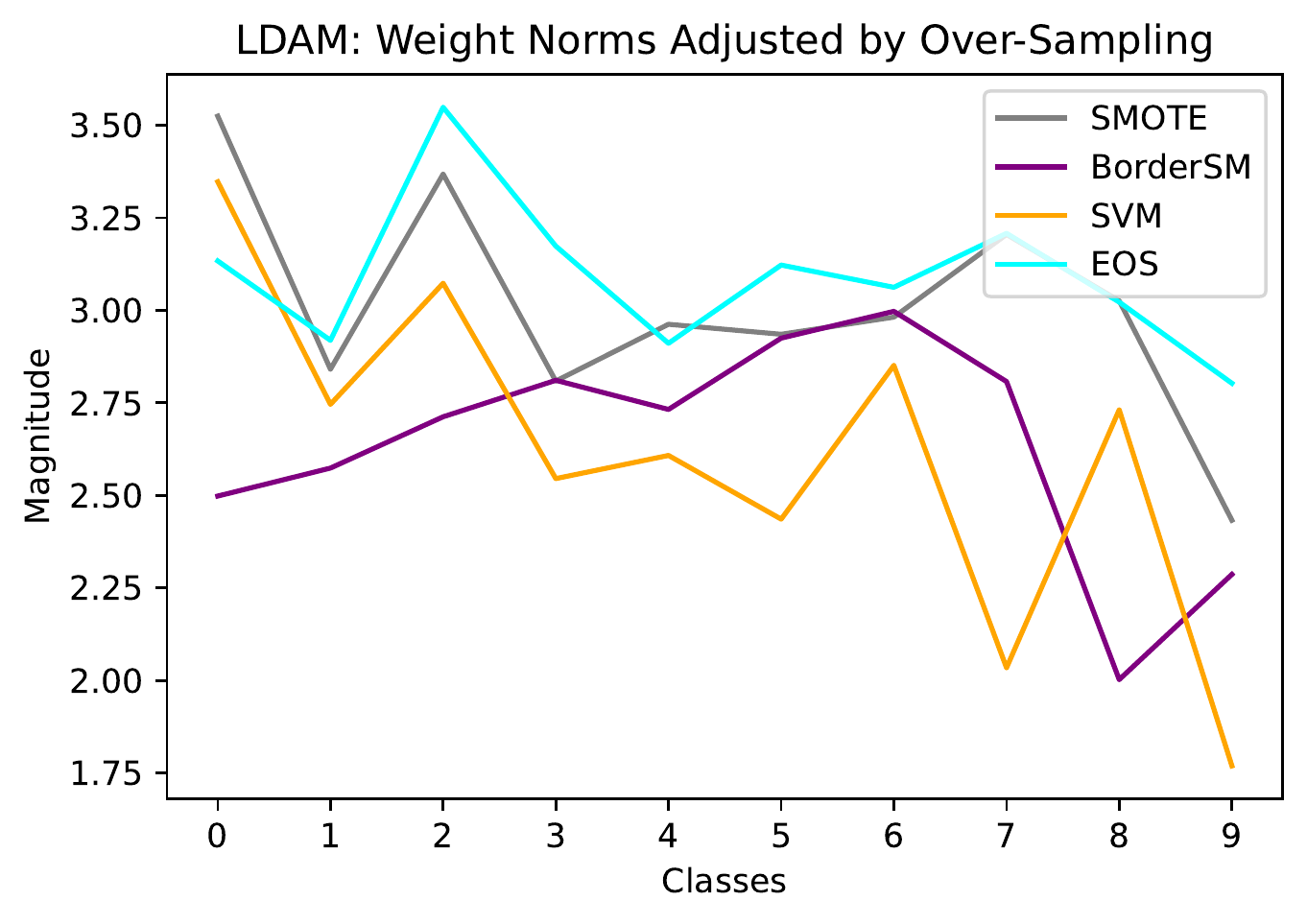}\label{fig:f10d}}
   \hfill
  \subfloat[CIFAR100:Base]{\includegraphics[width=0.2\textwidth]{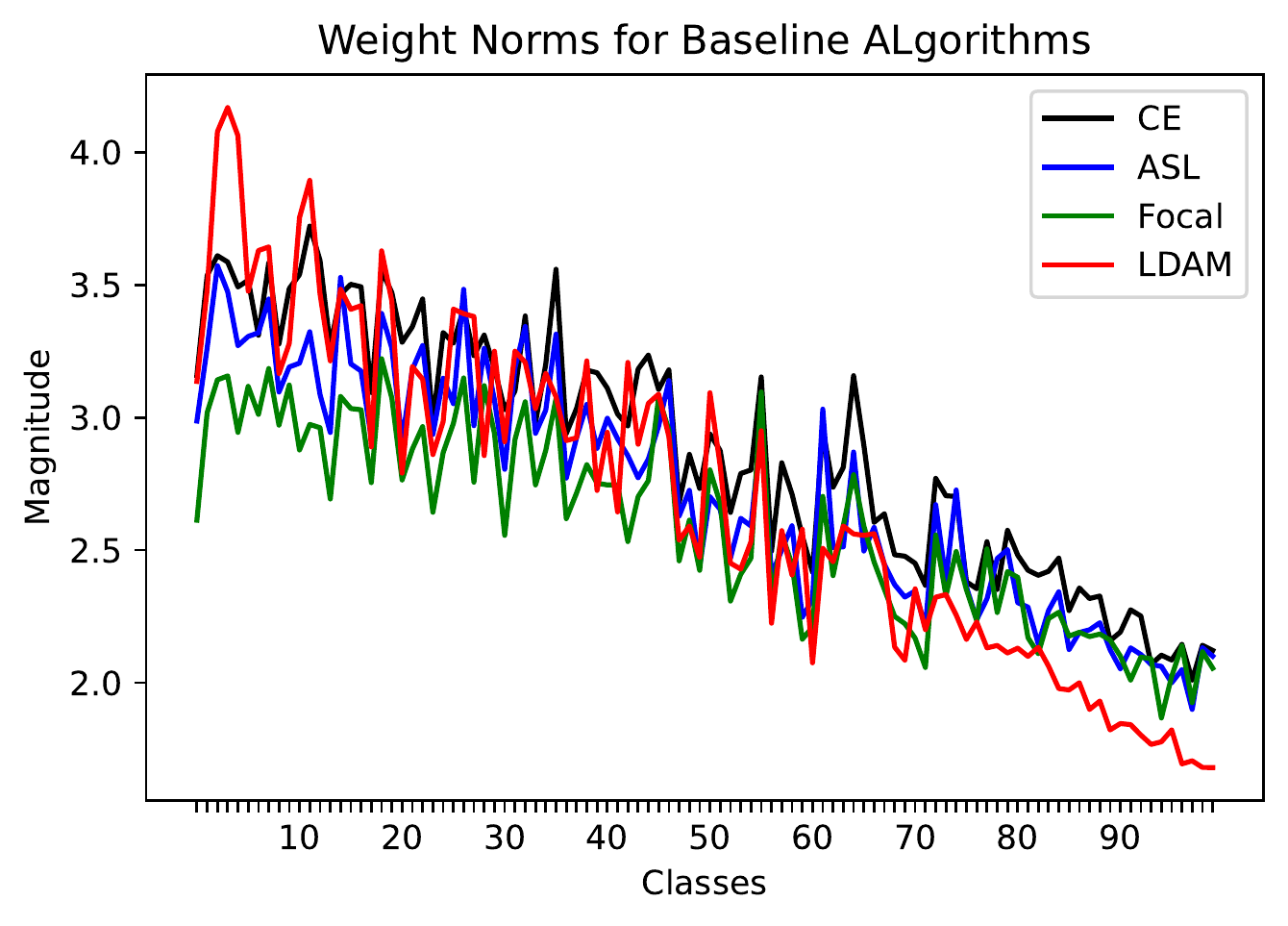}\label{fig:f10e}}
  \hfill
  \subfloat[CIFAR100:CE]{\includegraphics[width=0.2\textwidth]{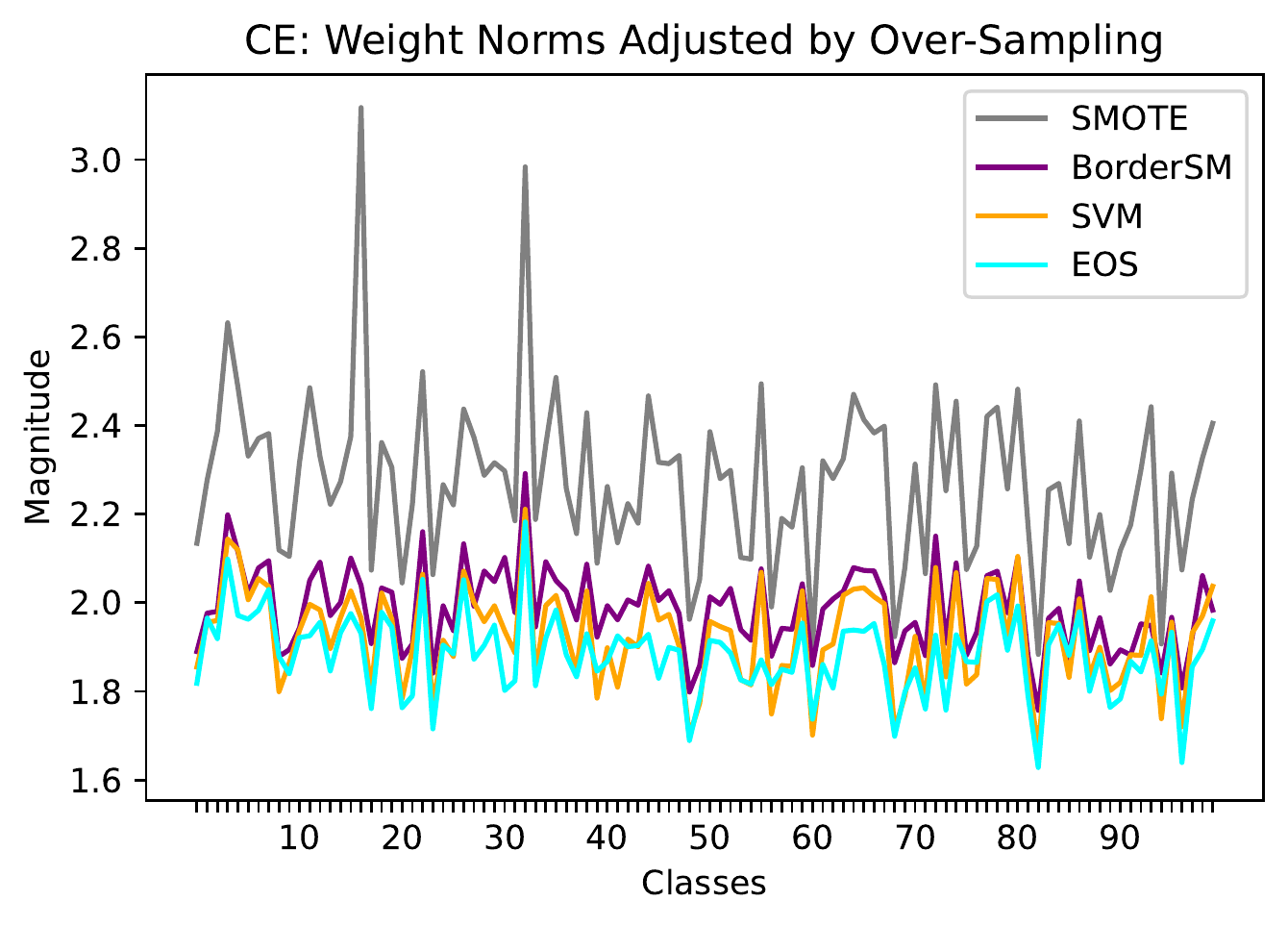}\label{fig:f10f}}
  \hfill
  \subfloat[CIFAR100:ASL]{\includegraphics[width=0.2\textwidth]{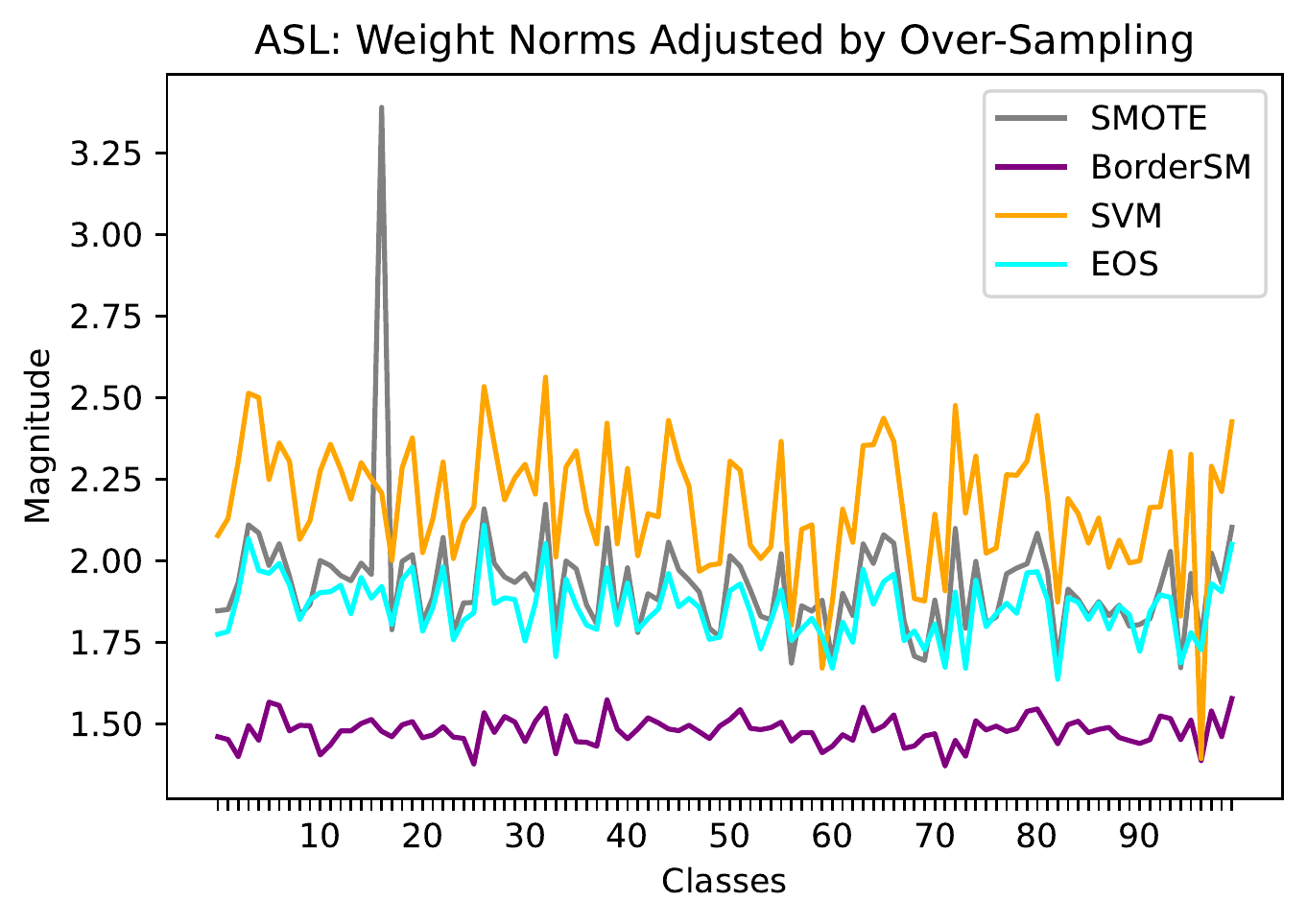}\label{fig:f10g}}
  \hfill
  \subfloat[CIFAR100:Focal]{\includegraphics[width=0.2\textwidth]{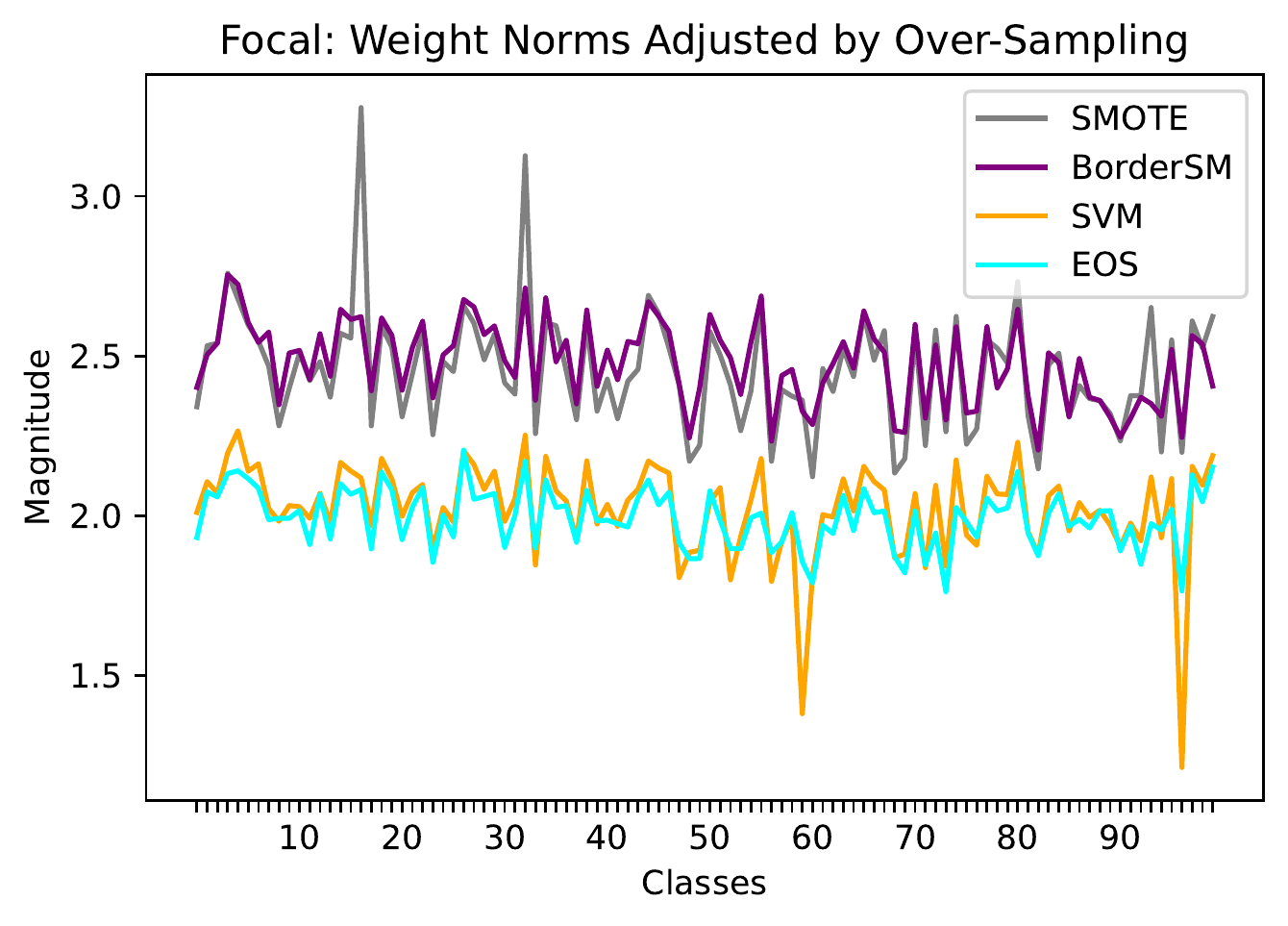}\label{fig:f10h}}
  \hfill
  \subfloat[CIFAR100:LDAM]{\includegraphics[width=0.2\textwidth]{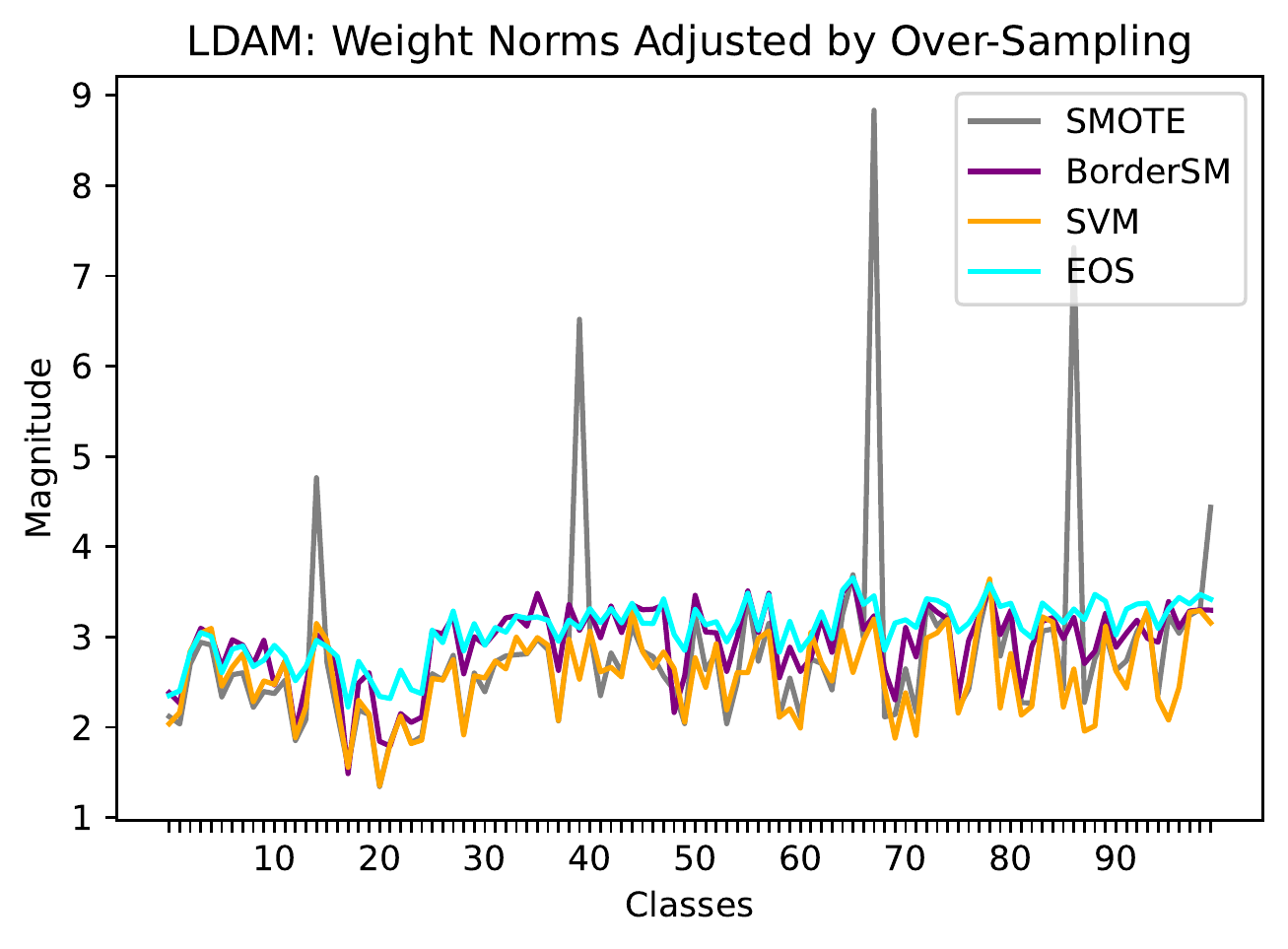}\label{fig:f10i}}
  
  \hfill
  \subfloat[CelebA:Base]{\includegraphics[width=0.2\textwidth]{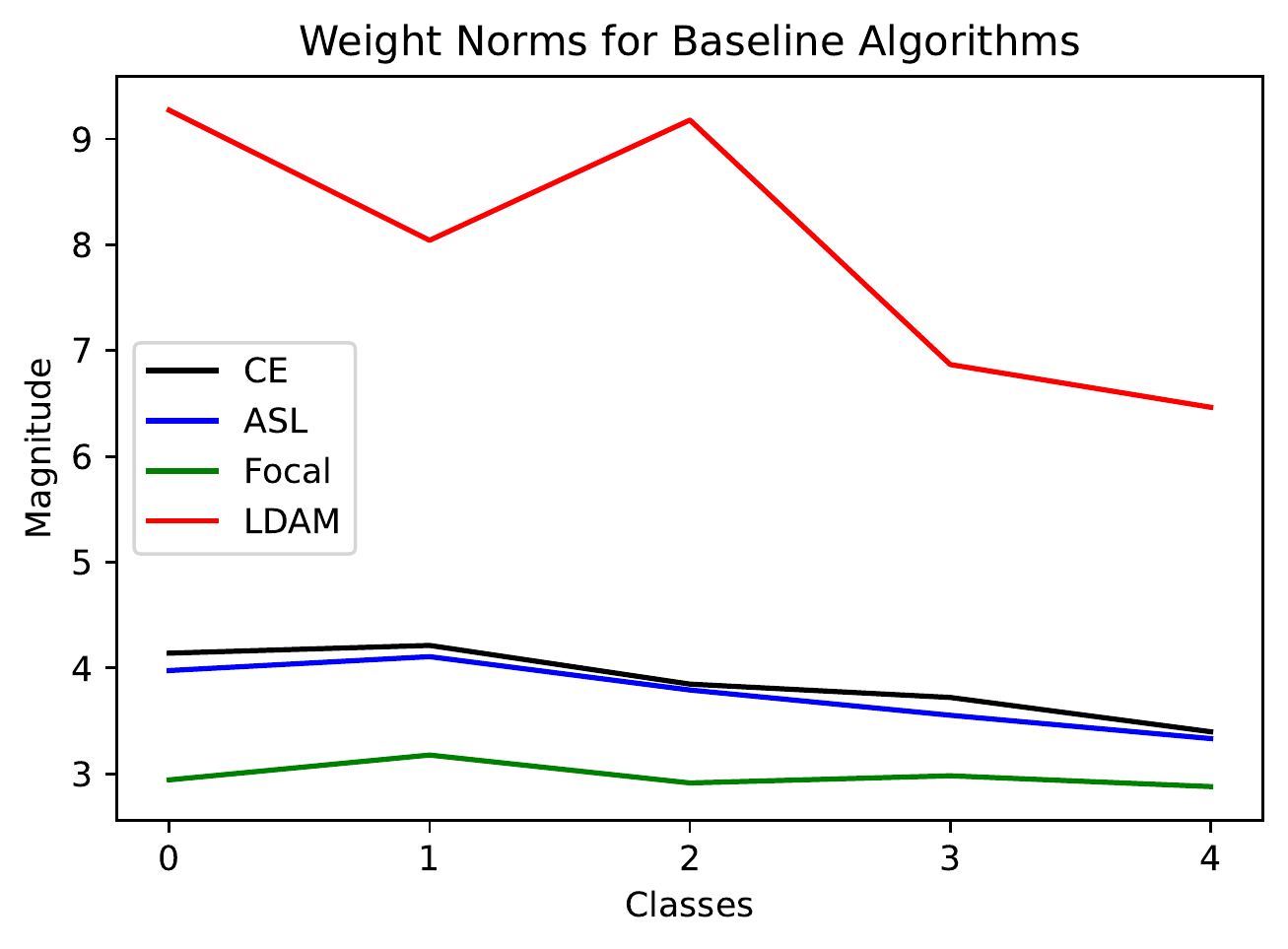}\label{fig:f10e2}}
  \hfill
  \subfloat[CelebA:CE]{\includegraphics[width=0.2\textwidth]{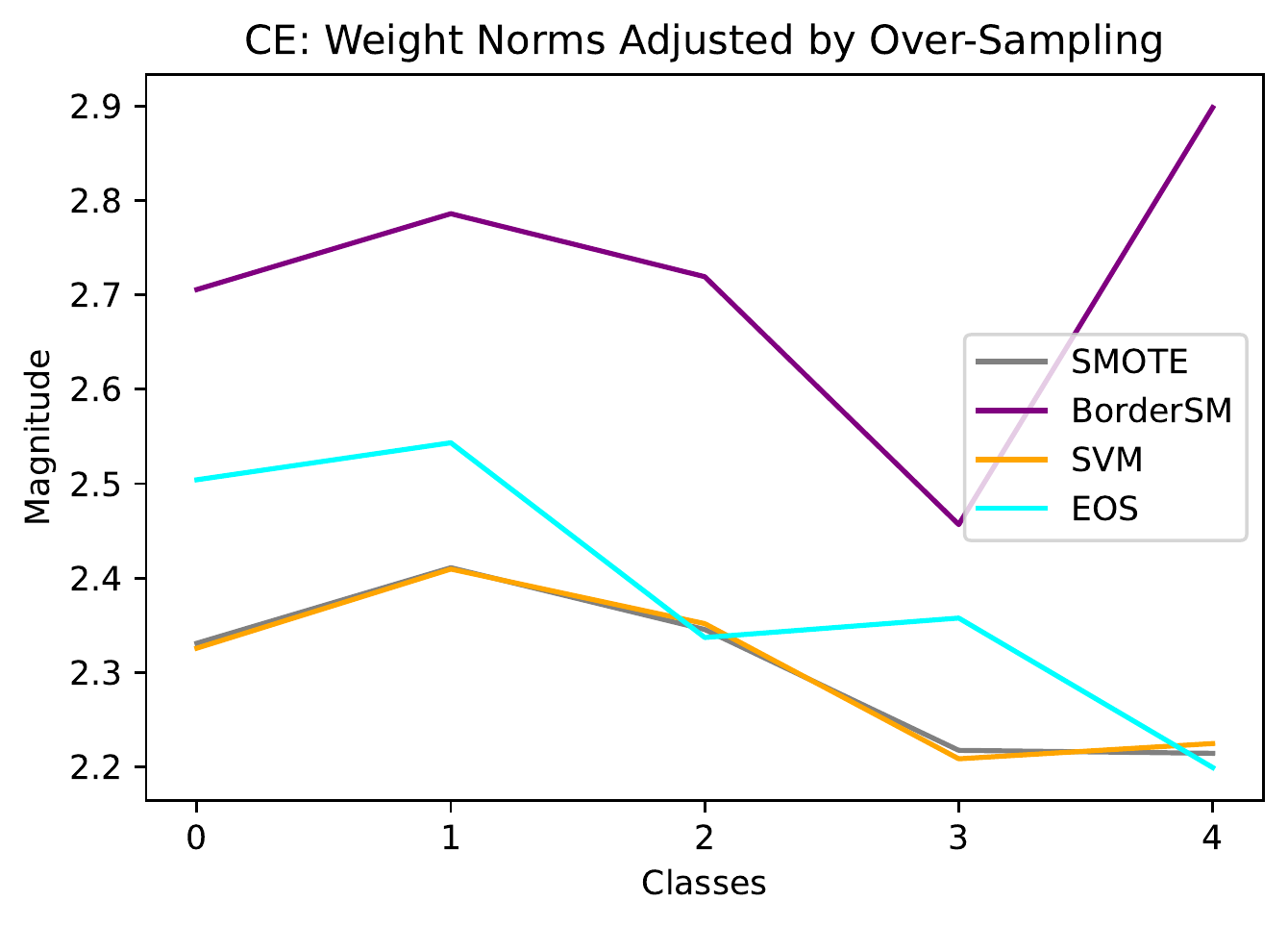}\label{fig:f10f2}}
  \hfill
  \subfloat[CelebA:ASL]{\includegraphics[width=0.2\textwidth]{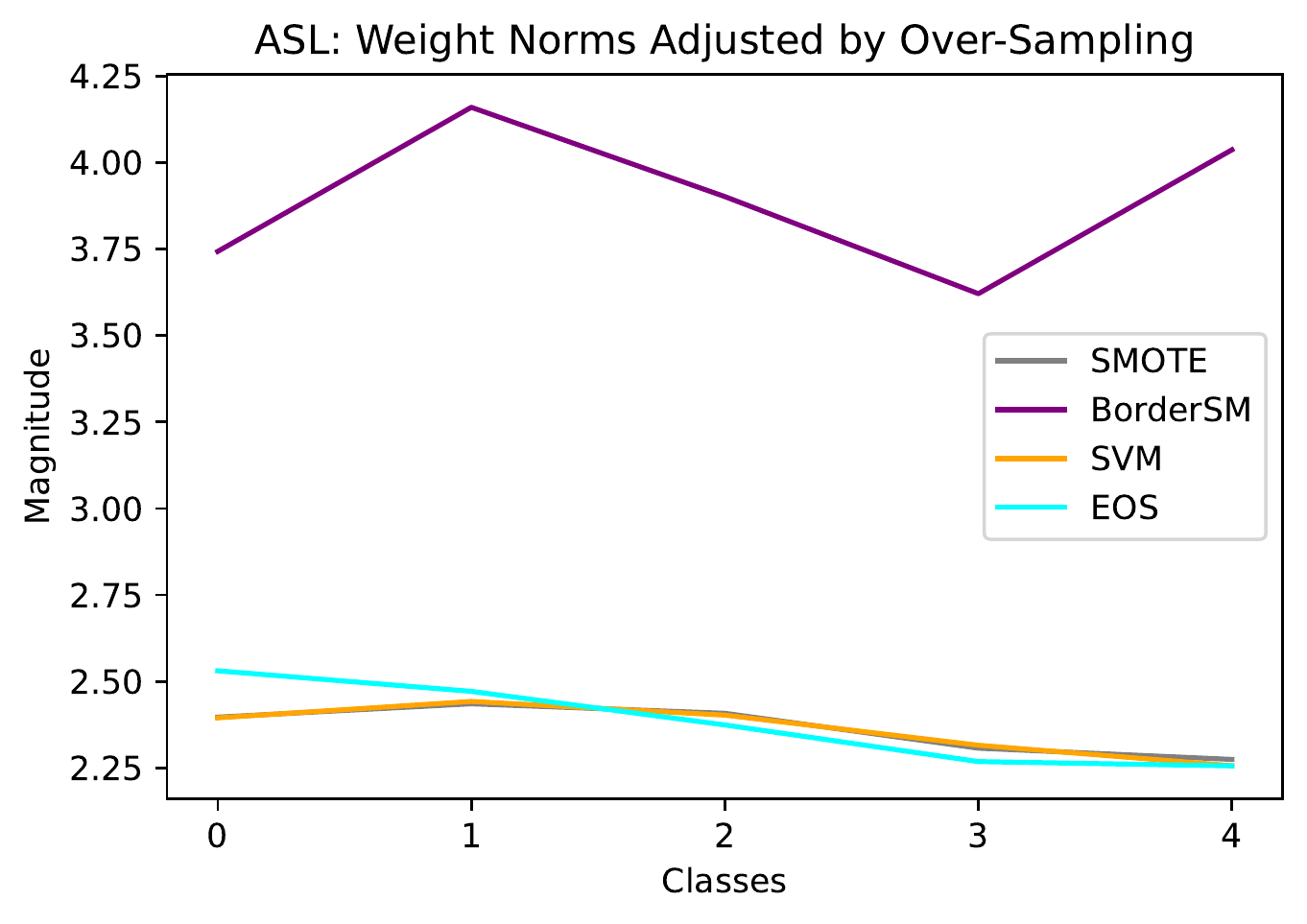}\label{fig:f10g2}}
  \hfill
  \subfloat[CelebA:Focal]{\includegraphics[width=0.2\textwidth]{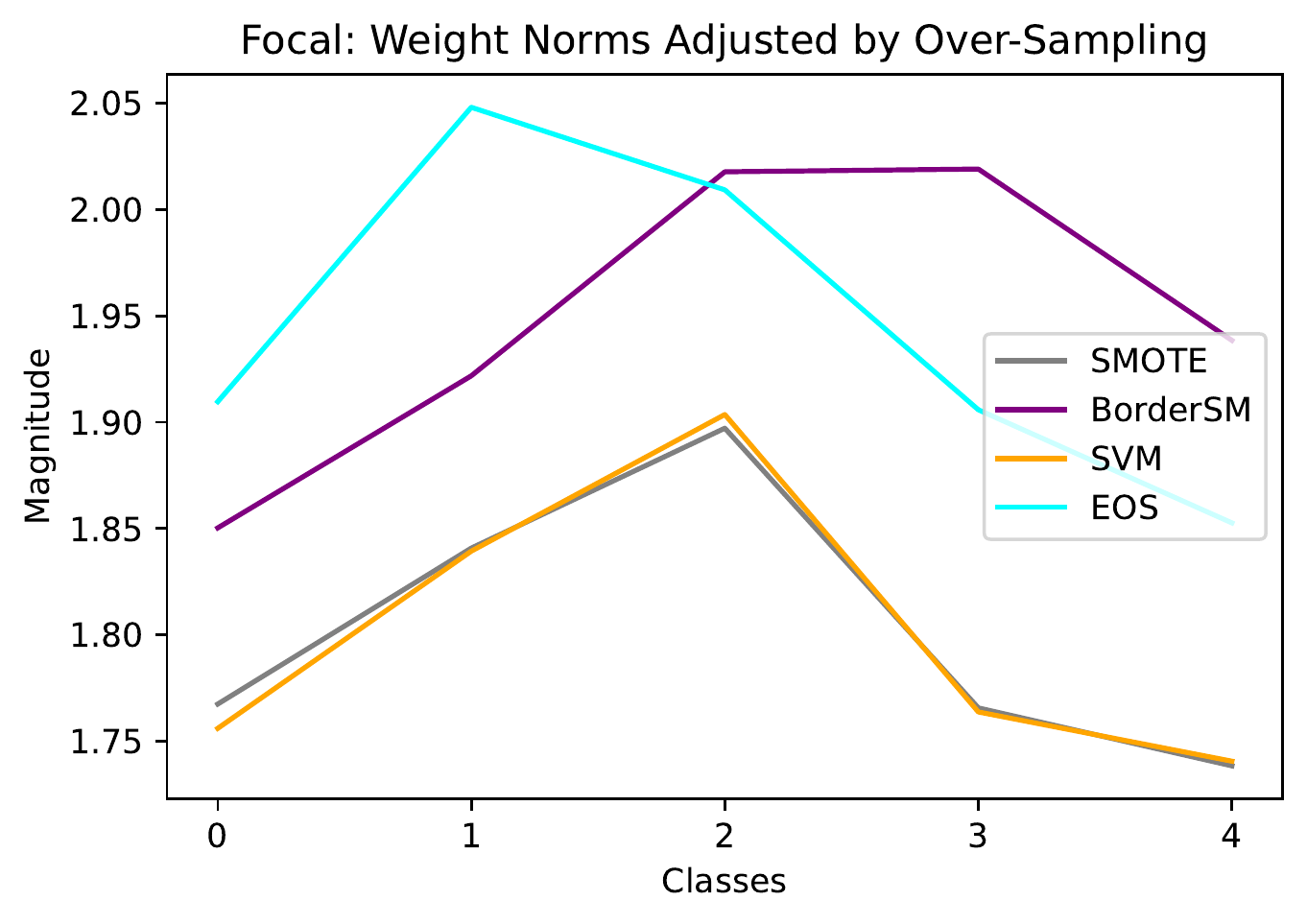}\label{fig:f10h2}}
  \hfill
  \subfloat[CelebA:LDAM]{\includegraphics[width=0.2\textwidth]{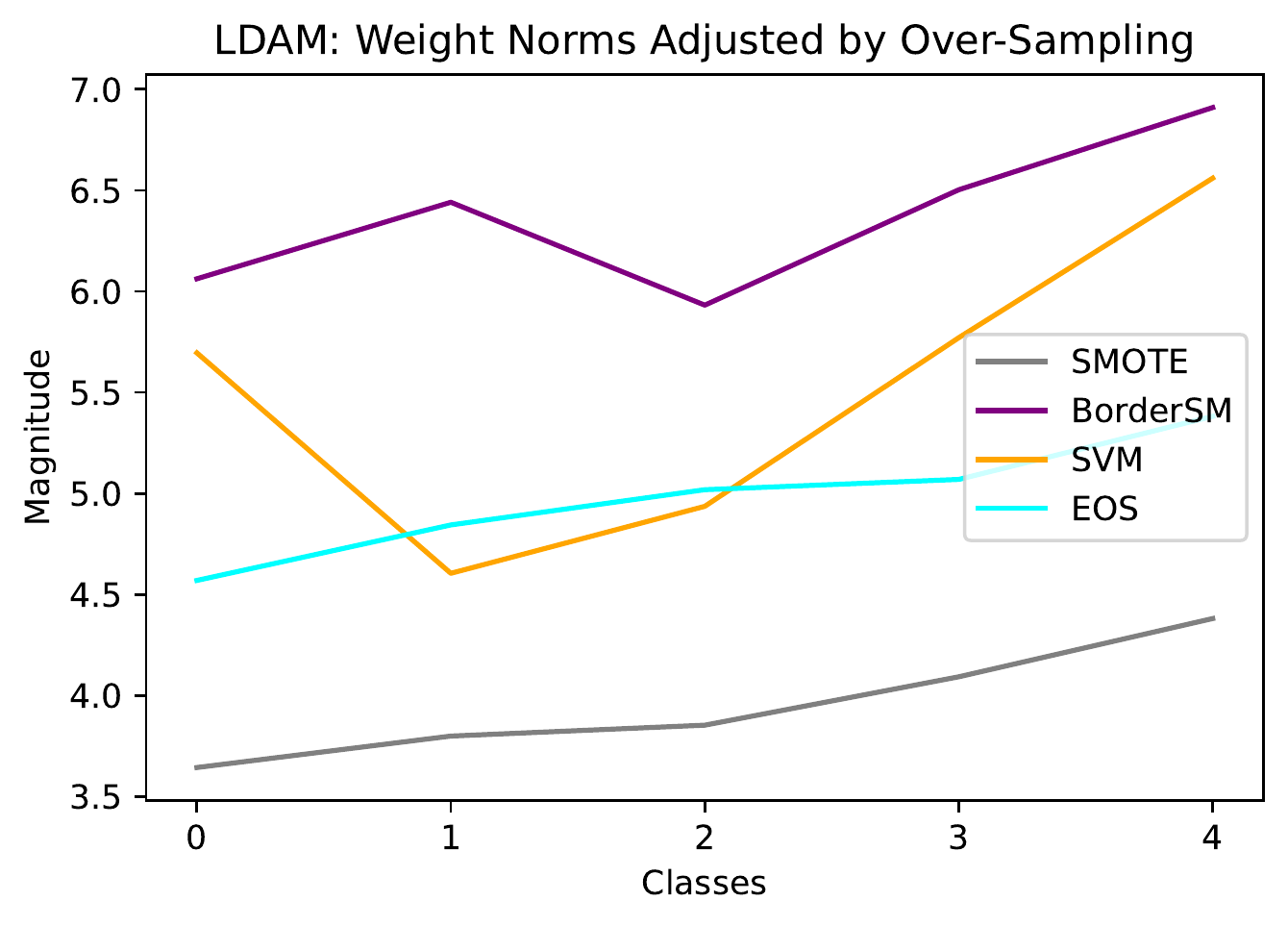}\label{fig:f10i2}}

  \caption{This diagram shows the weight norms (y-axis) for each class (x-axis) before over-sampling for each algorithm, with imbalance (class 0 has the largest number of samples). EOS generally displays larger, more evenly balanced weight norms between classes compared to other over-sampling algorithms.}
  \label{fig_norm}
  \vspace{-0.4cm}
\end{figure*}

 \begin{figure*}[b!]
  \vspace{-0.6cm}
  \centering
  \subfloat[Baseline]{\includegraphics[width=0.2\textwidth]{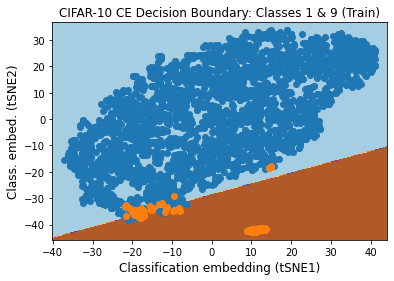}\label{fig:f1}}
  \hfill
  \subfloat[SMOTE]{\includegraphics[width=0.2\textwidth]{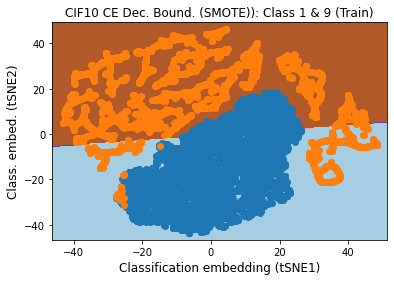}\label{fig:f2}}
   \hfill
  \subfloat[B. SMOTE]{\includegraphics[width=0.2\textwidth]{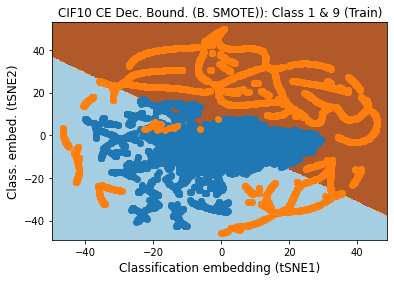}\label{fig:f3}}
   \hfill
  \subfloat[Balanced SVM]{\includegraphics[width=0.2\textwidth]{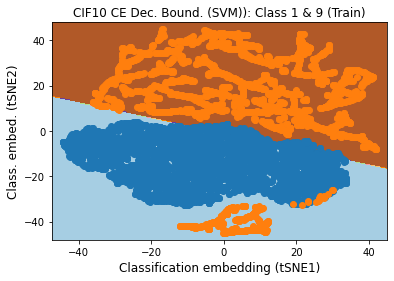}\label{fig:f4}}
  \hfill
  \subfloat[EOS]{\includegraphics[width=0.2\textwidth]{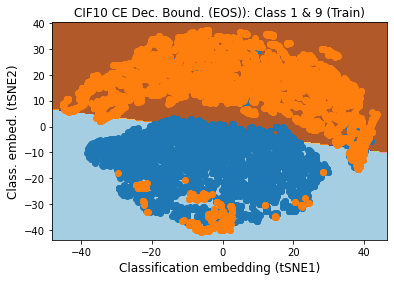}\label{fig:f5}}
  \caption{(a) Illustration of decision boundary for a Resnet-32 trained on cross-entropy loss for the  CIFAR-10 dataset. To simplify the visualization, only two classes are shown: automobiles in blue (majority class) and trucks in red (minority class), with t-SNE. The EOS local structure representation is more dense, uniform and conveys a larger margin than the other methods.}
  \label{fig_dec_bnd}
  \vspace{-0.5cm}
\end{figure*}

As can be seen in Figure~\ref{fig_norm}, traditional over-sampling methods have an uneven impact on classifier weight norms, when they are applied in feature embedding space. Before over-sampling, the weight norms gradually decrease in magnitude for minority classes (see sub-figures (a), (f), (k), and (p) of Figure~\ref{fig_norm}).  This trend continues even after traditional over-sampling methods are applied, although in some cases, the methods do succeed in flat-lining the weight norms between majority and minority classes.  EOS, depicted as a blue line in Figure~\ref{fig_norm}, generally balances the weight norms for classes, albeit by no means perfectly, and it usually exhibits the largest weight norms. However, the picture is uneven, which implies that EOS must have some other effect on the classifier than merely equalizing weight norms.    

We hypothesize that EOS' expansion of the range of FE and its reduction in the generalization gap are the reasons for its performance.  The effect of EOS on the generalization gap is clearly illustrated in Figure~\ref{fig_gen_gap}.  The baseline cost-sensitive methods and traditional over-sampling methods all show a rising generalization gap for minority classes, which generally follows the imbalance level.  In fact, the graphically illustrated generalization gap curves for all of the over-sampling methods overlap, except for EOS (orange line), which reduces the gap for all algorithms and datasets for the extreme minority classes.

Traditional over-sampling methods, such as SMOTE, Border-line SMOTE and Balanced SVM do not expand the range of feature embeddings because they are inherently interpolative.  In other words, they randomly find new FE \textit{within} same-class training examples.  This approach does not reduce the generalization gap between the training and test feature embedding distributions because it merely samples from \textit{within} a fixed feature embedding \textit{range}.  

In contrast, EOS expands the FE range for minority classes.  It randomly generates FE by identifying a same class and non-same class example that are close in space. It calculates a difference between a base, same-class example and a nearest enemy example and then adds a portion of this difference to the base example multiplied by a randomly selected number between [0,1]. 

\begin{table*}[t!]
\vspace{-0.2cm}
\centering
\footnotesize
\caption{Comparison with GAN-based oversampling approaches.}
\label{tab:gan}
\begin{tabular}{ p{.8cm}p{.6cm}p{.6cm}
p{.6cm}p{.6cm}p{.6cm}p{.6cm}
p{.6cm}p{.6cm}p{.6cm}p{.6cm}
p{.6cm} p{.6cm}}
\toprule

\multicolumn{1}{l}{\textbf{Descr}} & 
\multicolumn{3}{l}{\textbf{GAMO}} & 
\multicolumn{3}{l}{\textbf{BAGAN}} & 
\multicolumn{3}{l}{\textbf{CGAN}} &
\multicolumn{3}{l}{\textbf{EOS}}
\\
\midrule

Algo & BAC & GM & FM & BAC & GM & FM &
BAC & GM & FM  & BAC & GM & FM  \\

\midrule

\multicolumn{12}{l}{\textbf{CIFAR 10}}\\
\midrule

CE & .7382 & .8403 & .7351 &
.7304 & .8299 & .7288 &
\textbf{.7653} & \textbf{.8614} & \textbf{.7619} &
.7581 & .8589 & .7571 \\

ASL & .7294 & .8315 & .7156 &
.7248 & .8219 & .7196 &
.7802 & .8699 & .\textbf{7829} &
\textbf{.7825} & \textbf{.8738} & .7827\\

Focal & .7187 & .8093 & .7148 &
.7205 & .8146 & .7188 &
.7792 & .8698 & .7815 &
\textbf{.7831} & \textbf{.8742} & \textbf{.7830}\\

LDAM & .7692 & .8691 & .7651 &
.7644 & .8581 & .7619 &
.7862 & .8708 & .7850 &
\textbf{.7865} & \textbf{.8763} & \textbf{.7862}\\

\midrule

\multicolumn{12}{l}{\textbf{SVHN}}\\
\midrule

CE & .8548 & .9013 & .8482 &
.8581 & .9081 & .8504 &
.8903 & .9297 & .8893 &
\textbf{.9016} & \textbf{.9443} & \textbf{.9014} \\

ASL & .8319 & .8947 & .8288 &
.8346 & .8958 & .8291 &
.8826 & .9155 & .8807 &
\textbf{.9005} & \textbf{.9437} & \textbf{.9003}\\

Focal & .8321 & .8953 & .8301 &
.8358 & .9009 & .8301 &
.8719 & .9028 & .8711 &
\textbf{.8913} & \textbf{.9384} & \textbf{.8909}\\

LDAM & .8611 & .9102 & .8602 &
.8607 & .9116 & .8591 &
.8921 & .9313 & .8910 &
\textbf{.9093} & \textbf{.9487} & \textbf{.9092}\\

\midrule

\multicolumn{12}{l}{\textbf{CIFAR 100}}\\

\midrule
CE & .5204 & .7288 & .5216 &
.5244 & .7348 & .5251 &
.5788 & .7588 & \textbf{.5786} &
\textbf{.5794} & \textbf{.7596} & \textbf{.5786} \\

ASL & .5173 & .7206 & .5194 &
.5214 & .7268 & .5239 &
.5692 & .7503 & .5701 &
\textbf{.5722} & \textbf{.7548} & \textbf{.5727}\\

Focal & .5182 & .7214 & .5207 &
.5213 & .7266 & .5232 &
.5599 & .7411 & .5603 &
\textbf{.5633} & \textbf{.7489} & \textbf{.5623}\\

LDAM & .5262 & .7301 & .5274 &
.5248 & .7312 & .5261 &
\textbf{.5744} & \textbf{.7583} & \textbf{.5624} &
.5732 & .7555 & \textbf{.5624}\\

\midrule

\multicolumn{12}{l}{\textbf{CelebA}}\\

\midrule
CE & .6600 & .7771 & .6301 &
.6609 & .7777 & .6284 &
\textbf{.8044} & \textbf{.8747} & \textbf{.8023}  &
\textbf{.8044} & \textbf{.8747} & \textbf{.8023} \\

ASL & .6281 & .7429 & .6004 &
.6294 & .7502 & .6038 &
.7019 & .7922 & .7281 &
\textbf{.7758} & \textbf{.8558} & \textbf{.7729}\\

Focal & .6580 & .7753 & .6319 &
.6628 & .7581 & .6392 &
.8003 & .8698 & .7877 &
\textbf{.8010} & \textbf{.8742} & \textbf{.7990}\\

LDAM & .6736 & .7825 & .6411 &
.6751 & .7829 & .6448 &
.8056 & \textbf{.8760} & .8032 &
\textbf{.8064} & \textbf{.8760} & \textbf{.8055}\\

\bottomrule

\end{tabular}
\vspace{-0.2cm}
\end{table*}

The impact of EOS on class decision boundaries is graphically illustrated in Figure~\ref{fig_dec_bnd} for cross-entropy on the CIFAR-10 dataset with the automobile (depicted in blue) and trucks (orange) classes. When compared to each other, the imbalance level of autos to trucks is 60:1, with trucks as the minority class.  These two classes were selected due to their perceived similarity and potential for overlap. Figure~\ref{fig_dec_bnd} uses the t-distributed Stochastic Neighbor Embedding (t-SNE) method \cite{van2008visualizing} to visualize the class decision boundary. The t-SNE method is effective at preserving local distances in high-dimensional data, such as neural network embeddings, when those embeddings are converted to a two-dimensional visualization \cite{wattenberg2016use,zhou2018using}. In the t-SNE plots, the local structure of the minority class examples (trucks) in the baseline, SMOTE, Borderline SMOTE and Balanced SVM cases are 
uneven, with varying levels of density, such that there are intra-class gaps in data points.  In the baseline and Borderline SMOTE cases, there are gaps in the auto class, which is also over-sampled.  Only in the EOS visualization is the local structure of the classes represented with uniform, dense points.  Additionally, in the EOS t-SNE illustration, there is a wider local boundary between the nearest enemy points when compared to the other methods.  We hypothesize that this improved representation is due to EOS expanding the range of FE and closing the generalization gap, which results in a denser, more uniform class manifold in embedding space (\textbf{RQ3 answered}).

\subsection{Comparison with GAN-based over-sampling}

 Recent trends in deep learning from imbalanced data focus on using generative models (GANs) to create artificial images and use them to balance the training set. While GANs have shown great success in creating life-like artificial images, we argue that they are not well-suited for imbalanced data and that the massive computational cost incurred by them does not justify the obtained results. 

Table~\ref{tab:gan} presents a comparison between EOS and three state-of-the-art GAN-based over-sampling approaches. GAN-based methods focus on an adversarial approach for generating artificial images that follow the distribution of the training data. They can be seen as model-agnostic pre-processing methods, since datasets balanced by them can be used by any classifier. Thus, they are similar to standard over-sampling approaches, but rely on generative models to infer new instances. EOS works within a specified deep learning classifier and offers superior computational performance, as it does not require model induction.

Looking at the results, we can see that both GAMO and BAGAN display inferior results when compared to EOS. This can be explained by the fact that while the images generated by them look similar to real images, they do not sufficiently enhance the classifier training procedure. Recent advances in imbalanced learning point to the fact that over-sampling should focus not on the quantity of generated instances, but on their quality \cite{Krawczyk:2020}. Artificial minority instances should be injected in dedicated areas of the feature space to alleviate the bias towards majority classes \cite{Koziarski:2020} and address instance-level learning difficulties such as small sub-concepts and borderline overlap between classes \cite{Ghosh:2021}. The obtained results show that both GAMO and BAGAN fail to inject meaningful instances into the training set, leading to underwhelming performance, especially when dealing with a high number of classes (see CIFAR-100). At the same time, EOS utilizes information about the generalization gap and other class topologies (realized as nearest enemy distance) and is capable of informed generation of artificial instances in such areas of the feature space (in case of EOS in the embedding space) that directly empower learning from minority classes and effectively overcomes the bias. 

Only CGAN comes close to EOS performance, being able to offer slightly better results when handling a high number of classes (see CIFAR-100). However, there is a cost associated with using the CGAN approach. It trains a separate generative model for each class, making it computationally infeasible with an increased number of classes. Therefore, it is impractical to use CGAN for CIFAR-100 or long-tail recognition problems that have hundreds or thousands of classes. EOS offers comparable performance to CGAN while relying on very simple and lightweight instance generation, allowing it to scale efficiently to hundreds or thousands of classes (\textbf{RQ4 answered}). 

\subsection{Hyper-parameters, model architecture \& training}

In this section, we discuss the sensitivity of hyper-parameter, model architecture, and training regime changes on EOS performance, using the CIFAR-10 dataset.

\subsubsection {Number of epochs} 
\begin{figure}[h!]
\vspace{-0.2cm}
  \centering
  \includegraphics[width=0.4\textwidth]{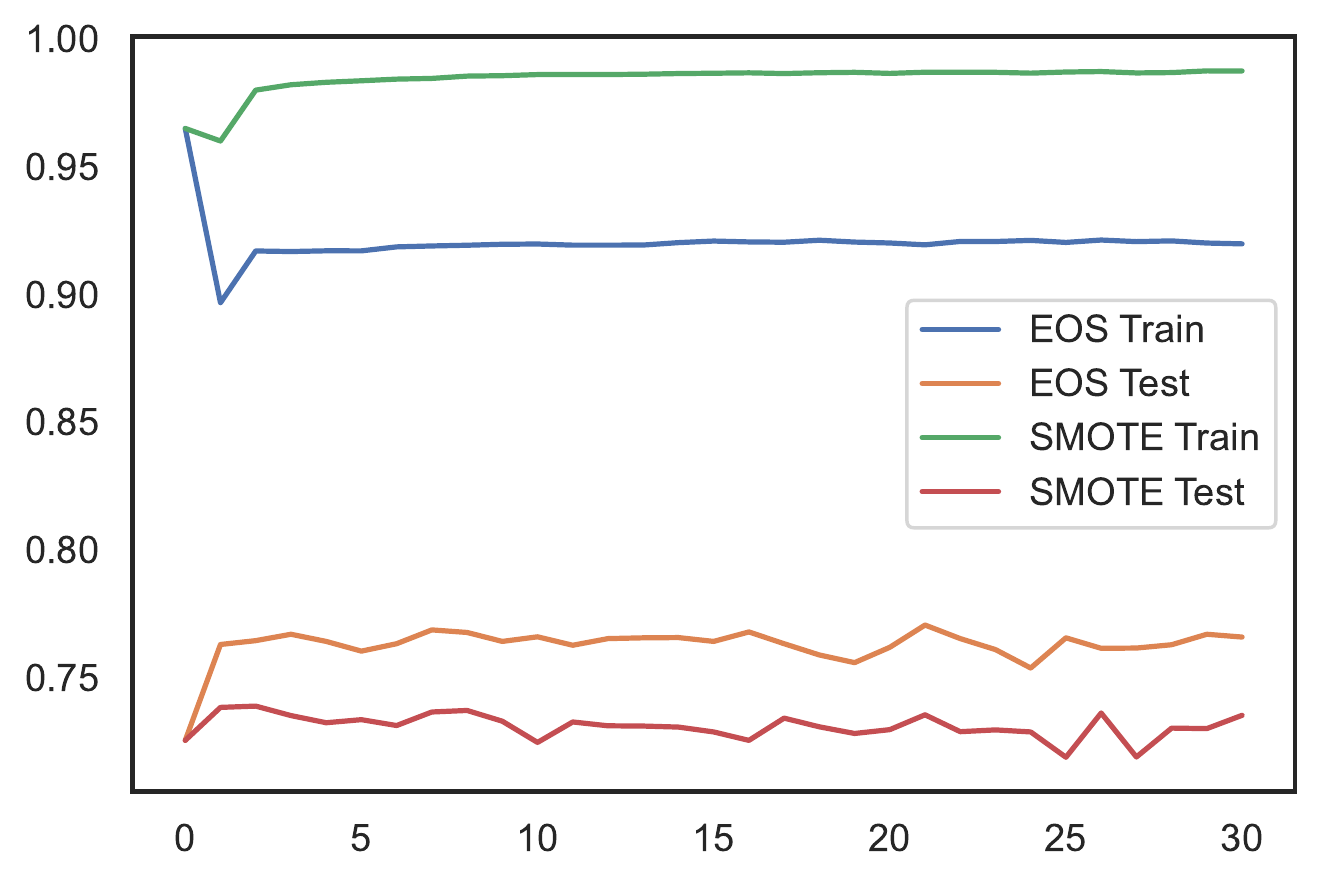}\label{fig:epoch1}
  
  \caption{This figure shows that there is a marginal improvement in EOS test balanced accuracy for classifier re-training beyond 10 epochs and no balanced accuracy improvement for SMOTE. Both methods essentially "flat-line" by epoch 10 for CIFAR-10.}
  \label{fig_epochs}
  \vspace{-0.5cm}
\end{figure}

We consider the impact of increasing the number of epochs on training and test balanced accuracy (BAC). Figure~\ref{fig_epochs} plots the balanced training and test accuracy of EOS and a baseline algorithm, SMOTE, for 30 epochs, using cross-entropy loss on the CIFAR-10 dataset.  Both methods essentially plateau by epoch 10 in terms of training and test accuracy; however, EOS achieves some marginal improvement by training longer.

\subsubsection{Model run time} In this section, we offer a comparison of model run time for EOS versus pre-processing methods. For this purpose, we use the same training regime discussed in Section~\ref{sec:set}. We compare EOS to the average training time for a Resnet-32 trained with a CIFAR-10 dataset that is alternatively augmented with SMOTE, Borderline SMOTE and Balanced SVM. The average training time for these three pre-processing methods is 126.9 minutes. For EOS, the training time is 43.9 minutes, which consists of the time to train a CNN and  classifier re-training.  Thus, model training time with front-end data augmentation is approx. 3 times more expensive in terms of compute time.

There are three reasons for this difference: (1) data augmentation with lower dimensional data, (2) training for fewer epochs with augmented data, and (3) implementing augmentation with a model that has fewer parameters. In our experiment, pre-processing with augmented data involves a Resnet-32 with approx. 464K parameters with 50K data instances of size 3X32X32. In contrast, EOS involves initial training of a Resnet-32 with 464K parameters on only 12K data instances (due to imbalance) and re-training a classifier with less than 1K parameters with 50K feature instances of size 1X64. Therefore, the reduction in training time is due to re-training a smaller model with fewer parameters with compressed data. The classifier re-training also only occupies 10 epochs, whereas the larger model is trained for 200 epochs.


\label{sec:knn} In this section, we show the effect of K Nearest Neighbor selection on EOS performance. For purposes of the experiments discussed in previous sections, we used $K=10$. Table~\ref{tab: nnb} displays the effect of $K \in \{10, 50, 100,200,300\}$  with cross-entropy loss.  In general, increasing K results in improved BAC; however, BAC plateaus at $K=300$, which also experiences a slight decline in F1 measure.

\begin{table}[h!]
\scriptsize
\caption{EOS: Nearest Neighbor Size Analysis}
\label{tab: nnb}
\centering
\begin{tabular}{ p{1cm}p{.6cm}
p{.6cm}p{1.2cm}p{.6cm}
p{.6cm}p{.6cm}}
\toprule

\multicolumn{1}{l}{\textbf{Neighbors}} & 
\multicolumn{1}{l}{\textbf{BAC}} & 
\multicolumn{1}{l}{\textbf{GM}} &
\multicolumn{1}{l}{\textbf{FM}} &
\multicolumn{1}{l}{\textbf{BAC}} & 
\multicolumn{1}{l}{\textbf{GM}} &
\multicolumn{1}{l}{\textbf{FM}}

\\
\midrule
\multicolumn{4}{l}{CIFAR-10}&\multicolumn{3}{l}{SVHN}\\

\midrule

10 & .7581 & .8589 & .7571  & .9016 & .9443 & .9014 \\

50 & .7618 & .8612 & .7618  & .9024 & .9461 & .9028 \\

100 & .7676 & .8647 & .7670  & .9081 & .9489 & .9049 \\

200 & .7717 & .8673 & .7723  & .9102 & .9499 & .9072  \\

300 & .7722 & .8676 & .7715 & .9103 & .9500 & .9098 \\

\midrule
\multicolumn{4}{l}{CIFAR-100}&\multicolumn{3}{l}{CelebA}\\

\midrule

10 & .5794 & .7596 & .5786  & .8044 & .8747 & .8023 \\

50 & .5837 & .7620 & .5795  & .8081 & .8769 & .8057 \\

100 & .5871 & .7649 & .5812  & .8095 & .8792 & .8078 \\

200 & .5884 & .7651 & .5815  & .8123 & .8804 & .8101  \\

300 & .5883 & .7643 & .5814 & .8127 & .8802 & .8116 \\

\bottomrule

\end{tabular}
\vspace{-.2cm}
\end{table}

The reason why an enlarged neighborhood generally improves EOS performance is that it facilitates an expansion in the range of minority features.  EOS operates by selecting nearest adversary class neighbors.  If the neighborhood of adversary class instances is larger, then there is a more diverse range of features that are available to expand minority class instances.  

\subsubsection{EOS performance in pixel space} We also assess the performance of EOS in pixel, versus feature embedding, space. For this experiment, we use the same training strategy outlined in Section~\ref{sec:set} on CIFAR-10, except that we implement EOS as a pre-processing step. Top BAC, GM and FM are .6881, .8150, and .6889, respectively. These results are lower than EOS performance when the algorithm is implemented in feature embedding space by approx. 7 BAC points (see Table~\ref{tab: results}).  We conjecture that the lower results are attributable to nearest adversary and minority class pairs in pixel space not being as discriminative as feature embedding pairs. In the case of EOS augmentation implemented in feature embedding space, the nearest adversary and minority class pairs reflect the model's learned embeddings. Hence, it  allows for richer gradients and a more precise expansion of minority class features. 

\subsubsection{EOS performance on different CNN architectures}

Here, we test EOS on three additional CNN architectures: a ResNet 56 \cite{he2016deep}, a Wide Residual Network (WideResNet) \cite{zagoruyko2016wide}, and a Densely Connected CNN (DenseNet) \cite{huang2017densely}.


\begin{table}[h!]
\footnotesize
\caption{Different CNN Architectures With \& Without EOS (CIFAR-10)}
\label{tab: oth_nets}
\centering
\begin{tabular}{ p{2.2cm}p{.6cm}
p{.6cm}p{.6cm}}
\toprule

\multicolumn{1}{l}{\textbf{Networks}} & 
\multicolumn{1}{l}{\textbf{BAC}} & 
\multicolumn{1}{l}{\textbf{GM}} &
\multicolumn{1}{l}{\textbf{FM}} 

\\
\midrule

ResNet 56 & .7219 & .8364 & .7188 \\
WideResNet & .7618 & .8612 & .7618 \\
DenseNet & .7425 & .8493 & .7392 \\
\midrule
EOS: ResNet 56 & .7362 & .8454 & .7339 \\

EOS: WideResNet & .7879 & .8771 & .7867 \\

EOS: DenseNet & .7820 & .8735 & .7817 \\

\bottomrule
\end{tabular}
\vspace{-.1cm}
\end{table}


We trained the ResNet 56 and DenseNet for 200 epochs on an exponentially imbalanced CIFAR-10 dataset.  We trained the WideResNet for 100 epochs because it contains approx. 5X the number of parameters as the other two models and hence began to over-fit at an earlier stage.  The performance results of the models with and without EOS classifier re-training are presented in Table~\ref{tab: oth_nets}, based on $K=10$.  This table demonstrates that EOS can be applied successfully on a variety of CNN architectures to improve imbalanced image data classification performance.

\section{Lessons learned}
\label{sec:les}

In order to summarize the knowledge we extracted through the extensive experimental evaluation, this section presents the lessons learned and recommendations for future researchers.

\smallskip
\noindent \textbf{Superiority of oversampling in embedding space.} Two common assumptions in the over-sampling literature are: (i) generating artificial images in pixel space to enrich the training set is the best solution to make deep models skew-insensitive; and (ii) over-sampling is treated as a pre-processing and model-agnostic solution to class imbalance. We have showed that for deep learning these two assumptions do not hold, as using representations learned by deep models for balancing the training set yields far superior results than operating in raw pixel space. This shows that effective augmentation via over-sampling does not require generative models and should be connected with a deep learning model. This is a significant shift from most existing approaches discussed in the literature and opens new directions for research in de-biasing deep neural networks.

\smallskip
\noindent \textbf{Need for measures to evaluate embeddings for imbalanced data.} The quality of feature embeddings has a direct impact on the effects of over-sampling. Therefore, a measure is needed to evaluate the effectiveness of these embeddings at discriminating between classes. We introduced the generalization gap as such a metric and showed how to embed it as a part of the EOS algorithm to improve accuracy in multi-class imbalanced problems. 

\smallskip
\noindent \textbf{GAN limitations at augmenting imbalanced data.} While GAN-based solutions for class imbalance attract attention from the deep learning community, we have shown limitations of such approaches. In learning from imbalanced data, we are less concerned with the visual quality of generated images and more with their impact on classifier bias. While GANs are capable of generating visually stunning artificial images, they display less control over the positioning of such images with regard to decision boundaries. Therefore, their impact on skewed classes is less significant than reference over-sampling methods operating in embedding spaces. Our experiments have highlighted that using topological information about embedding spaces together with a dedicated measure of embedding quality leads to superior placement of artificial instances. Furthermore, GANs rely on computationally costly models, often requiring users to store one model per class. This makes them unfeasible for problems with hundreds or thousands of classes (such as long-tailed recognition). On the contrary, EOS working in embedding space offers low cost and effective over-sampling for alleviating bias in deep neural networks.

\section{Conclusion and future work}
\label{sec:con}

\noindent \textbf{Conclusions.} Neural networks have been shown to memorize training data, and yet generalize to unseen test examples. We believe that a CNN's internal representations, in the form of feature and classification embeddings, offer insights into the inner workings of  machine learning models. We use a model's internal representations to measure the generalization gap between the training and test sets, and design a novel over-sampling technique to improve model accuracy.    

\smallskip
\noindent \textbf{Future work.} Our future work will focus on designing new measures complementary to the proposed generalization gap and investigating how to create and utilize effective embeddings for imbalanced data.  As we have seen that the generalization gap can lead to effective over-sampling, we envision  creating complementary measures will lead to a better understanding the characteristics of multi-class imbalanced problems. This will allow us to modify the feature extraction process to create unbiased embeddings that will enhance the effects of over-sampling. Combining imbalance-aware representation learning \cite{Khan:2018} with over-sampling in embedding space will allow for even more robust training of deep neural networks.

\bibliographystyle{IEEEtran}
\bibliography{references}

\end{document}